\documentclass[11pt]{article}
\pdfoutput=1
\usepackage{enumerate}
\usepackage{pdfsync}
\usepackage[OT1]{fontenc}
\date{}
\usepackage[usenames]{color}
\usepackage{smile}
\usepackage[colorlinks,
            linkcolor=red,
            anchorcolor=blue,
            citecolor=blue
            ]{hyperref}

\usepackage{fullpage}
\usepackage[protrusion=true,expansion=true]{microtype}
\usepackage{pbox}
\usepackage{setspace}
\usepackage{tabularx}
\usepackage{float}
\usepackage{wrapfig,lipsum}
\usepackage{enumitem}
\usepackage{microtype}
\usepackage{graphicx}
\usepackage{enumerate}
\usepackage{enumitem}
\usepackage{pgfplots}
\usepackage{natbib}

\usepackage{amsfonts}
\usepackage{bm}
\usepackage{xcolor}
\usepackage{multibib}
\usepackage{array}
\usepackage{multirow}
\usepackage{tabularx}
\usepackage{amsmath}
\usepackage{bbm}
\usepackage{caption}
\usepackage{subcaption}
\usepackage{booktabs}
\usepackage{url}
\usepackage{authblk}

\usetikzlibrary{arrows,shapes,snakes,automata,backgrounds,petri}
\usepackage{booktabs} 

\allowdisplaybreaks

\usepackage{enumitem}

\usepackage{colortbl}
\definecolor{LightCyan}{rgb}{0.8, 0.9, 1}
\usepackage{enumitem}

\makeatletter
\newcommand*{\rom}[1]{\expandafter\@slowromancap\romannumeral #1@}
\makeatother

\newcommand{\ip}[2]{{\langle #1, \, #2 \rangle}}
\newcommand{\lrf}[1]{{\lfloor #1 \rfloor}}
\newcommand{\lrc}[1]{{\lceil #1 \rceil}}

\newcommand{\mR}{\mathbb{R}}
\newcommand{\bB}{\mathbf{B}}
\newcommand{\mN}{\mathbb{N}}
\newcommand{\mE}{\mathbb{E}}

\newcommand{\ntheta}{\theta_n^\lambda}
\newcommand{\ceil}[1]{\left\lceil #1 \right\rceil}

\newcommand{\btheta}{\bar \theta^{\lambda}}
\newcommand{\1}{\mathbbm{1}}

\newcommand{\rd}{\mathrm{d}}

\def\rmd{\mathrm{d}}
\def\rset{\mathbb{R}}

\title{Polygonal Unadjusted Langevin Algorithms: Creating stable and efficient adaptive algorithms for neural networks}

\author[1]{Dong-Young Lim\thanks{\texttt{dlim@unist.ac.kr}}}
\author[2,3,4]{Sotirios Sabanis\thanks{\texttt{S.Sabanis@ed.ac.uk}}}
\affil[1]{Department of Industrial Engineering, UNIST, South Korea}
\affil[2]{School of Mathematics, The University of Edinburgh, UK}
\affil[3]{The Alan Turing Institute, UK}
\affil[4]{National Technical University of Athens, Greece}

\begin{document}

\maketitle

\begin{abstract}%
 We present a new class of Langevin-based algorithms, which overcomes many of the known shortcomings of popular adaptive optimizers that are currently used for the fine tuning of deep learning models. Its underpinning theory relies on recent advances of  Euler-Krylov polygonal approximations for stochastic differential equations (SDEs) with monotone coefficients. As a result, it inherits the stability properties of tamed algorithms, while it addresses other known issues, e.g. vanishing gradients in deep learning. In particular, we provide a nonasymptotic analysis and full theoretical guarantees for the convergence properties of an algorithm of this novel class, which we named TH$\varepsilon$O POULA (or, simply, TheoPouLa). Finally, several experiments are presented with different types of deep learning models, which show the superior performance of TheoPouLa over many popular adaptive optimization algorithms.
\end{abstract}

\section{Introduction}

Modern machine learning models including deep neural networks are successfully trained when they are finely tuned by solving the following optimization problem: 
\begin{align}\label{eq:optimizaton}
\text{minimize } u(\theta):= \mE\left[U(\theta, X)\right],
\end{align}
where $\theta \in \mR^d$, $X$ is a $\mR^m$-valued random variable, and $U:\mR^d \times \mR^m \rightarrow \mR$ is the associated loss function.
Two aspects of such optimization tasks pose significant challenges, namely the nonconvex nature of loss function $U$ and the highly nonlinear features of many types of neural networks. Moreover, the analysis in \citet{sabanis:20} shows that the gradients of such nonconvex loss functions typically grow faster than linearly and are only locally Lipschitz continuous. Naturally, stability issues are observed, which are known as the `exploding gradient' phenomenon (\citealt{bengio:94, Pascanu_et_al:13}), when vanilla stochastic gradient descent (SGDs) or certain types of adaptive algorithms are used for fine tuning. In addition, the sparsity of gradients of neural networks is another challenging issue, which is extensively studied in the literature. For example, momentum methods and adaptive learning rate methods such as AdaGrad (\citealt{duchi:11}), RMSProp (\citealt{hinton:12}) and Adam (\citealt{kingma:15}) have been developed to tackle this problem by diagonally scaling the gradient by some function taking the past gradients, also known as preconditioner.

Langevin-based algorithms have been another important stream of literature on stochastic optimization. They are built on the theoretical fact that the Langevin stochastic differential equation (Langevin SDE) defined on $t\in [0,\infty)$ given by,
\begin{equation} \label{eq:lagenvin_sde}
{\rm d} Z_t=-h(Z_t) {\rm d} t+ \sqrt{2\beta^{-1}} {\rm d} B_t, \quad Z_0=\theta_0,
\end{equation}
where $\theta_0 \in \mR^d$ is a (possibly random) initial condition, $h=\nabla u$, $\beta>0$ is the so-called inverse temperature parameter, and $(B_t)_{t\ge0}$ is a $d$-dimensional Brownian motion, converges to a unique invariant measure $\pi_\beta(\theta)\propto \exp(-\beta u(\theta))$ under mild conditions. Moreover, $\pi_\beta$ concentrates around the global minimizers of the objective function $u$ as $\beta\rightarrow \infty$, see \cite{hwang:80}, even in the case of nonconvex potentials. Based on this very attractive property, \citep{welling:11, raginsky:17} proposed the stochastic gradient Langevin dynamics (SGLD), which is the Euler-Maruyama discretization scheme for the Langevin SDE~\eqref{eq:lagenvin_sde} with an unbiased estimator $H:\mR^d \times \mR^m \rightarrow \mR^d$ satisfying $h(\theta)=\mE[H(\theta, X)]$ for all $\theta \in \mR^d$. More specifically, the SGLD algorithm is given by, for $n\in\mN_0$,
\begin{equation}\label{eqn:sgld}
\theta^{\mathsf{SGLD}}_0 :=\theta_0, \quad \theta^{\mathsf{SGLD}}_{n+1}:=\theta^{\mathsf{SGLD}}_n-\lambda H(\theta^{\mathsf{SGLD}}_n,X_{n+1})+ \sqrt{2\lambda\beta^{-1}} \xi_{n+1},
\end{equation}
where $\lambda>0$ is the learning rate, $(X_n)_{n\in \mN_0}$ is an i.i.d. sequence of random variables, $(\xi_n)_{n\in \mN_0}$ is an independent sequence of standard $d$-dimensional Gaurssian random variables. Theoretical guarantees for the (global) convergence of SGLD and its variants has been extensively studied in a nonconvex setting, see, e.g., \cite{raginsky:17, xu:18, erdo:18, nicolas:18, sabanis:20}. Moreover, it is worth noting that Langevin-based algorithms have been a key element in Bayesian statistics and in Markov Chain Monte Carlo (MCMC) theory, see, e.g., \cite{roberts:96, durmus:17, dalalyan:17, tula:19, welling:11, deng:20a, deng:20b}.

SGLD is basically a simple variant of stochastic gradient descent (SGD) in which scaled Gaussian noise is injected at each iteration. This design, however, makes it prone to the exploding and vanishing gradient problems during the training of deep learning models, thereby limiting its ability to fully leverage its theoretical advantages. This motivates us to propose a new class of stochastic gradient Langevin algorithms that addresses several challenges in deep learning. The theoretical foundation of this new class relies on recent advances in Euler-Krylov polygonal approximations for stochastic differential
equations (SDEs) with monotone coefficients, which originate from the articles \cite{Krylov:85} and \cite{Krylov:90}. We name this new class as \textit{polygonal unadjusted Langevin algorithms}, which replace $H$ in the SGLD algorithm with the Euler-Krylov polygonal approximations $H_\lambda$. Mathematically, it is described as follows: Given an i.i.d. sequence of random variables  $\{X_n\}_{n\ge0}$ of interest, which typically represent available data, the algorithm follows: for $n\in\mN_0$
\begin{equation}\label{POULA}
\theta_0^\lambda:= \theta_0, \quad \theta_{n+1}^\lambda := \theta_n^\lambda - \lambda H_{\lambda}(\theta_{n}^\lambda, X_{n+1}) + \sqrt{2\lambda \beta^{-1}}\xi_{n+1}, 
\end{equation}
where $\theta_0$ is an $\mathbb{R}^d$-valued random variable, $\lambda>0$ denotes the learning rate, and $H_{\lambda}:\mR^d \times \mR^m \rightarrow \mR^d$ satisfies the following three conditions:
\begin{enumerate}
  \item For every $\lambda>0$, there exist constants $K_\lambda>0$ and $\rho_1 \ge0$ such that
  \[|H_{\lambda}(\theta, x)| \le K_\lambda(1+|x|)^{\rho_1}(1+|\theta|) \mbox{ for every } \theta \in \mR^d \mbox{ and } x \in \mR^m.\]
  \item There exist constants $\gamma \ge 1/2$, $K_2>0$ and $\rho_2, \, \rho_3 \ge0$ such that for all $\lambda>0$,
  \[|H_{\lambda}(\theta, x) - H(\theta, x)| \le \lambda^{\gamma} K_2(1+|x|)^{\rho_2}(1+|\theta|)^{\rho_3}  \mbox{ for every } \theta \in \mR^d \mbox{ and } x \in \mR^m,\] where $H$ is the (unbiased) stochastic gradient of the objective function of the optimization problem.
  \item There exist constants $\lambda_{max}$ and $\delta\in\{1,\,2\}$ such that for any $\lambda \le \lambda_{max}$,
  \[ \liminf_{|\theta|\to\infty}\mathbb{E}\left[\langle \frac{\theta}{|\theta|^{\delta}}, H_{\lambda}(\theta, X_0) \rangle - \frac{2\lambda}{|\theta|^{\delta}}|H_{\lambda}(\theta, X_0)|^2 \right]>0. \]
\end{enumerate}

We note that it is not always guaranteed that the Euler-Krylov polygonal scheme $H_\lambda$ maintains the dissipative behavior of the original dynamical system, which is necessary for ensuring moment estimates and stability of the associated numerical algorithm. Therefore, the condition 3 is imposed for controlling high-order moments of the proposed algorithm. Typically, the case of $\delta =1$ refers to the log-Sobolev condition, consequently leading to a finite exponential moment of the algorithm. See \cite[Assumption A2]{tula:19} for the deterministic gradient case. The case of $\delta = 2$ refers to the application of Ito's formula for proving finite polynomial moments for the proposed algorithm, see proof for Lemma~\ref{lem:l2-norm} for further details.

One obtains our new algorithm TheoPouLa by considering the case where $H_{\lambda}(\theta, x)$ is the vector with entries $H_{\lambda,c}^{(i)}(\theta, x)$ as given by \eqref{eq:theo_poula}, for $i\in\{1,\ldots, d\}$. The flexibility of Euler-Krylov polygonal approximations allows TheoPouLa to combine an element-wise taming function and a boosting function, which effectively resolve the `exploding gradient' and `vanishing gradient' problems that are frequently observed in deep learning. More specifically, the element-wise taming function $1+\sqrt{\lambda}|G^{(i)}(\theta, x)|$ is proposed to control the super-linearly growing stochastic gradient\footnote{\cite{hutzenthaler:12} show that the absolute moments of the Euler discretization with super-linearly growing coefficients could diverge to infinity in finite time.} in high-dimensional optimization problems, which successfully extends the taming technique of TUSLA (\citealt{sabanis:20}). Here, $G^{(i)}$ denotes the $i$-th component of the stochastic gradient of $G(\theta, x)$ given in \eqref{eq:stochastic_gradientH}. In the literature, the taming techniques have been widely studied in the construction of stable numerical approximations. For example, see \cite{hutzenthaler:12, sabanis:13, sabanis:16} for nonlinear SDEs and \cite{tula:19, hola:19} for MCMC algorithms. Furthermore, the boosting function $1 + \frac{\sqrt{\lambda}}{\varepsilon + |G^{(i)}(\theta, x)|}$ is introduced to address the sparsity in deep learning by adaptively adjusting the stepsize of the algorithm in a region where the loss function is flat, i.e., the gradient is small. As a result, the flat gradients can increase by up to some point that is controlled by the hyperparameter of TheoPouLa, denoted by $\varepsilon$. Moreover, properly scaled isotropic Gaussian noise is added at each iteration since TheoPouLa is essentially a type of Langevin-based algorithms. Hence, its name is formed from the above description, so called Tamed Hybrid $\varepsilon$-Order POlygonal Unadjusted Langevin Algorithm (TH$\varepsilon$O POULA or TheoPouLa). We note that TheoPouLa and TUSLA (\citealt{sabanis:20}) satisfy the above three properties with $\delta=2$ and $\gamma=1/2$, whereas TULA (\citealt{tula:19}) satisfies them with $\delta= \gamma =1$ as it assumes only deterministic gradients (and thus the i.i.d. data sequence reduces to a constant).

In Section~\ref{sec:algo}, the precise formula of TheoPouLa and its full detailed analysis (including its convergence properties) are given. Furthermore, we provide in Section~\ref{sec:empirical_exp} extensive numerical experiments which demonstrate remarkable empirical performance of TheoPouLa on real-world datasets such as CIFAR10 and CIFAR100 for image classification, and the Penn Treebank for language modeling. Section~\ref{sec:sensitive} investigates the effect of the key hyperparameters of TheoPouLa on its performance and Section~\ref{sec:add_exp} presents additional experiments to support the effectiveness of the boosting function. All the proofs of main results in Section~\ref{sec:algo} are provided in Section~\ref{sec:proofs}.


\subsection{Related work: Langevin-based algorithms and adaptive learning rate methods}

In this paper, we focus on reviewing the literature studying Langevin-based algorithms for optimization problems. We refer to \cite{welling:11, ahn:12, chen:14, deng:20a, deng:20b, zhang:20} and references therein for recent progress on MCMC algorithms and Bayesian neural networks. Most research on Langevin-based algorithms for nonconvex optimization in the literature has been focused on theoretical aspects. \citet{raginsky:17} demonstrated the links between Langevin-based algorithms and stochastic optimization in neural networks, stimulating further the development and analysis of such algorithms. \citet{xu:18} analyzed the global convergence of gradient Langevin dynamics (GLD), stochastic gradient Langevin dynamics (SGLD) and stochastic variance reduced gradient Langevin dynamics (SVRG-LD). The incorporation of dependent data streams in the analysis of SGLD algorithms has been achieved in \citet{barkhagen:21} and in \citet{chau:19}, and local conditions have been studied in \citet{zhang:19b}. Recently, TUSLA in \citet{sabanis:20} has been proposed based on a new generation of tamed Euler approximations for stochastic differential equations (SDEs) with monotone coefficients in nonconvex optimization problems. Despite their elegant theoretical results, the use of Langevin-based algorithms for training deep learning models has been limited in practice as their empirical performance lacked behind in comparison to popular adaptive learning rate methods. 

Adaptive learning rate methods such as AdaGrad (\citealt{duchi:11}), RMSProp (\citealt{hinton:12}), and Adam (\citealt{kingma:15}) have been successfully applied to neural network models due to their fast training speed. In particular, Adam-type optimizers can be generally written as follows, for $n\in \mN_0$,
\begin{eqnarray}
\theta_{n+1} &=& \theta_n - \lambda \frac{m_n}{\varepsilon + \sqrt V_n} \label{eq:rule-adam}
\end{eqnarray}
where $m_n = \phi_n(H_1, \cdots, H_n)$, $V_n = \psi_n(H_1, \cdots, H_n)$ is a preconditioner, $H_i := H(\theta_i, X_i)$ is the stochastic gradient evaluated at the $i$-th iteration, $\lambda$ is the learning rate and all operations are applied element-wise. Table~\ref{tab:summary_alg} provides the details for some of popular stochastic optimization methods with corresponding averaging functions $\phi_n$ and $\psi_n$.
\begin{table*}[ht]
\renewcommand{\arraystretch}{2}
\centering
    \caption{Summary of stochastic optimization methods within the general framework. Note that $\widehat v_n = \max \{\widehat v_{n-1}, v_n\}$ is defined as $v_n = (1-\beta_2) v_{n-1} + \beta_2 H_n^2$ with $\widehat v_0 = v_0 = 0$ and all operations are applied element-wise.}
    \label{tab:summary_alg}
\begin{sc}
\scriptsize
 \begin{tabular}{c | c c c c}
 \toprule
  $\phantom0$   & SGD  & RMSProp & Adam & AMSGrad   \\
 \midrule
$\phi_n$ & $H_n$ & $H_n$ & $(1-\beta_1)\sum_{i=1}^n \beta_1^{n-i}H_i$ & $(1-\beta_1)\sum_{i=1}^n \beta_1^{n-i}H_i$  \\
$\psi_n$ & $\mathbbm{1}=(1,\ldots, 1)$ & $(1-\beta_2)\sum_{i=1}^n \beta_2^{n-i}H_i^2$  & $(1-\beta_2)\sum_{i=1}^n \beta_2^{n-i}H_i^2$ & $\max \{\widehat v_{n-1}, v_n\}$  \\
\bottomrule
\end{tabular}
\end{sc}
\end{table*}
Since the appearance of Adam, a large number of variants of Adam-type optimizers have been proposed to address the theoretical and practical challenges of Adam by suggesting a new preconditioner, $V_n$ in \eqref{eq:rule-adam}, to scale the stochastic gradient. For example, \citet{reddi:19} provided a simple example that demonstrates the non-convergence issue of Adam and proposed a simple modification, called AMSGrad, to solve the problem. \citet{chen:19} discussed the convergence of Adam-type optimizers in a nonconvex setting. RAdam to rectify the variance of adaptive learning rate has been proposed in \citet{liu:20}. \citet{wilson:17} revealed that the generalization ability of adaptive learning rate methods is worse than a global learning method like SGD. AdaBound of \citet{luo:19} attempts to overcome the drawback by employing dynamic bounds on learning rates. Recently, AdaBelief (\citealt{juntang:20}) and AdamP (\citealt{heo:21}) demonstrated their fast convergence and good generalization via extensive experiments. Nevertheless, these (and other) adaptive learning rate methods have an obvious theoretical drawback as they are only guaranteed  to converge to a stationary point, which can be a local minimum or even a saddle point in nonconvex settings. In addition, the theoretical results require strong assumptions such as the global Lipschitz continuity and boundedness conditions on the stochastic gradient. One should note that none of these two assumptions hold true in a typical optimization problem involving neural networks. This is particularly evident in complex neural network architectures.
\subsection{Our contributions}
The newly proposed algorithm, TheoPouLa, combines both advantages: global convergence in Langevin-based algorithms and powerful empirical performance in adaptive learning rate methods. To the best of the authors' knowledge, our algorithm is the first Langevin-based algorithm to achieve a comparable (or even better) empirical performance in deep learning tasks compared to popular stochastic optimization methods such as SGD, Adam, AMSGrad, RMSProp, AdaBound and AdaBelief. The major strengths of our work over related algorithms are summarized as follows:
\begin{itemize}
    \item 
    We provide a global convergence analysis of TheoPouLa in Wasserstein 1 and 2 distances when the stochastic gradient of the objective function is locally Lipscthiz continuous. Moreover, a non-asymptotic estimate for the expected excess risk of the algorithm is derived.
    \item 
    Polygonal unadjusted Langevin algorithms significantly extend possible approximations for the drift term of the Langevin SDE, which allows the algorithm to deal with the exploding and vanishing gradient problems. In particular, TheoPouLa achieves a stable and fast training process due to the element-wise taming technique and boosting function, which are theoretically well-designed for the algorithm to adaptively take a desirable stepsize. Furthermore, the effectiveness of both taming and boosting functions is confirmed through several empirical experiments.
    \item 
    While TheoPouLa behaves like adaptive learning rate methods in the early training phase, it takes an almost global learning rate near an optimal point. In other words, TheoPouLa is quickly switched from adaptive methods to SGD. As a result, it inherits the good generalization ability of SGD. Our experiments support this fact by showing that TheoPouLa outperforms the other optimization methods in \textit{generalization} measured by test accuracy for various deep learning tasks.
\end{itemize}

\section{New Algorithm: TH$\varepsilon$O POULA}\label{sec:algo}
We propose a new stochastic optimization algorithm by combining ideas from taming methods specifically designed to approximate Langevin SDEs with a hybrid approach based on recent advances of Euler-Krylov polygonal approximations. The latter is achieved by identifying a suitable boosting function (of order $\varepsilon \ll 1$) to efficiently deal with the sparsity of the stochastic gradients of neural networks. The novelty of our algorithm is to utilize a taming function and a boosting function instead of designing a new preconditioner $V_n$ in \eqref{eq:rule-adam} as in Adam-type optimizers.

We proceed with the necessary preliminary information, main assumptions and formal introduction of the new algorithm.

\subsection{Preliminaries and Assumptions}\label{sec:ass}
Let $(\Omega,\mathcal{F},P)$ be a probability space. We denote by $\mathbb{E}[X]$  the expectation of a random variable $X$.
Fix an integer $k\geq 1$. For an $\mathbb{R}^k$-valued random variable $X$, its law on $\mathcal{B}(\mathbb{R}^k)$, i.e. the Borel sigma-algebra of $\mathbb{R}^k$, is denoted by $\mathcal{L}(X)$. Scalar product is denoted by $\langle \cdot,\cdot\rangle$, with $|\cdot|$ standing for the
corresponding norm (where the dimension of the space may vary depending on the context). For $\mu\in\mathcal{P}(\mathbb{R}^k)$ and for a non-negative measurable $f:\mathbb{R}^k\to\mathbb{R}$, the notation $\mu(f):=\int_{\mathbb{R}^k} f(\theta)\mu(\rmd \theta)$ is used. For any integer $q \geq 1$, let $\mathcal{P}(\mathbb{R}^q)$ denote the set of probability measures on $\mathcal{B}(\mathbb{R}^q)$. For $\mu,\nu\in\mathcal{P}(\mathbb{R}^k)$, let $\mathcal{C}(\mu,\nu)$ denote the set of probability measures $\zeta$
on $\mathcal{B}(\mathbb{R}^{2k})$ such that its respective marginals are $\mu,\nu$. For two probability measures $\mu$ and $\nu$, the Wasserstein distance of order $p \geq 1$ is defined as
\begin{eqnarray}
{W}_p(\mu,\nu):=\inf_{\zeta\in\mathcal{C}(\mu,\nu)}
\left(\int_{\rset^k}\int_{\rset^k}|\theta-\theta'|^p\zeta(\rmd \theta \rmd \theta')\right)^{1/p} \label{eq:definition-W-p}
\end{eqnarray}
for $\mu,\nu\in\mathcal{P}(\rset^k)$. Let $(X_n)_{n\in \mN_0}$ be a sequence of i.i.d. $\mR^m$-valued random variables generating the filtration $(\mathcal G_n)_{n\in \mN_0}$ and $(\xi_n)_{n\in \mN_0}$ be an $\mR^d$-valued Gaussian process with independent components. It is assumed throughout the paper that the random variable $\theta_0$, $\mathcal G_\infty:=\sigma\left(\cup_{n\in \mN_0}\mathcal G_n\right)$, and $(\xi_n)_{n\in \mN_0}$ are independent.

Let $F:\mR^d \times \mR^m \rightarrow \mR^d$ be a continuously differentiable function such that $\mathbb{E}[|F(\theta, X_0)|]<\infty$, for all $\theta \in \mR^d$, where $X_0$ is a given $\mR^m$-valued random variable with probability law $\cL(X_0)$. We then consider the following optimization problem
\begin{eqnarray}
\min_{\theta\in \mR^d} u(\theta) = \min_{\theta\in \mR^d} \bigg(\mE[F(\theta, X_0)]  + \frac{\eta}{2(r+1)}|\theta|^{2(r+1)} \bigg) \label{eq:reg_opt}
\end{eqnarray}
where $r>0$ and $\eta \in (0,1)$. Assume that $u:\mR^d \rightarrow \mR$ is a continuously differentiable function and denote by $h:=\nabla u$ its gradient.

In the context of fine tuning of neural networks, $F$ represents the loss function for the task at hand and $\theta$ denotes the vector of the model's parameters. In particular, $r$ is determined by the property of the model.

\begin{remark}
For the reader who prefers to consider the optimization problem without the regularization term, i.e., $\eta=0$, the dissipativity condition \eqref{eq:diss} has to be additionally assumed as in the literature \citep{raginsky:17, xu:18, erdo:18}. Then, the same analysis can be applied to obtain our main results without any additional effort. However, it is yet to be proven theoretically that such an assumption holds in general for neural networks and thus it becomes a case-by-case investigation. In other words, we present here the formal theoretical statement with the appropriate regularization term which covers all of these cases.
\end{remark}

We denote by $H:\mR^d\times \mR^m \rightarrow \mR^d$ the stochastic gradient of the objective function, which is given by
\begin{equation}
H(\theta, x):= G(\theta, x) + \eta \theta |\theta|^{2r}, \label{eq:stochastic_gradientH}
\end{equation}
where $G(\theta, x) := \nabla_\theta F(\theta, x)$ for all $x\in\mR^m$, $\theta \in \mR^d$. Note that $\eta=0$ if the dissipativity condition holds for $G$. In addition, it is assumed that $H(\theta, x)$ is an unbiased estimator of $h(\theta)$ for all $\theta \in \mR^d, x\in \mR^m$.

To derive our main results, we introduce the following assumptions. Let $q\in [1, \infty)$, $r\in [q/2, \infty)$, $\rho \in [1, \infty)$ be fixed. We then impose conditions on the initial value $\theta_0$ and data process $(X_n)_{n\in \mN_0}$. As it is common to use a weight initialization technique using the uniform or normal distribution, Assumption~\ref{ass:init_ass} is mild.
\begin{assumption}\label{ass:init_ass}
The process $(X_n)_{n\in \mN_0}$ has a finite $16\rho(2r+1)$-th moment, i.e.,  $\mE[|X_0|^{16\rho(2r+1)}]<\infty$ and the initial condition has a finite $16(2r+1)$-th moment, i.e., $\mE[|\theta_0|^{16(2r+1)}]<\infty$.
\end{assumption}

The second requirement is that $G$ is locally Lipschitz continuous satisfying a polynomial growth condition, which is substantially weaker than a (globally) Lipschitz continuity or a bounded condition in the existing literature.
\begin{assumption}\label{ass:stoc_grad}
There exists a constant $
L_G>0$ such that, for all $x\in \mR^m$, $\theta, \theta' \in \mR^d$,
\begin{equation*}
|G(\theta, x) - G(\theta', x)| \leq L_G (1+|x|)^\rho (1+|\theta|+|\theta'|)^{q-1}|\theta-\theta'|.
\end{equation*}

\end{assumption}

\begin{remark}\label{rem:growth_G}
From Assumption~\ref{ass:stoc_grad}, one obtains, for all $\theta\in \mR^d$, $x\in \mR^m$,
$$
|G(\theta,x) |\leq K_G(x)(1+|\theta|)^q,
$$
where $K_G(x)= L_G(1+|x|)^\rho + |G(0,x)|$.
\end{remark}

Under Assumption~\ref{ass:init_ass} and \ref{ass:stoc_grad}, one can derive a dissipativity condition for $h$, which is presented in the next remark.

\begin{remark}\label{rem:dissipativity}
From Assumption~\ref{ass:init_ass} and ~\ref{ass:stoc_grad}, it can be shown that $h$ satisfies the following dissipativity condition, for all $\theta\in \mR^d$,
\begin{equation}
\ip{\theta}{h(\theta)} \geq A|\theta|^2 -B, \label{eq:diss}
\end{equation}
where $A=2^q \mE[1 + K_G(X_0)]$, $B=3 (2^{q+1} \mE[1+K_G(X_0)])^{q+2} /\eta^{q+1}$.
\end{remark}


\begin{remark}
We note that the regularization term $\frac{\eta}{2(r+1)}|\theta|^{2(r+1)}$ is incorporated into the optimization problem~\eqref{eq:reg_opt} to ensure the dissipativity condition given in \eqref{eq:diss}, which implies that all stationary points of the optimization problem are contained within the ball of radius $\sqrt{B/A}$ centered at the origin.  \cite{raginsky:17} introduced the dissipativity condition to establish the convergence results of SGLD in Wasserstein distances, mentioning that the dissipativity condition can be enforced through quadratic regularization. However, this justification is valid only when $G$ is globally Lipschitz continuous. In contrast, our work considers a more relaxed assumption, namely local Lipschitz continuity with a polynomial growth in Assumption~\ref{ass:stoc_grad}. Under this weaker assumption, quadratic regularization alone does not guarantee the dissipativity condition. Therefore, higher-order regularization is necessary for the convergence analysis of the optimization problem.
\end{remark}

Furthermore, under Assumption~\ref{ass:init_ass} and \ref{ass:stoc_grad}, the following proposition states that one can obtain an one-sided Lipschitz continuity condition for $h$. The proof of Proposition~\ref{prop:2.6} can be found in \cite[Proposition 1]{sabanis:20}.

\begin{proposition}\label{prop:2.6} Let Assumption~\ref{ass:init_ass} and ~\ref{ass:stoc_grad} hold. Then, one obtains, for all $\theta$, $\theta' \in \mR^d$,
$$
\ip{\theta-\theta'}{h(\theta)-h(\theta')} \geq -L_R|\theta-\theta'|^2,
$$
where $L_R= L_G\mE[(1+|X_0|)^\rho](1+2|R|)^{q-1}>0$ and $R$ is given by
$$
R = \max\bigg\{\bigg(\frac{2^{3(q-1)+1}L_G\mE[(1+|X_0|)^\rho]}{\eta}\bigg)^{\frac{1}{2r-1}}, \bigg( \frac{2^qL_G\mE[(1+|X_0|)^\rho]}{\eta}\bigg)^{\frac{1}{2r}}   \bigg\}.
$$
\end{proposition}

Under Assumption~\ref{ass:init_ass} and ~\ref{ass:stoc_grad}, $H(\theta, x)$ given in \eqref{eq:stochastic_gradientH} is locally Lipschitz continuous in $\theta$, which is explicitly stated in the following proposition. The proof follows the same idea in \cite[Proposition 2]{sabanis:20}.

\begin{proposition}\label{prop:2.7} Let Assumption~\ref{ass:init_ass} and ~\ref{ass:stoc_grad} hold. Then, one obtains that, for all $\theta\in \mR^d, x\in \mR^m$,
$$
|H(\theta, x)-H(\theta', x)| \leq L_H(1+|x|)^\rho (1+|\theta|+|\theta'|)^{2r+1} |\theta-\theta'|,
$$
where $L_H=L_G+8r\eta$.
\end{proposition}

\begin{remark}\label{rem:h_lipschitz} Let Assumption~\ref{ass:init_ass} and ~\ref{ass:stoc_grad} hold. Then, Proposition~\ref{prop:2.7} implies that $h$ is locally Lipschitz continuous. That is, there exists a $L_h>0$ such that for all $\theta\in\mR^d$,
$$
|h(\theta)- h(\theta')| \leq L_h(1+|\theta|+|\theta'|)^{2r+1}|\theta-\theta'|,
$$
where $L_h = L_H(1+ \mE[|X_0|)^{\rho}$.
\end{remark}

Remark~\ref{rem:h_lipschitz} implies that the Langevin SDE~\eqref{eq:lagenvin_sde} has a super-linear drift function. We note that the one-sided Lipschitz condition on $h$ obtained in Proposition~\ref{prop:2.6} allows for a unique solution of the Langevin SDE with the superlinear growth drift \eqref{eq:lagenvin_sde}, see \cite[Theorem~1]{Krylov:90}.


\subsection{Mechanism of TH$\varepsilon$O POULA}

We introduce the mechanism of TheoPouLa, which iterately updates as follows: for $n\in\mN_0$ and $\theta_0^\lambda :=\theta_0$,
\begin{equation} \label{Tamed-Hybrid-PoULA}
\theta_{n+1}^\lambda := \theta_n^\lambda - \lambda H_{\lambda, c}(\theta_{n}^\lambda, X_{n+1}) + \sqrt{2\lambda \beta^{-1}}\xi_{n+1},
\end{equation}
where $\lambda>0$ is the learning rate and $H_{\lambda, c} =\bigg(H_{\lambda,c}^{(1)}(\theta, x), \ldots, H_{\lambda,c}^{(d)}(\theta, x)\bigg): \mR^d \times \mR^m \rightarrow \mR^d$ is given by, for all $\theta\in \mR^d, x\in \mR^m$,
\begin{align}
H_{\lambda,c}^{(i)}(\theta, x) = \frac{G^{(i)}(\theta, x)}{\underbrace{1+ \sqrt{\lambda }| G^{(i)}(\theta, x)|}_{\mbox{taming function}}}\bigg(\underbrace{1 + \frac{\sqrt{\lambda}}{\varepsilon + |G^{(i)}(\theta, x)|}}_{\mbox{boosting function}}\bigg)
+ \underbrace{\eta\frac{\theta^{(i)}|\theta|^{2r}}{1 + \sqrt{\lambda}|\theta|^{2r}}}_{\mbox{regularization term}}, \label{eq:theo_poula}
\end{align}
for $i=1, \ldots, d$ and $0<\varepsilon<1$. In the formula of TheoPouLa \eqref{eq:theo_poula}, we call the functions $1+ \sqrt\lambda |G^{(i)}(\theta, x)|$ and $1 + \frac{\sqrt{\lambda}}{\varepsilon + |G^{(i)}(\theta, x)|}$ as the taming function and boosting function of the newly proposed algorithm, respectively.


TheoPouLa has several distinct features over the existing optimization methods in the literature. We give an intuitive explanation as to how these features are complementarily harmonized to efficiently tackle the exploding and vanishing gradient problems. We omit the regularization term, i.e., $\eta=0$, and the noise term, $\sqrt{2\lambda \beta^{-1}}\xi_{n+1}$, throughout the exposition for simplicity. Also, we refer to $\lambda$ as the \textit{learning rate} and $|\Delta \theta_n^\lambda|:=|\theta_{n+1}^\lambda - \theta_n^\lambda|=\frac{\lambda| G^{(i)}(\theta_{n}^\lambda, X_{n+1})|}{1+ \sqrt{\lambda }| G^{(i)}(\theta_{n}^\lambda, X_{n+1})|}\times
\bigg(1 + \frac{\sqrt{\lambda}}{\varepsilon + |G^{(i)}(\theta_{n}^\lambda, X_{n+1})|}\bigg)$ as the \textit{stepsize} by the convention in \citet{kingma:15}.


\begin{figure}
     \centering
    \includegraphics[width=0.6\textwidth]{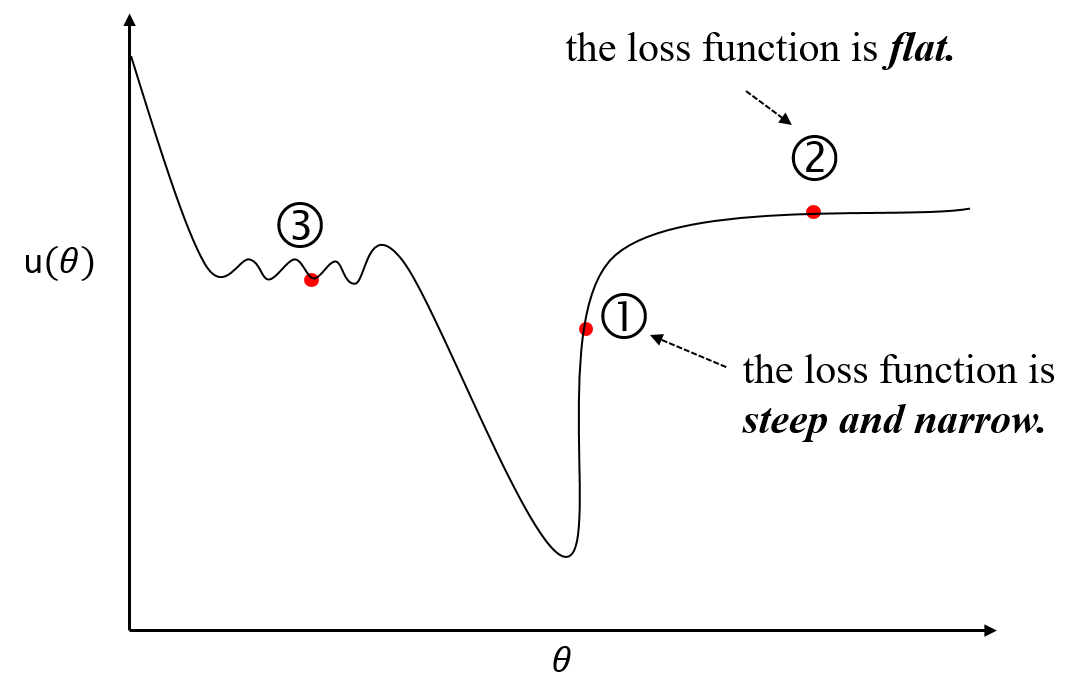}
    \caption{A landscape of a loss function.}
    \label{fig:lossfunction}
\end{figure}

Firstly, the new algorithm adopts the taming function to control the super-linearly growing gradient. In region \textcircled{\small{1}} of Figure~\ref{fig:lossfunction} where the loss function is steep and narrow, i.e., the gradient is huge, it is ideal for the optimizer to take a small stepsize. This is effectively achieved because the growth of the taming function is proportional to $G$, which relieves the huge gradient. On the other hand, the boosting is close to one, i.e. it becomes useless in this case. The effectiveness of the taming function is confirmed in the motivating example in Section~\ref{sec:sim_example}. In particular, we emphasize that the taming function of TheoPouLa is applied element-wise to scale the effective element-wise learning rate in contrast to TUSLA of \citet{sabanis:20}. This significantly improves the performance of TheoPoula when solving high-dimensional optimization problems such as the training of neural network models.

Secondly, we first introduce the boosting function to accelerate training speed and prevent the vanishing gradient problem. When the current parameter is located in region \textcircled{\small{2}} of Figure~\ref{fig:lossfunction} where the loss function is (almost) flat, i.e. the gradient is small, the boosting function helps TheoPoula to adaptively increase its stepsize. Specifically, the boosting function can increases the stepsize by up to $\sqrt{\lambda}/\varepsilon$,  whereas the taming function barely contributes to the stepsize. We highlight that the taming and boosting functions do not interfere with each other in any adverse way. On the contrary, they complement each other in a harmonious way.  We verify the effectiveness of the boosting function, which can be found in Section~\ref{app:boosting}. The experiments show that the boosting function brings a significant improvement in test accuracy across different models and data sets for deep learning models.

Thirdly, TheoPouLa is quickly converted from adaptive learning rate methods to SGD. In the early training phase, TheoPouLa certainly behaves like adaptive learning rate methods. Then, when the current position is approaching an optimal solution, TheoPouLa is similar to SGD with a learning rate $(1 + \sqrt{\lambda}/\varepsilon)$. Consequently, TheoPouLa simultaneously attains two favorable features of fast training in adaptive learning rate methods and good generalization in SGD. The switching from adaptive learning rates to SGD has been also investigated by different strategies in \citet{luo:19} and \citet{keskar:17}.

Lastly, a scaled Gaussian noise, $\sqrt{2\lambda \beta^{-1}}\xi_{n+1}$, is added as a consequence of the discretization of the Langevin SDE. The term is essential to prove the convergence property of TheoPouLa. Moreover, adding properly scaled Gaussian noise allows the new algorithm to escape from region \textcircled{\small{3}} of Figure~\ref{fig:lossfunction}, local minima or saddle points, in a similar manner to the standard SGLD method, see \citet{raginsky:17}. 

\begin{remark}   To address the comparison between our proposed algorithm and ADAM-type optimizers, we provide a detailed discussion on the mechanism of ADAM under the same loss function landscape depicted in Figure~\ref{fig:lossfunction}. Both ADAM-type optimizers and TheoPouLa exhibit similarities in their approach to handling steep and flat regions of the loss landscape. In steep regions, they both take smaller step sizes, while in flat regions, they increase the step size adaptively. This strategy effectively mitigates the problems of exploding and vanishing gradients.

However, the underlying ideas to achieve the properties are markedly different. For ADAM-type optimizers, the key lies in the preconditioner\footnote{For ADAM, $V_n$ is defined as the exponential moving average of squared gradients.} $V_n$ defined in \eqref{eq:rule-adam} . In steep regions like \textcircled{\small{1}} of Figure~\ref{fig:lossfunction}, $V_n$ becomes large, reducing the step size, whereas in flat areas \textcircled{\small{2}} of Figure~\ref{fig:lossfunction} , a smaller $V_n$ amplifies the step size. Unlike the harmonious interaction between the boosting function and the taming function in TheoPouLa, the preconditioner in ADAM-type optimizers works in a complex and sometimes adversarial manner with the momentum term $m_n$. In Subsection~\ref{sec:sim_example}, we demonstrate that an excessively large preconditioner can lead to the failure of ADAM's convergence in the case of superlinear gradients. We also refer to \cite{reddi:19} for fundamental flaw in the preconditioner of ADAM. 

Furthermore, the update rule of ADAM does not contain scaled noise, relying solely on momentum to escape local minima or saddle points. This reliance becomes particularly problematic in scenarios depicted in \textcircled{\small{3}} of Figure~\ref{fig:lossfunction}, where small local minima are clustered together, making it challenging for momentum alone to facilitate an effective escape.

\end{remark}

\subsection{Convergence Analysis}

In this subsection, we present the main convergence results of TheoPouLa in Wasserstein-1 and Wasserstein-2 distances which are defined in \eqref{eq:definition-W-p}. The convergence is guaranteed when  $\lambda$ is less than $\lambda_{\max}$, which is given by
\begin{equation}\label{eq:lambda_max}
\lambda_{\max} = \min\bigg\{1, \frac{1}{4\eta^2},\frac{1}{2^{14}\eta^2 ({}_{8l}\cC_{4l})^2 }\bigg\}.
\end{equation}
where  ${}_n\cC_k$ is the binomial coefficient `$n$ choose $k$' and $l=2r+1$.
Note that the learning rate restriction causes no issues as $\eta$ is typically very small ($\eta \ll 1$).   

Theorem~\ref{Thrm1} and Corollary~\ref{Thrm2} state the non-asymptotic estimates for the Wasserstein-1 and -2 distances between $\mathcal{L}\left(\theta_{n}^{\lambda}\right)$ and $\pi_\beta$. The proofs of the main results can be found in Section~\ref{sec:proofs}.

\begin{theorem}\label{Thrm1}
Let Assumption \ref{ass:init_ass} and \ref{ass:stoc_grad} hold.  Then, for all $0<\lambda\leq \lambda_{\max}$, $n\in \mN_0$, we have that
\begin{align*}
W_{1}\left(\mathcal{L}\left(\theta_{n}^{\lambda}\right), \pi_{\beta}\right) \leq & C_1 \sqrt \lambda + C_2e^{- C_0\lambda n } ,
\end{align*}
where $C_0$, $C_1$ and $C_2$ are explicitly given in Table~\ref{tab:constant}. Moreover, the constants $C_0$, $C_1$, $C_2$ are independent of $n$ and $\lambda$.
\end{theorem}

\begin{corollary}\label{Thrm2}
Let Assumption \ref{ass:init_ass} and \ref{ass:stoc_grad} hold.  Then, for all $0<\lambda \leq \lambda_{max}$, $n\in \mN_0$, we have
\begin{align*}
W_{2}\left(\mathcal{L}\left(\theta_{n}^{\lambda}\right), \pi_{\beta}\right)  \leq & C_3 \lambda^{\frac{1}{4}} +C_4 e^{-C_5\lambda n},
\end{align*}
where $C_3$, $C_4$ and $C_5$ are explicitly given in Table~\ref{tab:constant}. Moreover, the constants $C_3$, $C_4$, $C_5$ are independent of $n$ and $\lambda$.
\end{corollary}

We are now concerned with the expected excess risk of TheoPouLa, so called the optimization error of $\theta_n^\lambda$, which is defined as
\begin{equation}\label{eq:excess_risk}
\mE[u(\theta_n^\lambda)] - u(\theta^*),
\end{equation}
where $\theta^* := \arg\min_{\theta \in \mR^d} u(\theta)$. Using the result in Corollary~\ref{Thrm2}, one can further obtain an error bound of the expected excess risk as stated in the below.

\begin{theorem}\label{Thrm3}
Let Assumption~\ref{ass:init_ass} and ~\ref{ass:stoc_grad} hold. For any $n\in \mN_0$, the expected excess risk of the $n$-th iterate of TheoPouLa is bounded by
\begin{align*}
\mE[u(\theta_n^\lambda)] - u(\theta^*) &\leq C_6 W_2(\mathcal L(\theta_n^\lambda, \pi_\beta)) +  \frac{C_7}{\beta},
\end{align*}
where $W_2(\mathcal L(\theta_n^\lambda), \pi_\beta)$ is given in Corollary~\ref{Thrm2} and constants $C_6$, $C_7$ are explicitly given in Table~\ref{tab:constant}. Moreover, the constants $C_6$, $C_7$ are independent of $n$ and $\lambda$.
\end{theorem}

\begin{remark}
The constants $C_0$, $C_1$, $C_2$, $C_3$, $C_4$, $C_6$, $C_7$ are independent of $n$ and $\lambda$, but might depend on $\beta$ and $d$. Their full expressions and dependency on $d$ and $\beta$ can be found in Table~\ref{tab:constant} and ~\ref{tab:basic constants}. In particular, the constants have exponential dependence on the dimension $d$ because our nonconvex setting should encompass possible pathological scenarios. In particular, the exponential dependence on $d$ only comes from the contraction property of the Langevin SDE in Lemma~\ref{eberle}, inherited from the result in \cite{eberle:19}. In other words, if the contraction estimate can be improved under reasonable regularities, the exponential dependence on $d$ is accordingly relaxed without affecting our analysis.
\end{remark}

Using Corollary~\ref{Thrm2}, the expected excess risk of TheoPoula in Theorem~\ref{Thrm3} is rewritten as
\begin{equation*}
\mE[u(\theta_n^\lambda)] - u(\theta^*) \leq C_3C_6 \lambda^{\frac{1}{4}} + C_4C_6 e^{-C_5 \lambda n}+  \frac{C_7}{\beta}. \label{eq:thm3}
\end{equation*}
Then, the error bound of the expected excess risk in Theorem~\ref{Thrm3} can be interpreted via the following three steps: (i) For any $\delta>0$, choose $\bar \beta >0$ such that
$$
\frac{C_7}{\bar \beta} \leq \frac{\delta}{3},
$$
and fix $\bar \beta$,
(ii) Then, pick and fix $\bar \lambda>0$ such that
$$
C_{3}C_{6} \bar \lambda^{\frac{1}{4}} \leq \frac{\delta}{3},
$$
by using that $C_3$, $C_6$ are independent of $\lambda$, (iii) Lastly, choose $\bar n>0$ such that
$$
C_{4}C_{6} e^{-C_{5} \bar \lambda \bar n}\leq \frac{\delta}{3}.
$$
Therefore, for any $\delta>0$, one can always find $(\bar \lambda, \bar n, \bar \beta)$ that achieves the expected excess risk of TheoPoula being less than $\delta$.

\section{Comparison with Other Langevin-based algorithms}

This section discusses the contributions of this work with an emphasis on assumptions, convergence results, and the algorithm structure. In particular, we provide a comprehensive point-by-point comparison of our work with recent progress on SGLD and its variants in the literature, such as \cite{raginsky:17}, \cite{xu:18}, \cite{zhang:19b}, \cite{sabanis:20}, and \cite{zou:21}.

In the seminal work of \cite{raginsky:17}, the authors establish non-asymptotic convergence results in Wasserstein-2 distance for SGLD, where the convergence rate is $\lambda^{5/4} n$ and depends on the number of iterations $n$. On the other hand, in Theorem~\ref{Thrm1} and Corollary~\ref{Thrm2}, we obtain the \textit{best known} convergence rates $\lambda^{1/2}$ and $\lambda^{1/4}$ in Wasserstein-1 and Wasserstein-2 distances, respectively, which are independent of $n$. In addition, Assumptions in \cite{raginsky:17} are stronger than ours. First, it assumes the (global) Lipschitz continuity on the stochastic gradient $H$ in $\theta$ \cite[Assumption (A.2)]{raginsky:17}, which is significantly stronger than our local Lipschitz continuity condition with a polynomial growth as given in Assumption~\ref{ass:stoc_grad}. Second, an estimate for the variance of the stochastic gradient $H$ is required \cite[Assumption (A.4)]{raginsky:17}, whereas our theoretical results can be derived without this quantity. Third, while it assumes the finiteness of an exponential moment of the initial value $\theta_0$ in \cite[Assumption (A.5)]{raginsky:17}, we only assume polynomial moments on the initial value $\theta_0$ in Assumption~\ref{ass:init_ass}.  

Although \cite{xu:18} achieves improved results for SGLD through a direct analysis of the ergodicity of numerical approximations to Langevin dynamics, its convergence rate remains dependent on $n$. Furthermore, the authors assume a global Lipschitz condition on the gradient.

In \cite{zhang:19b}, the authors replace the global Lipschitz continuity condition of the stochastic gradient with a \textit{local} Lipschitz continuity condition, accounting for the dependence on the data $X$, and also derive non-asymptotic convergence results with rates $1/2$ and $1/4$ for SGLD in Wasserstein-1 and Wasserstein-2 distances, respectively. While the local Lipschitz condition in \cite[Assumption 2]{zhang:19b} removes the uniform dependence on the data, the stochastic gradient remains globally Lipschitz continuous in $\theta$. This condition is not only inadequate for optimization problems involving neural networks, but also represents a stronger assumption than Assumption~\ref{ass:stoc_grad} in our paper. 

\cite{sabanis:20} proposes a variant of SGLD, known as the tamed unadjusted stochastic Langevin algorithm (TUSLA), which employs taming techniques to enhance the stability of the algorithm under the assumption of locally Lipschitz continuous stochastic gradients. Despite its theoretical motivation to address the exploding gradient problem in deep learning, the taming factor in TUSLA, $1+\sqrt\lambda |\theta|^{2r}$, proportional to the $2r$-moment of the norm of the model's parameters, aggravates another serious issue: the vanishing gradient problem. As a result, as demonstrated in our numerical experiments (see Tables~\ref{tab:cifar10} and \ref{tab:lstm}), TUSLA shows worse empirical performance than SGLD. Moreover, as the structure of TheoPouLa employs \textit{element-wise} taming and boosting functions, which significantly differ from SGLD and TUSLA, additional careful steps are required to obtain estimates for the Wasserstein distances between the proposed algorithm and the target distribution $\pi_\beta$. 

Lastly, we comment on the study by \cite{zou:21}, which focuses on convergence results in total variation for SGLD. The authors derive non-asymptotic convergence estimates in total variation with a precise dependence of the constants on key parameters. To obtain these convergence estimates, the authors assume stronger assumptions, such as the global Lipschitz continuity of the gradient. In addition, it is worth noting that the Wasserstein distance is often considered as a more suitable metric than total variation for quantifying the quality of approximate solutions, as it provides better intuition on the closeness of two distributions. We refer to \cite{dalalyan:17b} and \cite{bortoli:22} for further discussions.

\section{Numerical Experiments}\label{sec:exp}
This section provides extensive numerical experiments to demonstrate the empirical performance of TheoPouLa. In Section~\ref{sec:sim_example}, we present a simple example to illustrate that popular stochastic optimization algorithms may fail to find the optimal solution in the presence of the super-linearly growing stochastic gradient. In Section~\ref{sec:empirical_exp}, we present two real-world deep learning tasks such as image classification on CIFAR10 (\citealt{cifar10}) and CIFAR-100 (\citealt{cifar100}), and language modeling on Penn Treebank  (\citealt{treebank}). In Section~\ref{sec:sensitive}, we investigate the effect of key hyperparameters $\lambda$, $\varepsilon$,  $\beta$ on the performance of TheoPouLa and the effectiveness of the boosting function.

\subsection{Toy example}\label{sec:sim_example}
The super-linearly growing gradient and its effect on the performance of optimization methods are relatively under-studied because most relevant studies assume that the stochastic gradient is global Lipschitz continuous and bounded (\citealt{kingma:15, xu:18, nicolas:18, duchi:11, hinton:12, reddi:19, chen:19, liu:20, luo:19, juntang:20}). However, the assumptions are not true for the problem of training neural networks. This section provides a simple one-dimensional optimization problem that illustrates the convergence issue of popular optimization algorithms when the stochastic gradient is locally Lipschitz continuous that results in the super-linearly growing gradient\footnote{A function $f:\mR^k \rightarrow \mR^j$ for $k,j\in \mN$ is said to be super-linearly growing if $\sup_{\theta \in \mR^k} \frac{|f(\theta)|}{1+|\theta|}=\infty$.}. \citet{sabanis:20} considers a similar example to show the stability of TUSLA using a different taming function.

Consider the following optimization problem:
\begin{equation}
\min_{\theta \in \mR} u(\theta) = \min_{\theta \in \mR} \mathbb E [U(\theta, X)],\label{eq:optim_sim}
\end{equation}
where $U: \mR \times \mR \rightarrow \mR$ is defined as
$$
U(\theta, x) = \left\{ \begin{array}{ll}
\theta^2 \left(1+ \1_{x\leq 1} \right) + \theta^{30} \quad \text{if }|\theta|\leq 1, \\
(2 |\theta| - 1)\left(1+ \1_{x\leq 1} \right)+ \theta^{30} \quad \text{if } |\theta|>1,
\end{array}\right.
$$
and $X$ is uniformly distributed over $(-2, 2)$, that is, $f_X(x) = \frac{1}{4} \1_{|x|\leq 2}$. Furthermore, the stochastic gradient $H:\mR \times \mR \rightarrow \mR$ is given by
$$
H(\theta,x) = \left\{ \begin{array}{ll}
2\theta\left(1+ \1_{x\leq 1} \right) + 30\theta^{29} \quad \text{if } |\theta|\leq 1, \\
2(1 +  \1_{x\leq 1})sgn(\theta) +  30\theta^{29}   \quad \text{if } |\theta|>1,
\end{array}\right.
$$
where $sgn(\cdot)$ is the sign function. Then, one can easily show that the stochastic gradient $H$ is locally Lipschitz continuous:
$$
|H(\theta, x) - H(\theta', x) | \leq 34 (1 + |\theta|+ |\theta'|)^{28} |\theta -\theta'|,
$$
for all $x \in \mR$ and $\theta, \theta' \in \mR$. Moreover, the minimum value of the optimization problem~\eqref{eq:optim_sim} is attained at $\theta=0$.

\begin{figure*}[ht]
     \centering
     \begin{subfigure}[b]{0.45\textwidth}
         \centering
         \includegraphics[width=\textwidth]{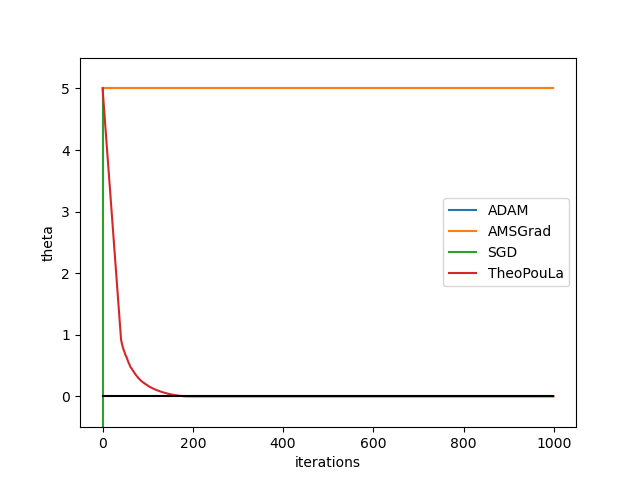}
         \caption{default settings}
         \label{fig:sim_exam_uniform_a}
     \end{subfigure}
     \begin{subfigure}[b]{0.45\textwidth}
         \centering
         \includegraphics[width=\textwidth]{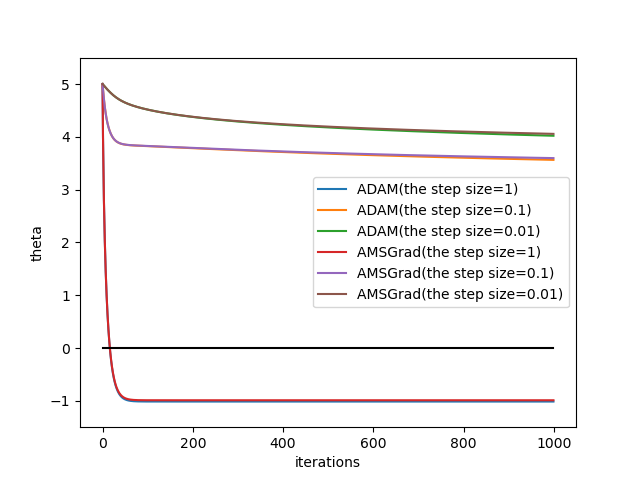}
         \caption{different step sizes}
         \label{fig:sim_exam_unifrom_b}
     \end{subfigure}
     \caption{Performance of SGD, Adam, AMSGrad and TheoPouLa on an artificial example with the initial value $\theta_0 = 5.0$}
    \label{fig:sim_exam_uniform}
\end{figure*}

We examine the behavior of SGD, Adam, and AMSGrad in solving the optimization problem~\eqref{eq:optim_sim} with the initial value $\theta_0 = 5$. For hyperparameters of the optimization algorithms, we use their default settings provided in PyTorch. More specifically, for Adam and AMSGrad, $\lambda=0.001$, $\beta_1= 0.9$ and $\beta_2=0.999$ are used. Figure~\ref{fig:sim_exam_uniform}(a) presents the trajectories of approximate solutions generated by the optimization algorithms, showing that SGD, Adam, and AMSGrad fail to converge to the optimal solution $0$ even after $1,000$ iterations.

Intuitively, the undesirable phenomenon of Adam-type optimizers occurs because, in the update rule \eqref{eq:rule-adam}, the denominator $\sqrt V_n$, so-called the preconditioner, excessively dominates the numerator $m_n$, causing the vanishing gradient problem in the presence of the super-linealy growing gradient. On the other hand, SGD suffers from the exploding gradient problem due to the huge gradient. Moreover, Figure~\ref{fig:sim_exam_uniform}(b) highlights that the problematic behavior cannot be simply resolved by adjusting the learning rate within the Adam-type framework. On the contrary, TheoPouLa rapidly finds the optimal solution only after 200 iterations due to its taming function that controls the super-linearly growing gradient.


\subsection{Empirical performance on real data sets}\label{sec:empirical_exp}
We compare the performance of TheoPouLa with that of other popular optimization algorithms including Adam (\citealt{kingma:15}), AdaBelief (\citealt{juntang:20}), AdamP (\citealt{heo:21}), AdaBound (\citealt{luo:19}), AMSGrad (\citealt{reddi:19}), RMSProp (\citealt{hinton:12}), SWATS (\citealt{keskar:17}), SGD (with momentum), ASGD (\citealt{merity:18}), SGLD (\citealt{raginsky:17}), and TUSLA (\citealt{sabanis:20}). In particular, we consider three deep convolution neural networks for image classification: VGG11 (\citealt{simon:14}), ResNet34 (\citealt{he:16}) and DenseNet121 (\citealt{huang:17}) models. For language modeling, AWD LSTMs with 1, 2 and 3 layers are considered. Each experiment is run three times to compute the mean and standard deviation of the best accuracy on the test dataset.

Recall that it is assumed that $r \geq \frac{q}{2} \geq \frac{1}{2}$ to obtain the main results. However, for the experiments in this section, we consider $r=0$ in \eqref{eq:reg_opt}, which is equivalent to $\ell_2$-regularization. This is justified by the fact that some form of dissipativity may already exist for specific problems such as the one considered here, although this has not been verified theoretical so far. In Section~\ref{app:nonzero_eta}, we perform additional experiments with $r \geq \frac{1}{2}$, which show similar performance of TH$\varepsilon$O POULA as in Table~\ref{tab:cifar10} without any noticeable loss of accuracy. This demonstrates that there is no gap between theory and practice of our work.



\paragraph{Image classification.}
We replicate the experiments for image classification based on the official implementation of \cite{juntang:20} as it provides a reliable baseline of the experiments by comparing the performance of various optimization algorithms with extensive hyperparameter search. More specifically, all the models are trained for 500 epochs with batch size of $128$. We set $\eta=0.0005$ and $r=0$ in \eqref{eq:reg_opt}, which is $\ell_2$-regularization. We decay the initial learning rate by 10 after 150 epochs to all optimization algorithms.

For TheoPouLa, we search the optimal hyperparameters as follows: $\lambda \in \{1, 0.5, 0.1, 0.05, \\ 0.01\}$, $\varepsilon \in \{1, 0.1, 0.01\}$ and $\beta \in \{10^8, 10^{10}, 10^{12}\}$. For SGLD and TUSLA, we use the following hyperparameters: $\lambda=\{1, 0.5, 0.1, 0.05, 0.01\}$, $r=0.5$, and $\beta=\{10^8, 10^{10}, 10^{12}\}$.  Regarding hyperparameter values of Adam, AdaBelief, AdamP, AdaBound, AMSGrad and RMSProp, the best hyperparameters are chosen among $\lambda \in \{1.0, 0.1, 0.01, 0.001\}$, $\beta_1 \in \{0.5, 0.6, 0.7, 0.8, 0.9\}$, $\beta_2=0.999$, and $\epsilon=10^{-8}$. For SGD, we set the momentum as $0.9$ and search learning rate $\lambda \in \{10.0, 1.0, 0.1, 0.01, 0.001\}$\footnote{Our results for Adam, AdaBelief, AdaBound, AMSGrad, and RMSprop are consistent with the test accuracies reported in \cite{luo:19} and \cite{juntang:20}.}.

\begin{figure}[htbp]
    \centering
    \begin{subfigure}[b]{0.325\textwidth}
        \includegraphics[width=\textwidth]{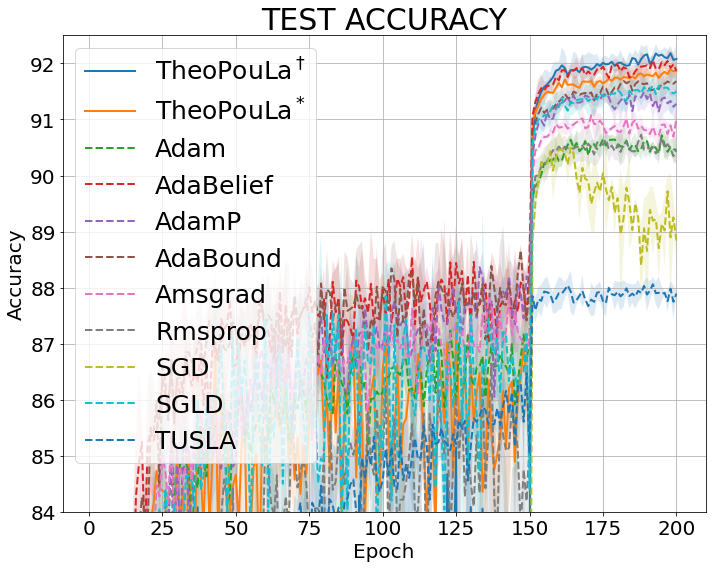}
        \caption{VGG11 on CIFAR10}        
    \end{subfigure}
    \begin{subfigure}[b]{0.325\textwidth}
        \includegraphics[width=\textwidth]{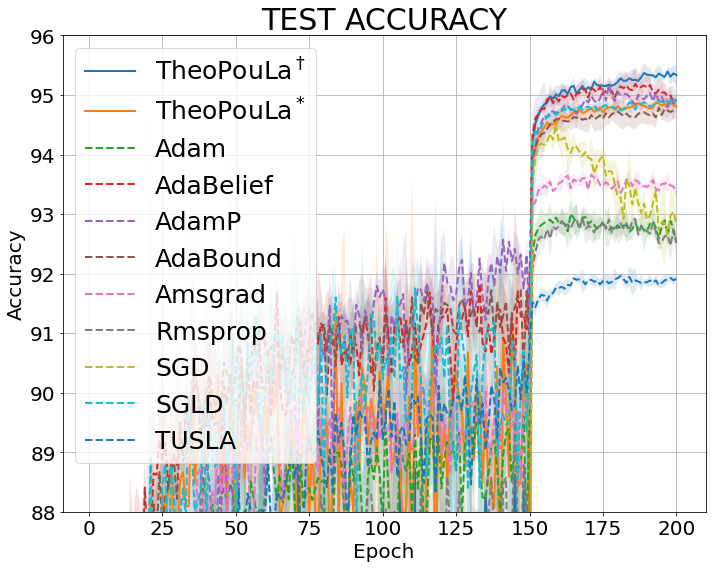}
        \caption{ResNet34 on CIFAR10}        
    \end{subfigure}
    \begin{subfigure}[b]{0.325\textwidth}
        \includegraphics[width=\textwidth]{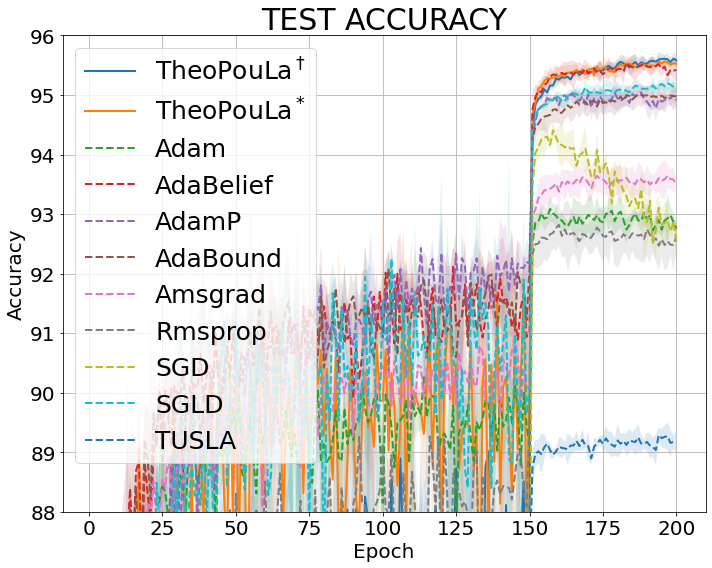}
        \caption{DenseNet121 on CIFAR10}        
    \end{subfigure}
        \begin{subfigure}[b]{0.325\textwidth}
        \includegraphics[width=\textwidth]{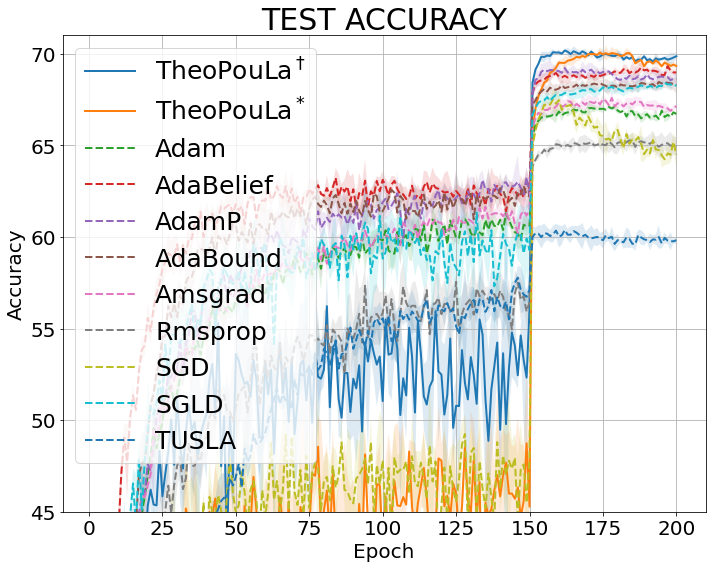}
        \caption{VGG11 on CIFAR100}        
    \end{subfigure}
    \begin{subfigure}[b]{0.325\textwidth}
        \includegraphics[width=\textwidth]{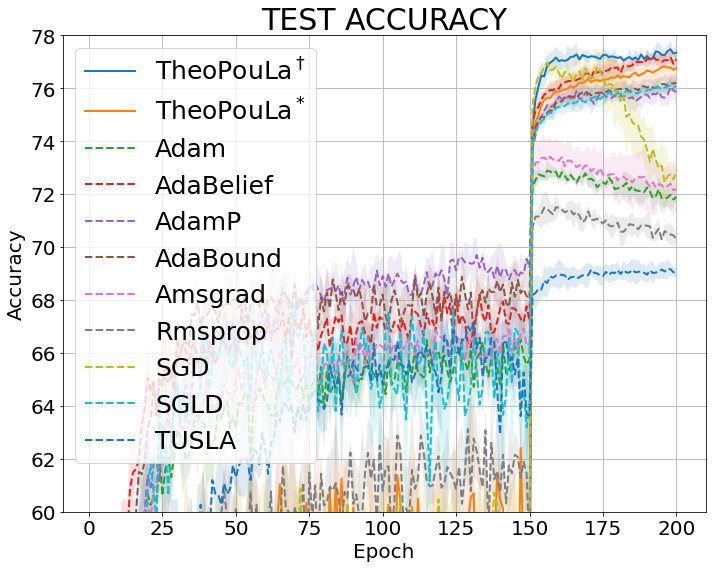}
        \caption{ResNet34 on CIFAR100}        
    \end{subfigure}
    \begin{subfigure}[b]{0.325\textwidth}
        \includegraphics[width=\textwidth]{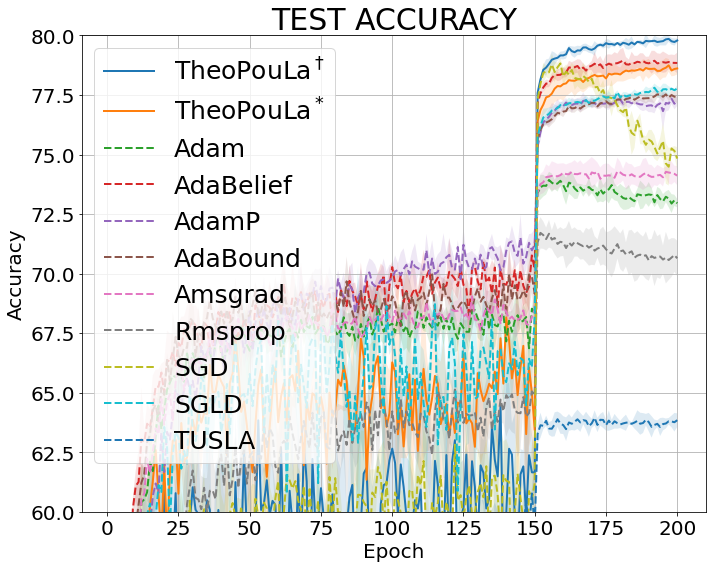}
        \caption{DenseNet121 on CIFAR100}
    \end{subfigure}
    \caption{Test accuracy for VGG11, ResNet34 and DenseNet121 on CIFAR-10 and CIFAR-100. TheoPouLa$^\dagger$ and TH$\varepsilon$OPOULA$^*$ represent the performances of TheoPouLa under the best and second best hyperparameters, respectively.}
    \label{fig:test_acc_cifar}
\end{figure}

Figure~\ref{fig:test_acc_cifar} shows test accuracy for VGG11, ResNet34 and DenseNet121 on CIFAR10 and CIFAR100. Table~\ref{tab:cifar10} shows the test accuracy for VGG11, ResNet34 and DenseNet121 on CIFAR10 and CIFAR100. As shown in Table~\ref{tab:cifar10}, our algorithm achieves the highest accuracy and significantly outperforms the other optimization algorithms across all the experiments. In particular, TheoPouLa with the second best hyperparameter is even comparable to the AdaBelief (the state-of-the-art algorithm) and outperforms the other methods, validating that the solutions found by TheoPouLa yield good generalization performance. Also, the improvement of our algorithm is increasingly prominent as the models and datasets are more complicated and large-scale.

\begin{table*}[ht]
\footnotesize
\centering
\caption{Mean and standard deviation of the best accuracy for VGG11, ResNet34 and DenseNet121 on CIFAR10. TheoPouLa$^\dagger$ and TheoPouLa$^*$ represent the performances of TheoPouLa with the best and second best hyperparameters, respectively. The numbers in parenthesises indicate the standard deviations.}
\begin{tabular}{c | c | c | c | c | c | c}
 \hline
dataset & \multicolumn{3}{c|}{CIFAR-10} & \multicolumn{3}{c}{CIFAR-100} \\\hline
model & VGG & ResNet & DenseNet & VGG & ResNet & DenseNet \\
 \hline\hline
\multirow{2}{*}{TheoPouLa$^\dagger$}  & \textbf{92.30} & \textbf{95.43} & \textbf{95.66} &  \textbf{70.31} & \textbf{77.60} & \textbf{79.90}\\
&(0.055) &(0.095) & (0.066)  & (0.117)  & (0.208)& (0.133)\\
 \hline
\multirow{2}{*}{TheoPouLa$^*$} & 91.92  & 94.92 & 95.59 & 70.24 & 76.88 & 78.76\\
&(0.119) & (0.076)& (0.067)& (0.227) & (0.536) & (0.269)\\
 \hline
AdaBelief & 92.17 & 95.29 & 95.58 & 69.50 &   77.33 & 79.12\\
 (baseline)& (0.035)&(0.196) & (0.095)& (0.111)  &   (0.172) & (0.382)\\
 \hline
\multirow{2}{*}{Adam} & 90.79 & 93.11 & 93.21 & 67.30 &   73.02 & 74.03\\
 & (0.075)&(0.184) & (0.240)& (0.137)  &   (0.231) & (0.334)\\
 \hline
 \multirow{2}{*}{AdamP} & 91.68 & 95.18 & 95.17 & 69.41 &  76.14 & 77.58\\
 & (0.162) &(0.116) &(0.079) & (0.297)  &   (0.347) & (0.091)\\
 \hline
 \multirow{2}{*}{AdaBound} & 91.81 & 94.83 & 95.05& 68.61 &  76.27 & 77.56\\
 & (0.272) &(0.131) &(0.176) & (0.312)  &   (0.256) & (0.120)\\
 \hline
\multirow{2}{*}{AMSGrad} & 91.24 & 93.76 & 93.74& 67.71    &   73.51 & 74.50 \\
        & (0.115) & (0.108)& (0.236)& (0.291)  & (0.692)  & (0.416)\\
 \hline
\multirow{2}{*}{RMSProp} & 90.82 & 93.06 & 92.89& 65.45 &  71.79 & 71.75\\
& (0.201) &(0.120) &(0.310) & (0.394) &  (0.287) & (0.632)\\
 \hline
 \multirow{2}{*}{SGD} & 90.73 & 94.61 & 94.46 & 67.78  &   77.16 & 78.95\\
 &(0.090) & (0.280)&(0.159) & (0.320)  &  (0.214) & (0.312)\\
 \hline
  \multirow{2}{*}{SWATS} & 87.29 & 94.76 & 95.04 & \multirow{2}{*}{N/A}  &  73.86 & 78.81\\
 &(4.210) &(0.565) &(0.339) &   &  (3.928) & (1.812)\\ \hline
 \multirow{2}{*}{SGLD} & 91.68 &  94.97 &  95.26 & 69.52  &  76.2 & 77.88\\
 &  (0.210) &  (0.023) &  (0.080) &   (0.100)  &   (0.185) &  (0.135)\\ \hline
 \multirow{2}{*}{{ TUSLA}} &  88.23 &  92.05 &   89.37 &  60.55  &   69.25 &  64.07\\
 & (0.060) &  (0.060) &  (0.135) &   (0.209) &   (0.304) &  (0.299)\\
 \hline
\end{tabular}
\label{tab:cifar10}
\end{table*}

\paragraph{Language modeling.}
We conduct language modeling over the Penn Treebank (PTB) with AWD-LSTM models of \citet{merity:18}. It is reported that Non-monotonically Triggered ASGD (NT-ASGD) achieves state-of-the-art performance for this task. Motivated by this observation, we also consider averaged TheoPouLa, which is performed by averaging of trajectories of the parameters after a user-specified trigger $Q$, $\frac{1}{n-Q+1} \sum_{i=Q}^n \theta_i^\lambda$, instead of the last updated parameter $\theta_n^\lambda$ (\citealt{polyak:92}). Moreover, we use a trigger strategy which starts the averaging when no improvement in the validation metric is seen for a patience number of epochs. For our experiments, we set the patience number to $5$. We only test NT-ASGD, AdaBelief, SGLD, and TheoPouLa rather than investigating the various optimization algorithms used in the image classification. Since AdaBelief significantly outperforms the other optimization algorithms including vanilla SGD, AdaBound, Yogi (\citealt{zaheer:18}), Adam, MSVAG (\citealt{balles:18}), RAdam, Fromage and AdamW (\citealt{losh:19}) in the same experiment, we believe that it is enough to compare the performance of AdaBelief, NT-ASGD, SGLD, and TheoPouLa. In addition, we observed that the performance of TUSLA in this task is poor compared to  the algorithms presented in Table~\ref{tab:lstm}.

For a fair comparison, the averaging scheme has also been applied to AdaBelief. However, it turns out that the averaging scheme does not improve the performance of AdaBelief. Instead, AdaBelief uses a development-based learning rate decay, which decreases the learning rate by a constant factor $\delta$ if the model does not attain a new best value for $k$ epochs. We search the optimal learning rate schedule among $\delta \in \{0.1, 0.5\}$ and $k \in \{5, 10, 20\}$. For ASGD and TheoPouLa, a constant learning rate is used without a learning rate decay. Moreover, in order to compare with the baseline, we apply gradient clipping of $0.25$ to all three optimization algorithms.

We train the AWD LSTM with 1,2 and 3 layers for $750$ epochs with $20$ batch size. The details of models can be found in the official implementation of AWD-LSTM \footnote{https://github.com/salesforce/awd-lstm-lm}. For NT-ASGD and averaged TheoPouLa, the constant learning rate of $30$ is used for 2 and 3-layer LSTMs. For 1-layer LSTMs, we set $\lambda$ to $10$. Moreover, we fix the following hyperparameters: $\varepsilon=100$ and $\beta=10^{10}$  across all the experiments. For AdaBelief, we used the best hyperparameters reported in \citet{juntang:20}. That is, we use $\lambda=0.01$ and $\epsilon=10^{-12}$ for 2 and 3-layer LSTMs, and $\lambda=0.001$ and $\epsilon=10^{-16}$ for 1-layer LSTMs where $\beta_1=0.9$ and $\beta_2=0.999$ are fixed.

\begin{figure}[htbp]
    \centering
    \begin{subfigure}[b]{0.325\textwidth}
        \includegraphics[width=\textwidth]{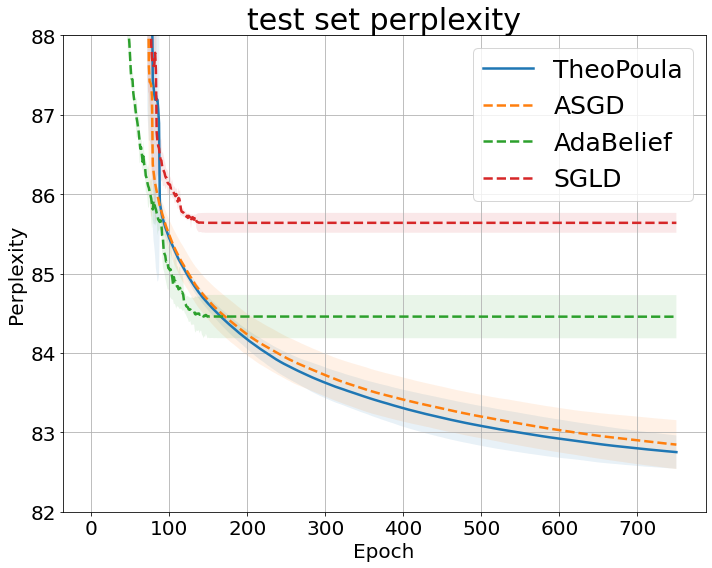}
        \caption{1-layer}        
    \end{subfigure}
    \begin{subfigure}[b]{0.325\textwidth}
        \includegraphics[width=\textwidth]{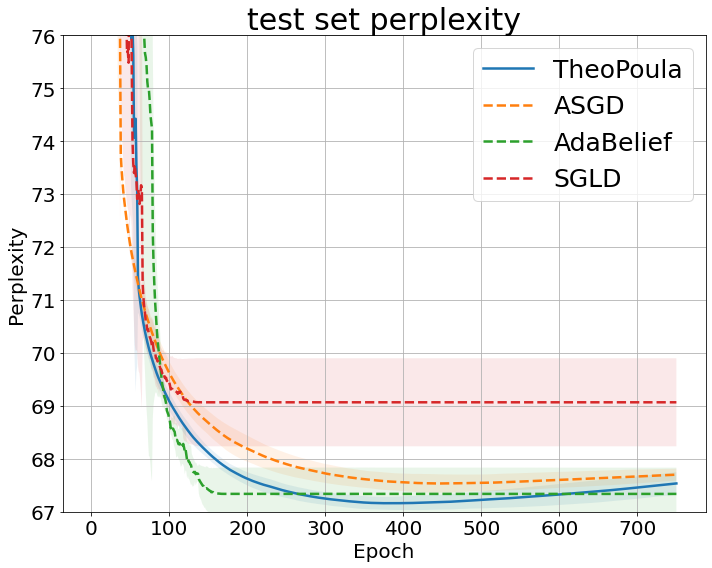}
        \caption{2-layer}        
    \end{subfigure}
    \begin{subfigure}[b]{0.325\textwidth}
        \includegraphics[width=\textwidth]{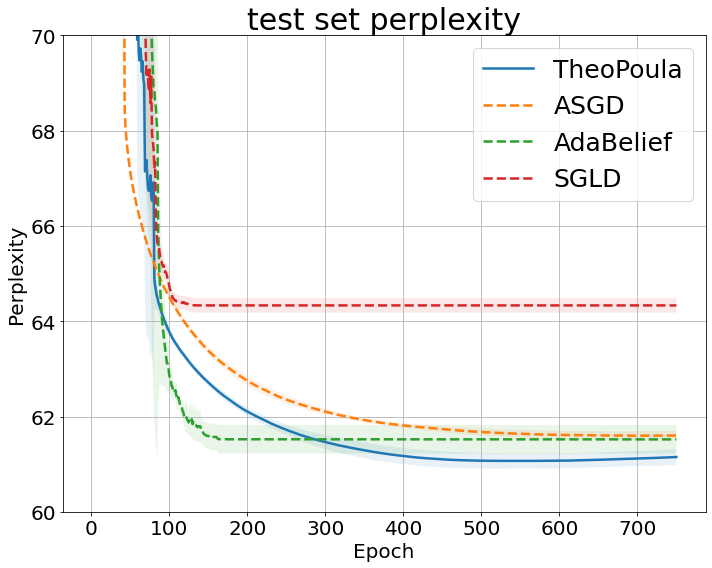}
        \caption{3-layer}        
    \end{subfigure}
    \caption{Test perplexity for 1, 2 and 3-layer AWD-LSTMs on PTB}
    \label{fig:ptb}
\end{figure}

The performance of each optimization algorithm is measured by test perplexity (The lower is better). Figure~\ref{fig:ptb} displays test perplexity of different algorithms for different AWD-LSTM models on PTB. Table~\ref{tab:lstm} shows that TheoPouLa  attains the lower test perplexity against the baselines for AWD-LSTM with $1$, $2$, and $3$-layers. AdaBelief shows a comparable performance with ASGD for 2-layer and 3-layer models.

\begin{table}
\footnotesize
\centering
\caption{Test perplexity for language modeling tasks on PTB. Lower is better. }
\begin{tabular}{c | c | c | c}
 \hline
\# of layers & 1-layer  & 2-layer &  3-layer \\\hline
\multirow{2}{*}{TheoPouLa} &  \textbf{82.75} & \textbf{67.15} & \textbf{61.07}\\
& (0.209) & (0.126) & (0.161)\\
 \hline
 ASGD & 82.85  &  67.53 & 61.60\\
(baseline) & (0.308) & (0.171) & (0.094)\\ \hline
\multirow{2}{*}{AdaBelief} & 84.46 & 67.34 & 61.52\\
& (0.272) & (0.496) & (0.302)\\ \hline
\multirow{2}{*}{SGLD} & { 85.64} & { 69.07} & { 64.33}\\
& { (0.126)} & { (0.832)} & { (0.150)}\\
 \hline
\end{tabular}
\label{tab:lstm}
\end{table}

\subsection{Effect of $\lambda$, $\varepsilon$, and $\beta$ on the performance of TheoPouLa}\label{sec:sensitive}
This section perform a sensitive analysis to understand the effect of key hyperparameters  $\lambda$, $\varepsilon$, and $\beta$, on the performance of TheoPouLa. We consider image classification for VGG11 and ResNet34 on CIFAR10 and CIFAR100. As in Section~\ref{sec:empirical_exp}, we train the models for $200$ epochs with $128$ batch size and then evaluate their test accuracy under varying $\lambda$, $\varepsilon$, and $\beta$.

We first report the effect of $\lambda$ on the performance of TheoPouLa. To see this, we evaluate the models trained with $\lambda \in \{0.5, 0.1, 0.05, 0.01\}$, $\varepsilon=0.1$, $\beta=10^{10}$, $r=0$, and $\eta=0.0005$. As shown in Table~\ref{tab:sen_lambda}, $\lambda=0.1$ yields the best accuracy for VGG on CIFAR10 and CIFAR100, and ResNet on CIFAR10, but $\lambda=0.05$ achieves the highest accuracy for ResNet on CIFAR100.

\begin{table}[h]
\footnotesize
\centering
\caption{The test accuracy for VGG11 and ResNet34 on CIFAR-10 and CIFAR-100 with different $\lambda$. We report mean and standard deviation of the accuracy from three repetitive experiments.}
\begin{tabular}{c | c | c | c | c | c }
 \hline
\multicolumn{2}{c|}{} & \multicolumn{4}{c}{$\lambda$} \\ \hline
model & dataset & $0.5$ & $0.1$ & $0.05$ & $0.01$  \\
\hline
\multirow{4}{*}{VGG} & \multirow{2}{*}{CIFAR10} & 42.31  & 92.14 & 91.82 & 89.89  \\
  & &  (4.996) & (0.201) & (0.081) & (0.060) \\ \cline{2-6}
  & \multirow{2}{*}{CIFAR100} & 31.59  & 70.57 & 68.96 & 63.98  \\
  & &  (2.929) & (0.279) & (0.205) & (0.185)  \\ \cline{2-6}
\hline
\multirow{4}{*}{ResNet} & \multirow{2}{*}{CIFAR10} & 91.23  & 95.41 & 95.09 & 93.54  \\
  & &  (0.312) & (0.175) & (0.117) & (0.277)  \\ \cline{2-6}
  & \multirow{2}{*}{CIFAR100} & 1.90  & 74.02 & 77.31 & 75.49 \\
  & &  (0.316) & (0.609) & (0.248) & (0.066) \\ \cline{2-6}
\hline
\end{tabular}
\label{tab:sen_lambda}
\end{table}

The hyperparameter $\varepsilon$ controls the intensity of the boosting function as the stepsize can increase by up to $\sqrt{\lambda}/\varepsilon$. The effect of the boosting function can be exaggerated when $\varepsilon$ is small, whereas a large $\varepsilon$ depresses the role of the boosting function. To see the effect of $\varepsilon$, we fix $\lambda=0.1$ and $\beta=10^{10}$ and then conduct experiments with different $\varepsilon \in \{1, 0.1, 0.01, 0.001\}$. Table~\ref{tab:sen_eps} summarizes the test accuracy with varying $\varepsilon$. TheoPouLa with $\varepsilon=0.1$ shows the highest accuracy for VGG on CIFAR10 and CIAR100, and ResNet on CIFAR10. On the other hand, $\varepsilon=0.01$ is the best hyperparameter for ResNet on CIFAR100. Moreover, it is observed that too small $\varepsilon$, $0.001$, leads to unstability for VGG on CIFAR10.

\begin{table}[h]
\footnotesize
\centering
\caption{The test accuracy for VGG11 and ResNet34 on CIFAR-10 and CIFAR-100 with different $\varepsilon$. We report mean and standard deviation of the accuracy from three repetitive experiments.}
\begin{tabular}{c | c | c | c | c | c }
 \hline
\multicolumn{2}{c|}{} & \multicolumn{4}{c}{$\varepsilon$} \\ \hline
model & dataset & $1$ & $0.1$ & $0.01$ & $0.001$  \\
\hline
\multirow{4}{*}{VGG} & \multirow{2}{*}{CIFAR10} & 91.80  & 92.14 & 86.75 & 26.03  \\
  & &  (0.262) & (0.201) & (0.489) & (14.473) \\ \cline{2-6}
  & \multirow{2}{*}{CIFAR100} & 68.75  & 70.57 & 60.55 & 44.74  \\
  & &  (0.523) & (0.279) & (0.202) & (0.383)  \\ \cline{2-6}
\hline
\multirow{4}{*}{ResNet} & \multirow{2}{*}{CIFAR10} & 91.23  & 95.41 & 95.09 & 93.54  \\
  & &  (0.312) & (0.175) & (0.117) & (0.277)  \\ \cline{2-6}
  & \multirow{2}{*}{CIFAR100} & 1.90  & 74.02 & 77.31 & 75.49 \\
  & &  (0.316) & (0.609) & (0.248) & (0.066) \\ \cline{2-6}
\hline
\end{tabular}
\label{tab:sen_eps}
\end{table}

Lastly, we evaluate the effect of the inverse temperature $\beta>0$, which is a unique feature of Langevin-based algorithms. Intuitively, a small inverse temperature offers relatively strong random shocks at each iteration, which is helpful to escape sharp local minima, so called \textit{the exploration effect}. On the other hand, the solutions generated with a large inverse temperature explores the local geometry of the objective function, so called \textit{the exploitation effect}. To leverage the trade-off of $\beta$, it is desirable to apply simulated annealing (\citealt{mang:18}) and simulated tempering (\citealt{lee:18}), which often requires intensive effort for the hyperparameter tuning. In this paper, we fix $\beta$ as a constant during the training. We obtain the test accuracy of TheoPouLa with different values of $\beta$ ranging from $\{10^4, 10^6, 10^8, 10^{10}, 10^{12}\}$. The other hyperparameters are as follows: $\lambda=0.1$, $\varepsilon=0.1$ for VGG on CIFAR10 and CIFAR100, and ResNet on CIFAR10, $\lambda=0.05$, $\varepsilon=0.01$ for ResNet on CIFAR100. Table~\ref{tab:sen_beta} shows that TheoPouLa achieves its best performance when $\beta$ is large, say $10^{12}$. We note that the result in Table~\ref{tab:sen_beta} is consistent with the cold posterior effect implying that a large inverse temperature $\beta$ improves the model's predictive power, see \cite{aitchison:21} and \cite{wenzel:20}.

\begin{table}[h]
\footnotesize
\centering
\caption{The accuracy for VGG11 and ResNet34 on CIFAR-10 and CIFAR-100 with different $\beta$. We report mean and standard deviation of the accuracy from three repetitive experiments. }
\begin{tabular}{c | c | c | c | c | c | c}
 \hline
\multicolumn{2}{c|}{} & \multicolumn{5}{c}{$\beta$} \\ \hline
model & dataset & $10^4$ & $10^6$ & $10^8$ & $10^{10}$& $10^{12}$ \\
\hline
\multirow{4}{*}{VGG} & \multirow{2}{*}{CIFAR10} & 73.10  & 91.53 & 92.31 & 92.29 & 92.10 \\
  & &  (0.407) & (0.141) & (0.055) & (0.120) & (0.023) \\ \cline{2-7}
  & \multirow{2}{*}{CIFAR100} & 20.69  & 70.0 & 70.28 & 70.16 & 70.31 \\
  & &  (0.718) & (0.343) & (0.124) & (0.110) & (0.117) \\ \cline{2-7}
\hline
\multirow{4}{*}{ResNet} & \multirow{2}{*}{CIFAR10} & 80.84  & 94.67 & 95.42 & 95.34 & 95.43 \\
  & &  (0.264) & (0.145) & (0.117) & (0.141) & (0.095) \\ \cline{2-7}
  & \multirow{2}{*}{CIFAR100} & 63.58  & 77.22 & 77.4 & 77.6 & 77.53 \\
  & &  (0.103) & (0.291) & (0.036) & (0.208) & (0.143) \\ \cline{2-7}
\hline
\end{tabular}
\label{tab:sen_beta}
\end{table}

\subsection{Additional Experiments}\label{sec:add_exp}


\paragraph{Experiments with $r \geq \frac{1}{2}$.}\label{app:nonzero_eta}

In Section~\ref{sec:empirical_exp}, we set $r=0$ in \eqref{eq:reg_opt} as $\ell_2$-regularization is widely used in practice. However, our main results are derived with $r\geq \frac{q}{2}\geq \frac{1}{2}$ to theoretically ensure the dissipativity condition. In this section, we perform additional experiments with $r\geq \frac{1}{2}$ to demonstrate that there is no gap between theory and practice.

When the regularization parameter $r$ is sufficiently large and the dimension $d$ is big, $|\theta|^{2r}$ becomes extremely large. As a result, the stochastic gradient of the regularization term $\eta\frac{\theta^{(i)}|\theta|^{2r}}{1+\sqrt\lambda |\theta|^{2r}}$ in \eqref{eq:theo_poula} approximately ends up with $\frac{\eta}{\sqrt \lambda} \theta^{(i)}$, which is equivalent to the same regularization effect of $\ell_2$-regularization. Therefore, by choosing $\eta = 5\times 10^{-4} \sqrt\lambda$ and large $r$, the performance of the models with $\eta = 5\times 10^{-4} \sqrt \lambda$ and large $r$ is similar to that of the models with $\ell_2$-regularization (i.e., $r=0$) and $\eta = 5\times 10^{-4}$.

Table~\ref{tab:nonzero_eta} shows that the accuracy for VGG, ResNet and DenseNet on CIFAR10 and CIFAR100 with $r =10$ and $\eta=5\times10^{-4}\sqrt\lambda$. We use the same best hyperparameters used in Section~\ref{sec:empirical_exp}. As shown in Table~\ref{tab:nonzero_eta}, one observes the experiments with large $r$ is highly similar to the results in Section~\ref{sec:empirical_exp}.

\begin{table}[h]
\footnotesize
\centering
\caption{The accuracy for VGG11, ResNet34 and DenseNet121 on CIFAR10 and CIFAR100 with $r=10$.}
\begin{tabular}{c | c | c | c | c | c | c}
 \hline
dataset & \multicolumn{3}{c|}{CIFAR10} & \multicolumn{3}{c}{CIFAR100} \\\hline
model & VGG & ResNet & DenseNet & VGG & ResNet & DenseNet \\
 \hline
TheoPouLa  & 92.2 & 95.38 & 95.69 &  70.07 & 77.78 & 80.47\\
 \hline
\end{tabular}
\label{tab:nonzero_eta}
\end{table}

\paragraph{Effectiveness of the boosting function.}\label{app:boosting}

This subsection empirically tests the effectiveness of the boosting function in our algorithm. TheoPouLa without the boosting function is given by
\begin{eqnarray}\label{eq:tuslac}
H_{\lambda,c}^{(i)}(\theta, x) = \frac{G^{(i)}(\theta, x)}{1+ \sqrt{\lambda }| G^{(i)}(\theta, x)|} + \eta\frac{\theta^{(i)}|\theta|^{2r}}{1 + \sqrt{\lambda}|\theta|^{2r}}.
\end{eqnarray}
Indeed, this is a special case of TheoPouLa with $\varepsilon = \infty$. We train VGG11, ResNet34 and DenseNet121 on CIFAR-10 and CIFAR-100 using TheoPoula without the boosting function. We use the hyperparameters as the best hyperparameters of TheoPouLa in Section~\ref{sec:empirical_exp}. Table~\ref{tab:effect_boosting} clearly shows that the performance of TheoPouLa without the boosting function deteriorates, confirming that the addition of the boosting function brings meaningful increase in test accuracy. 

\begin{table}[h]
\footnotesize
\centering
\caption{The best accuracy for VGG11, ResNet34 and DenseNet121 on CIFAR-10 and CIFAR-100 obtained from TheoPouLa with/without the boosting function.  }
\begin{tabular}{c | c | c | c | c | c | c}
 \hline
dataset & \multicolumn{3}{c|}{CIFAR-10} & \multicolumn{3}{c}{CIFAR-100} \\\hline
model & VGG & ResNet & DenseNet & VGG & ResNet & DenseNet \\
 \hline
TheoPouLa  & \textbf{92.30} & \textbf{95.43} & \textbf{95.66} &  \textbf{70.31} & \textbf{77.60} & \textbf{79.90}\\ \hline
TheoPouLa ($\varepsilon=\infty$)  & 91.48  & 94.31 & 94.22 & 68.11 & 75.91 & 77.99\\
 \hline
\end{tabular}
\label{tab:effect_boosting}
\end{table}

\section{Overview of the Proofs}\label{sec:proofs}

This section provides an overview of the proofs of Theorem~\ref{Thrm1}, Corollary~\ref{Thrm2} and Theorem~\ref{Thrm3}. In Section~\ref{sec:aux_process}, we introduce suitable Lyapunov functions and auxiliary processes, which are necessary to analyze the convergence of TheoPouLa. Then, necessary momentum bound for the auxiliary processes are estimated. Lastly, the proofs of main results can be found in Section~\ref{sec:main_proofs}.

\subsection{Auxiliary processes.}\label{sec:aux_process} For each $p\geq 1$, define the Lyapunov function $V_p$ by, for all $\theta\in \mR^d$,
\begin{equation}
V_p(\theta):= (1+|\theta|^2)^{\frac{p}{2}},  \label{def:Lyapunov}
\end{equation}
and similarly, define  $\mathrm{v}_p(x) = (1+x^2)^\frac{p}{2}$ for $x\geq 0$. Both functions are continuously differentiable and $\lim_{|\theta|\rightarrow \infty} \nabla V_p(\theta)/V_p(\theta)=0$. Also, denote by $Z_t^\lambda := Z_{\lambda t}, t\in \mR_+$, the time-changed Langevin dynamics of \eqref{eq:lagenvin_sde} given by
\begin{equation}\label{time_changed_Langevin}
\rd Z_t^\lambda = -\lambda h(Z_t^\lambda)\rd t + \sqrt{2\lambda\beta^{-1}} \rd B_t^\lambda,
\end{equation}
with $t>0$, $Z_0:=\theta_0$, where $B_t^\lambda: = B_{\lambda t} / \sqrt{\lambda}$ is a $d$-dimensional standard Brownian motion. We then consider the continuous-time interpolation of the TheoPouLa algorithm \eqref{Tamed-Hybrid-PoULA}, denoted by $(\bar{\theta}_t^{\lambda})_{t\in \mR_+}$ as
\begin{equation} \label{eq:CI_Tamed-Hybrid-PoULA}
    \rd \bar{\theta}_{t}^{\lambda}=-\lambda H_\lambda\left(\bar{\theta}_{\lfloor t\rfloor}^{\lambda}, X_{\lceil t \rceil}\right) \rd t+ \sqrt{2 \lambda\beta^{-1}} \rd B_t^\lambda
\end{equation}
with the initial condition $\bar{\theta}_{0}^{\lambda}=\theta_0$. Here, $\lfloor x\rfloor$ denotes the integer part of a positive real $x$ and $\lceil x \rceil := \lfloor x\rfloor+1$. Due to the construction of \eqref{eq:CI_Tamed-Hybrid-PoULA}, the law of interpolated process is equivalent to the law of the TheoPouLa algorithm at grid points, i.e., $\mathcal{L}( \bar{\theta}_{n}^{\lambda})=\mathcal{L}(\theta_{n}^{\lambda})$, for all $n \in \mN_0$.

Furthermore, define the continuous-time process $\zeta_{t}^{s, v, \lambda}, t \geq s$ which is the solution to the following SDE:
\begin{equation} \label{zeta}
    \mathrm{d} \zeta_{t}^{s, v, \lambda}=-\lambda h\left(\zeta_{t}^{s, v, \lambda}\right) \mathrm{d} t+ \sqrt{2 \lambda\beta^{-1}} \mathrm{d} B_t^\lambda,
\end{equation}
with the initial condition $\zeta_{s}^{s, v, \lambda}:=v\in \mathbb{R}^{d}$. 

\begin{definition}
For each fixed $\lambda >0$ and $n\in \mN_0$,  define $\bar{\zeta}_{t}^{\lambda, n}:=\zeta_{t}^{n T, \bar{\theta}_{n T}^{\lambda}, \lambda}, t\geq nT$
where $T=\lfloor 1/\lambda \rfloor$ and $\zeta_{t}^{s,v,\lambda}$ is given in \eqref{zeta}.
\end{definition}



\subsection{Primary estimates}

We provide moment estimates for $(\theta_n^\lambda)_{n\geq 1}$ in the following two lemmas. Recall that $\lambda_{\max}$ is defined in \eqref{eq:lambda_max}.

\begin{lemma}\label{lem:l2-norm}
Let Assumption \ref{ass:init_ass} and \ref{ass:stoc_grad} hold. Then there exists $M_0>0$ such that ,for $0<\lambda\leq\lambda_{\max}$, $n\in \mN_0$,
\begin{align*}
\mE|\theta_{n+1}^\lambda|^2 \leq & \bigg(1- \frac{\eta}{2}\sqrt{\lambda}\bigg)^n \mE|\theta_0|^2 + \bigg[5M_0^2 + \frac{4d}{\eta}\bigg(\beta^{-1} +4\bigg)  + \frac{8\sqrt{d} M_0}{\eta} +4\eta M_0^2 \bigg],
\end{align*}
and
\begin{align*}
\sup_n \mE|\theta_{n+1}^\lambda|^2 \leq & \mE|\theta_0|^2 + \bigg[5M_0^2 + \frac{4d}{\eta}\bigg(\beta^{-1} +4\bigg) + \frac{8\sqrt{d} M_0}{\eta} +4\eta M_0^2 \bigg].
\end{align*}
\end{lemma}

\begin{proof}
See Appendix \ref{proof:l2-norm}.
\end{proof}

\begin{lemma}\label{lem:l2p-norm}
Let Assumption \ref{ass:init_ass} and \ref{ass:stoc_grad} hold. Then, one obtains that, for all $0<\lambda <\lambda_{\max})$, $n\in \mN_0$, $p\in [1, 8(2r+1)]$,
$$
\mE|\theta_{n+1}^\lambda|^{2p} \leq  (1-\eta^2\lambda)^n \mE|\theta_0^\lambda|^{2p} + \frac{A_p }{\eta^2},
$$
and
$$
\sup_{n\in\mN} \mE|\theta_{n+1}^\lambda|^{2p} \leq  \mE|\theta_0^\lambda|^{2p} + \frac{A_p }{\eta^2},
$$
where $A_p$ is given in Table~\ref{tab:constant}.
\end{lemma}
\begin{proof}
See Appendix \ref{proof:l2p-norm}.
\end{proof}

Using Lemma~\ref{lem:l2-norm} and Lemma~\ref{lem:l2p-norm}, one can establish the fourth moment bounds for $(\bar{\theta}_t^{\lambda})_{t\geq nT}$ and $(\bar{\zeta}_t^{\lambda, n})_{t\geq nT}$. To this end, we introduce a drift condition for the Lyapunov function, which is stated in the following lemma.
\begin{lemma}\label{lem:drift_lyapunov}
Let Assumption \ref{ass:init_ass} and \ref{ass:stoc_grad} hold. Then, one obtains, for any $p\in [2, \infty)\cap \mN$, $\theta\in \mR^d$,
$$
\frac{\nabla V_p(\theta)}{\beta} -\ip{\nabla V_p(\theta)}{h(\theta)} \leq -\bar{c}(p)V_p(\theta) + \tilde{c}(p),
$$
where $\bar{c}(p) = Ap/4$, $\tilde{c}(p) = (3/4)Ap\mathrm{v}_p(\bar{M}_p)$, $\bar M_p =(1/3 + 4B/(3A) + 4d/(3A\beta) + 4(p-2)/(3A\beta))^{1/2}$, and $A$, $B$ are explicitly given in Remark~\ref{rem:dissipativity}.
\end{lemma}
\begin{proof}
See \cite[Lemma 3.5]{chau:19}.
\end{proof}
\begin{lemma}\label{lem:V4}
Let Assumption \ref{ass:init_ass} and \ref{ass:stoc_grad} hold. Then, one obtains, for $n\in \mN_0$, $0<\lambda \leq \lambda_{\max}$,
$$
\mE [V_4(\bar \theta^\lambda_{nT})] \leq 2 \mE|\theta_0|^4 + 2+ 2\frac{A_2}{\eta^2},
$$
where $A_2$ is given in Table~\ref{tab:constant}.
\end{lemma}
\begin{proof}
From the definition of the Lyapunov function $V_m$ given in \eqref{def:Lyapunov} and Lemma~\ref{lem:l2p-norm}, we have
\begin{eqnarray*}
\mE [V_4(\bar \theta^\lambda_{nT})] &=& \mE [ (1+ |\bar \theta^\lambda_{nT}|^2)^2] \\
&\leq& 2 + 2\mE  |\bar \theta^\lambda_{nT}|^4 \\
&\leq& 2 + 2\mE |\theta_0|^4 + 2\frac{A_2}{\eta^2}.
\end{eqnarray*}
\end{proof}

\begin{lemma}\label{lem3.7} Let Assumption \ref{ass:init_ass} and \ref{ass:stoc_grad} hold. Then, one obtains that, for $n\in \mN_0$, $0<\lambda\leq \lambda_{\max}$, $t\in (nT, (n+1)T]$,
\begin{eqnarray*}
\mE[V_4(\bar{\zeta}_{t}^{\lambda, n})] &\leq& 2\mE|\theta_0|^4 + 2 + \frac{2A_2}{\eta^2} + \frac{\tilde c(4)}{\bar c(4)},
\end{eqnarray*}
where $\bar c(4)$, $\tilde c(4)$, $A_2$ are given in Table~\ref{tab:constant}.
\end{lemma}
\begin{proof}
See Appendix \ref{proof:V24}.
\end{proof}

Denote by $\mathcal{P}_{V_2}(\mR^d)$ the subset of $\mathcal{P}(\mathbb{R}^d)$ such that $\mu \in\mathcal{P}_{V_2}(\mR^d)$ satisfies $\int_{\mathbb{R}^d} V_2(\theta)\mu(d\theta)<\infty$. Moreover, we consider the following functional, for all $p\geq 1$, $\mu, \nu \in \mathcal{P}_{V_2}(\mR^d)$,
\begin{equation} \label{seminorm}
    w_{1, 2}(\mu, \nu):=\inf _{\zeta \in \mathcal{C}(\mu, \nu)} \int_{\mathbb{R}^{d}} \int_{\mathbb{R}^{d}}\left[1 \wedge | \theta-\theta|^{\prime}\right]\left[\left(1+V_{2}(\theta)+V_{2}\left(\theta^{\prime}\right)\right) \zeta\left(\mathrm{d} \theta \mathrm{d} \theta^{\prime}\right)\right.,
\end{equation}
where $\mathcal{C}(\mu, \nu)$ is defined in \eqref{eq:definition-W-p}. For any $ \mu,\nu \in \mathcal{P}_{V_2}(\mR^d)$, the following inequalities hold:
\begin{equation}
W_1(\mu,\nu)\leq w_{1,2}(\mu,\nu), \quad W_2(\mu,\nu)\leq \sqrt{2w_{1,2}(\mu,\nu)}. \label{ineq:semi_norm}
\end{equation}

The following lemma states the contraction property of the Langevin SDE \eqref{time_changed_Langevin} in $w_{1,2}$, which is a key result of our analysis.

\begin{lemma}\label{eberle}
Let Assumption \ref{ass:init_ass} and \ref{ass:stoc_grad} hold. Let $Z_{t}^{\prime}, t \in \mathbb{R}_{+}$ be the solution of the Langevin SDE \eqref{eq:lagenvin_sde} with initial condition $Z_{0}^{\prime}=\theta'_{0}$ which is independent of $\mathcal{G}_{\infty}$ and $\left|\theta'_{0}\right| \in L^{2} .$ Then, one obtains
\[
w_{1,2}\left(\mathcal{L}(Z_{t}^\lambda ), \mathcal{L}(Z_{t}^{\prime})\right) \leq \hat{c} e^{-C_0 t} w_{1,2}\left(\mathcal{L}\left(\theta_{0}\right), \mathcal{L}\left(\theta_{0}^{\prime}\right)\right)
\]
where $w_{1,2}$ is given in \eqref{seminorm}. The constant $C_0$ is given by
$$
C_0 := \min \{\bar{\phi}, \bar{c}(2), 4\tilde{c}(2) \bar{c}(2)\epsilon\}/2,
$$
where $\bar{c}(2) = A/2$, $\tilde{c}(2) = (3/2) A\mathrm{v}_2(\bar {M}_2)$, $\bar{M}_2 = (1/3 + 4B/(3A) + 4d/(3A\beta))^{1/2}$, $\bar \phi$ is defined by
$$
\bar\phi = \bigg( \sqrt{8\beta\pi/L_R}\dot{c}_0\exp\left\{\left(\dot{c}_0 \sqrt{\beta L_R/8} + \sqrt{8/(\beta L_R)}\right)^2\right\}\bigg)^{-1},
$$
where $L_R$ is defined in Proposition~\ref{prop:2.6}, and $\epsilon$ is chosen such that the following inequality is satisfied
$$
\epsilon \leq 1 \wedge \bigg( 4\tilde{c}(2)\sqrt{2\beta\pi/L_R}\int_0^{\dot{c}_1} \exp\left\{\left(s\sqrt{\beta L_R/8}+\sqrt{8/(\beta L_R)}\right)^2\right\} \mathrm{d}s    \bigg)^{-1},
$$
with $\dot{c}_0 = 2\left(4\tilde{c}(2) (1+\bar{c}(2))/\bar{c}(2) -1\right)^{1/2}$ and $\dot{c}_1=2\left(\tilde{c}(2)/\bar{c}(2)-1\right)^{1/2}$. Moreover, the constant $\hat{c}$ is given by
$$
\hat{c}= 2(1+\dot{c}_0)\exp(\beta L_R \dot{c}_0^2/8 + 2\dot{c}_0)/\epsilon.
$$
\end{lemma}
\begin{proof}
See \cite[Proposition 3.14]{chau:19}.
\end{proof}

We are now able to provide non-asymptotic bounds in the Wasserstein distances between the Auxiliary processes, namely $W_{2}\left(\mathcal{L}(\bar{\theta}_{t}^{\lambda}), \mathcal{L}(\bar{\zeta}_{t}^{\lambda, m})\right)$, $W_{1}\left(\mathcal{L}(\bar{\zeta}_{t}^{\lambda, m}), \mathcal{L}(Z_{t}^{\lambda})\right)$, and $W_{2}\left(\mathcal{L}(\bar{\zeta}_{t}^{\lambda, m}), \mathcal{L}(Z_{t}^{\lambda})\right)$. These results are key components to derive our main results.

\begin{lemma} \label{Lemma 4.7}
Let Assumptions \ref{ass:init_ass} and \ref{ass:stoc_grad} hold. Then, one obtains, for all $0<\lambda\leq \lambda_{max}$, $n\in \mN_0$, $t\in (nT, (n+1)T]$,
\[W_{2}\left(\mathcal{L}(\bar{\theta}_{t}^{\lambda}), \mathcal{L}(\bar{\zeta}_{t}^{\lambda, n})\right) \leq \sqrt \lambda \sqrt{e^{3 L_R} (\bar{C}_1+\bar{C}_2+\bar{C}_3)}\]
where the constants $\bar {C}_1$, $\bar{C}_2$, $\bar{C}_3$ are given explicitly in Table~\ref{tab:constant}, and $L_R$ is given in Proposition~\ref{prop:2.6}.
\end{lemma}
\begin{proof}
See Appendix~\ref{proof:thetazeta}.
\end{proof}

\begin{lemma}\label{Lemma 4.8}
Let Assumptions \ref{ass:init_ass} and \ref{ass:stoc_grad} hold. Then, one obtains, for all $0<\lambda\leq \lambda_{max}$, $n\in \mN_0$, $t\in (nT, (n+1)T]$,
\[W_{1}\left(\mathcal{L}(\bar{\zeta}_{t}^{\lambda, n}), \mathcal{L}(Z_{t}^{\lambda})\right)\leq \sqrt \lambda z_1 \]
where $z_1$ is  given explicitly in Table~\ref{tab:constant}.
\end{lemma}
\begin{proof}
See Appendix~\ref{proof:zetaZ1}.
\end{proof}

\begin{lemma}\label{Lemma A.5.}Let Assumptions \ref{ass:init_ass} and \ref{ass:stoc_grad} hold. Then, one obtains, for all $0<\lambda\leq \lambda_{max}$, $n\in \mN_0$, $t\in (nT, (n+1)T]$,
$$
W_{2}\left(\mathcal{L}\left(\bar{\zeta}_{t}^{\lambda, n}\right), \mathcal{L}\left(Z_{t}^{\lambda}\right)\right)\leq \lambda^{\frac{1}{4}} z_2
$$
where $z_2$ is  given explicitly in Table~\ref{tab:constant}.
\end{lemma}
\begin{proof}
See Appendix~\ref{proof:zetaZ2}.
\end{proof}

\subsection{Proofs of main results}\label{sec:main_proofs}

To derive non-asymptotic upper bounds for $W_{1}\left(\mathcal{L}(\theta_{t}^{\lambda}), \pi_{\beta}\right)$ and $W_{2}\left(\mathcal{L}(\theta_{t}^{\lambda}), \pi_{\beta}\right)$, we consider the following decomposition in terms of the auxiliary processes $\bar \theta_t^\lambda$, $\bar \zeta_t^{\lambda, n}$, and $Z_t^\lambda$ as follows:
$$
W_{j}\left(\mathcal{L}(\theta_{t}^{\lambda}), \pi_{\beta}\right) \leq W_{j}\left(\mathcal{L}(\bar \theta_{t}^{\lambda}), \mathcal L (\bar \zeta_{t}^{\lambda, n})\right) + W_{j}\left(\mathcal{L}( \bar \zeta_{t}^{\lambda, n} ), \mathcal{L}\left( Z_t^\lambda \right)\right) + W_{j}\left(\mathcal{L}(Z_{t}^{\lambda}), \pi_{\beta}\right),
$$
for $j=1, 2$.

\begin{proof}[\textbf{Proof of Theorem~\ref{Thrm1}}]
Observe that $W_{1}\left(\mathcal{L}(\theta_{n}^{\lambda}), \mathcal L (Z_t^\lambda)\right)$ is decomposed as follows: for all $t\in (nT, (n+1)T]$, $n\in \mN_0$, $0<\lambda \leq \lambda_{\max}$,
\begin{equation}
W_{1}\left(\mathcal{L}(\bar{\theta}_{t}^{\lambda}), \pi_{\beta}\right) \leq W_{1}\left(\mathcal{L}(\bar \theta_{t}^{\lambda}), \mathcal{L}(Z_{t}^{\lambda})\right) + W_{1}\left(\mathcal{L}(Z_{t}^{\lambda}), \pi_{\beta}\right). \label{eq:w1_decom}
\end{equation}

Then, from the results of Lemma~\ref{Lemma 4.7} and \ref{Lemma 4.8}, the first term in \eqref{eq:w1_decom} is estimated by, for $t\in (nT, (n+1)T]$, $0<\lambda \leq \lambda_{\max}$, $n\in \mN_0$,
\begin{eqnarray}
W_{1}\left(\mathcal{L}(\bar \theta_{t}^{\lambda}), \mathcal L (Z_t^\lambda)\right) &\leq&  W_{1}\left(\mathcal{L}(\bar \theta_{t}^{\lambda}), \mathcal L (\bar \zeta_{t}^{\lambda, n})\right) + W_{1}\left(\mathcal{L}( \bar \zeta_{t}^{\lambda, n} ), \mathcal{L}( Z_t^\lambda )\right) \nonumber \\
&\leq&  W_{2}\left(\mathcal{L}(\bar \theta_{t}^{\lambda}), \mathcal L (\bar \zeta_{t}^{\lambda, n})\right) + W_{1}\left(\mathcal{L}( \bar \zeta_{t}^{\lambda, n} ), \mathcal{L}( Z_t^\lambda )\right) \nonumber \\
&\leq& \sqrt\lambda ( \sqrt{e^{3L_R } (\bar{C}_1+\bar{C}_2+\bar{C}_3)} + z_1) \nonumber \\
&\leq& C_1 \sqrt \lambda,  \label{ineq:w1_0}
\end{eqnarray}
where $C_1:=  \sqrt{e^{3L_R } (\bar{C}_1+\bar{C}_2+\bar{C}_3)} + z_1$.

Consequently,  using \eqref{ineq:w1_0}, \eqref{ineq:semi_norm}, and Lemma \ref{eberle}, we derive, for $t\in (nT, (n+1)T]$, $0<\lambda \leq \lambda_{\max}$, $n\in \mN_0$,
\begin{eqnarray*}
W_{1}\left(\mathcal{L}(\bar{\theta}_{t}^{\lambda}), \pi_{\beta}\right) &\leq& \sqrt\lambda ( \sqrt{e^{3L_R} (\bar{C}_1+\bar{C}_2+\bar{C}_3)} + z_1) + w_{1,2}\left(\mathcal{L}(Z_{t}^{\lambda}), \pi_{\beta}\right) \\
&\leq&\sqrt\lambda ( \sqrt{e^{3L_R} (\bar{C}_1+\bar{C}_2+\bar{C}_3)} + z_1) + \hat{c} e^{-C_0 \lambda t} w_{1,2}(\theta_0, \pi_\beta) \\
&\leq&\sqrt\lambda ( \sqrt{e^{3L_R} (\bar{C}_1+\bar{C}_2+\bar{C}_3)} + z_1)\\
&+& \hat{c} e^{-C_0\lambda t} \bigg[1 + \mE[V_2(\theta_0)] + \int_{\mR^d} V_2(\theta) \pi_\beta (d\theta)\bigg] \\
&\leq& C_1 \sqrt \lambda + C_2 e^{-C_0 n},
\end{eqnarray*}
where $C_2:= \hat{c}  \bigg(1 + \mE[V_2(\theta_0)] + \int_{\mR^d} V_2(\theta) \pi_\beta (d\theta)\bigg)$, which yields that, for all $n\in \mN_0$,
$$
W_{1}\left(\mathcal{L}(\bar{\theta}_{nT}^{\lambda}), \pi_{\beta}\right) \leq C_1 \sqrt \lambda + C_2 e^{-C_0 n}.
$$
By setting $nT$ to $n$ and noticing that $n\lambda \leq n/T$ in the above inequality, we obtain the desired result.
\end{proof}

\begin{proof}[\textbf{Proof of Corollary~\ref{Thrm2}}]
Similar to the proof of Theorem~\ref{Thrm1}, we consider the following decomposition: for $t\in(nT, (n+1)T]$, $0<\lambda \leq \lambda_{\max}$, $n\in \mN_0$,
\begin{eqnarray*}
W_{2}\left(\mathcal{L}(\theta_{t}^{\lambda}), \pi_{\beta}\right) \leq W_{2}\left(\mathcal{L}(\bar \theta_{t}^{\lambda}), \mathcal L (\bar \zeta_{n}^{\lambda, n})\right)  + W_{2}\left(\mathcal{L}(\bar \zeta_{t}^{\lambda, n}), \mathcal L (Z_t^\lambda)\right)
+  W_{2}\left(\mathcal{L}(Z_t^\lambda), \pi_{\beta}\right).
\end{eqnarray*}
Then, using Lemma~\ref{Lemma 4.7}, Lemma~\ref{Lemma A.5.} and \eqref{ineq:semi_norm}, one further obtains that
\begin{eqnarray*}
W_{2}\left(\mathcal{L}(\theta_{t}^{\lambda}), \pi_{\beta}\right)\leq  \sqrt{e^{3L_R}(\bar{C}_1+ \bar{C}_2+\bar{C}_3)}\sqrt \lambda +z_2 \lambda^{\frac{1}{4}} + \sqrt{2w_{1,2}(\mathcal L (Z_t^\lambda), \pi_\beta)},
\end{eqnarray*}
which yields from Lemma~\ref{eberle}
\begin{eqnarray*}
W_{2}\left(\mathcal{L}(\theta_{t}^{\lambda}), \pi_{\beta}\right) &\leq& \left(\sqrt{e^{3a}(\bar{C}_1+ \bar{C}_2+\bar{C}_3)} + z_2\right) \lambda^{\frac{1}{4}}\\
&+& \sqrt{2 \hat{c} }e^{-C_0\lambda t/2} \left(1+\mathbb{E}\left[V_{2}\left(\theta_{0}\right)\right]+\int_{\mathbb{R}^{d}} V_{2}(\theta) \pi_{\beta}(d \theta)\right)^{1/2} \\
&\leq& C_3 \lambda^{\frac{1}{4}} + C_4 e^{-C_5 n}
\end{eqnarray*}
where
\begin{eqnarray*}
C_3 &:=& \sqrt{e^{3a}(\bar{C}_1+ \bar{C}_2+\bar{C}_3)} + z_2, \\
C_4 &:=& \sqrt{2 \hat{c}}  \left(1+\mathbb{E}\left[V_{2}\left(\theta_{0}\right)\right]+\int_{\mathbb{R}^{d}} V_{2}(\theta) \pi_{\beta}(d \theta)\right)^{1/2}, \\
C_5 &=& \frac{C_0}{2}.
\end{eqnarray*}

Therefore, we have for $m\in \mN_0$, $0<\lambda <\lambda_{\max}$,
$$
W_{2}\left(\mathcal{L}(\theta_{mT}^{\lambda}), \pi_{\beta}\right) \leq C_3 \lambda^{\frac{1}{4}} + C_4 e^{-C_5 m},
$$
and by setting $n=mT$ and using $-\lambda m \leq -\lambda n/T \leq - n\lambda$, we complete the proof.

\end{proof}

\begin{proof}[\textbf{Proof of Theorem~\ref{Thrm3}}] 

We begin by decomposing expected excess risk~\eqref{eq:excess_risk} as follows:
\begin{eqnarray}
\mE[u(\theta_n^\lambda)] - u(\theta^*) \leq \mE[u(\theta_n^\lambda)] - \mE[u(Z_\infty)] + \mE[u(Z_\infty)] - u(\theta^*) \label{thm3_second}
\end{eqnarray}
where $Z_\infty$ follows the target invariant measure $\pi_\beta$, i.e., $\mathcal L(Z_\infty)=\pi_\beta$. Let us focus on estimating the first part, $\mE[u(\theta_n^\lambda)] - \mE[u(Z_\infty)]$. Due to Remark \ref{rem:growth_G}, it follows that
\begin{eqnarray*}
|\nabla u(\theta)| &=& |h(\theta)| \leq (2^{2r} \mE[K(X_0)] + \eta ) (1 + |\theta|^{2r+1})\\
&\leq& r_1(1+|\theta|^{2r+1}),
\end{eqnarray*}
where $r_1 := 2^{2r} \mE[K(X_0)] + \eta$. Then, one calculates that, for all $\theta, \theta \in \mR^d$,
\begin{eqnarray}
u(\theta)-u(\theta') &=& \int_0^1 \ip{\nabla u(t\theta+(1-t)\theta')}{\theta-\theta'} \rd t \nonumber \\
&\leq& \int_0^1  \bigg(r_1  + r_12^{2r}(t^{2r+1}|\theta|^{2r+1} + (1-t)^{2r+1}|\theta'|^{2r+1})\bigg) \rd t  |\theta-\theta'| \nonumber \\
&\leq& \bigg( r_1 +  \frac{r_1 2^{2r}}{2r+2}|\theta|^{2r+1}+  \frac{r_1 2^{2r}}{2r+2}|\theta'|^{2r+1}\bigg) |\theta-\theta'|. \label{eq:u-v}
\end{eqnarray}
Let $\mathbf{P}$ denote the coupling between $\mu$ and $\nu$ that achieves $W_2(\mu, \nu)$ with $\mu = \mathcal L(\theta_n^\lambda)$ and $\nu = \mathcal{L}(Z_\infty)$, i.e., $W_2^2\left(\mathcal L(\theta_n^\lambda), \mathcal L(Z_\infty)\right)= \mE_\mathbf{P}\left[|\theta_n^\lambda - Z_\infty|^2\right]$. By using \eqref{eq:u-v} and Cauchy-Schwarz inequality, we obtain
\begin{align}
\mE[u(\theta_n^\lambda)] - \mE[u(Z_\infty)] &= \mE_\mathbf{P} [u(\theta_n^\lambda) - u(Z_\infty)] \nonumber \\
&\leq \mE_\mathbf{P} \bigg[\bigg( r_1 +  \frac{r_1 2^{2r}}{2r+2}|\theta_n^\lambda|^{2r+1}+  \frac{r_1 2^{2r}}{2r+2}|Z_\infty|^{2r+1}\bigg) |\theta_n^\lambda-Z_\infty|\bigg] \nonumber \\
&\leq \bigg( r_1 + \frac{r_12^{2r}}{2r+2}\sqrt{\mE|\theta_n^\lambda|^{4r+2}} + \frac{r_12^{2r}}{2r+2}\sqrt{\mE|Z_\infty|^{4r+2}}\bigg)W_2\bigg(\mathcal L(\theta_n^\lambda), \pi_\beta\bigg) \nonumber \\
&\leq \bigg( r_1 + \frac{r_12^{2r}}{2r+2}\sqrt{\mE|\theta_0^\lambda|^{4r+2} + \frac{A_{2r+1}}{\eta^2}} + \frac{r_12^{2r}}{2r+2}\sqrt{\mE|Z_\infty|^{4r+2}}\bigg)\nonumber \\
&\times W_2\bigg(\mathcal L(\theta_n^\lambda), \pi_\beta\bigg) \nonumber \\ \label{eq:diff_gibbs_n}
&\leq C_6 W_2\bigg(\mathcal L(\theta_n^\lambda), \pi_\beta\bigg)
\end{align}
where we have used Lemma~\ref{lem:l2p-norm} for the last inequality and the constant $C_6$ is given by
$$
C_6 := r_1 + \frac{r_12^{2r}}{2r+2}\sqrt{\mE|\theta_0^\lambda|^{4r+2} + \frac{A_{2r+1}}{\eta^2}} + \frac{r_12^{2r}}{2r+2}\sqrt{\mE|Z_\infty|^{4r+2}}.
$$

We take a similar approach in \citet{raginsky:17} to estimate the second term in the RHS of \eqref{thm3_second}. From Equation (3.18), (3.20) in \citet{raginsky:17}, we obtain
\begin{eqnarray}\label{eq:raginsky}
\mE u(Z_\infty) - u(\theta^*) &\leq&  \frac{1}{\beta}
\bigg(-\int_{\mR^d}\frac{e^{-\beta u(\theta)}}{\Lambda}\log  \frac{e^{-\beta u(\theta)}}{\Lambda} d \theta - \log \Lambda \bigg) - u^* \nonumber \\
&\leq& \frac{d}{2\beta}\log \bigg(\frac{2\pi e(B+d/\beta)}{Ad}\bigg) - \frac{\log \Lambda}{\beta}-u^*
\end{eqnarray}
where $\Lambda=\int_{\mR^d} e^{-\beta u(\theta)}d\theta$ is the normalizing constant.

Using ~\eqref{eq:diss}, we obtain
$$
\ip{\theta^*}{h(\theta^*)} \geq A|\theta^*|^2 -B
$$
which yields
$$
|\theta^*|^2 \leq \sqrt{\frac{B}{A}}.
$$
Moreover, for $w \in \mR^d$, we have
\begin{eqnarray*}
u(\theta^*)-u(w) &=& \int_0^1 \ip{\nabla u(w+t(\theta^*-w))}{\theta^*-w} dt \nonumber \\
&=& \int_0^1 \ip{\nabla u(w+t(\theta^*-w)) - \nabla u(\theta^*)}{\theta^*-w} dt \nonumber \\
&=& \int_0^1 \frac{1}{t-1}\ip{\nabla u(w+t(\theta^*-w)) - \nabla u(\theta^*)}{w - \theta^* + t(\theta^*-w)} dt.
\end{eqnarray*}
From Remark~\ref{rem:h_lipschitz}, we further obtain
\begin{eqnarray*}
-\beta(u(\theta^*) - u(w)) &=& \beta|u(\theta^*) - u(w)| \\
&\leq& \beta \int_0^1 \frac{1}{t-1}|\ip{h(w+t(\theta^*-w)) - h(\theta^*)}{w - \theta^* + t(\theta^*-w)}| dt \\
&\leq& \beta L_h\int_0^1 (1 + |w+ t(\theta^* - w)| + |\theta^*|)^{2r+1} (1-t)|w-\theta^*|^2 dt  \\
&\leq& \beta L_h \int_0^1 (1 + |w|+ |\theta^* - w| + |\theta^*|)^{2r+1} (1-t)|w-\theta^*|^2 dt \\
&=&\beta L_h (1 + 2|\theta^*|+ 2|\theta^* - w|)^{2r+1} \frac{|w-\theta^*|^2}{2}
\end{eqnarray*}
where we have used the elementary inequality $0\leq |w|-|\theta^*| \leq |\theta^*-w|$ for the last inequality.

Define $R_0:= \max\{\sqrt{B/A}, \sqrt{2d/(\beta L_h)}\}$ and $\overline\bB_r(p) = \{x\in \mR^d | |x-p|>r\}$. Then, from the above inequality, one further calculates
\begin{eqnarray}\label{eq:normal_factor}
\frac{\log \Lambda}{\beta} &=& -u(\theta^*) + \frac{1}{\beta} \log \int_{\mR^d} e^{\beta(u(\theta^*)-u(w))}dw \nonumber \\
&\geq& -u(\theta^*) + \frac{1}{\beta} \log\int_{\mR^d} e^{-\beta L_h (1 + 2|\theta^*|+ 2|\theta^* - w|)^{2r+1} \frac{|w-\theta^*|^2}{2}}dw \nonumber \\
&\geq& -u(\theta^*) + \frac{1}{\beta} \log\int_{\overline\bB_{R_0}(\theta^*)} e^{-\beta L_h  (1 + 4R_0)^{2r+1} \frac{|w-\theta^*|^2}{2}}dw \nonumber \\
&=& -u(\theta^*) + \frac{1}{\beta} \log\bigg[\bigg(\frac{2\pi}{\beta K}\bigg)^{d/2}\int_{\overline\bB_{R_0}(\theta^*)} f_X(w)dw\bigg]\nonumber \\
&\geq& -u(\theta^*) + \frac{1}{\beta} \log\bigg(
\frac{1}{2}\bigg(\frac{2\pi}{K\beta}\bigg)^{d/2}\bigg)
\end{eqnarray}
where $K= L_h(1+4R_0)^{2r+1}$ and $f_X$ is the density function of a multivariate normal variable $X$ with mean $\theta^*$ and covariance $\frac{1}{K\beta}I_d$. Here, the last inequality is obtained from the following inequality:
\begin{eqnarray*}
\int_{\overline\bB_{R_0}(\theta^*)} f_X(w)dw &=& P(|X-\theta^*| > R_0) \\
&=& P\bigg( |X-\theta^*| >\sqrt{\frac{K\beta R_0^2}{d}}\sqrt{\frac{d}{K\beta}}\bigg)\\
&\leq& \frac{d}{K\beta R_0^2} \\
&\leq&\frac{1}{2(1+4R_0)^{2r+1}} \\
&\leq& \frac{1}{2}.
\end{eqnarray*}

Combining \eqref{eq:raginsky} and \eqref{eq:normal_factor}, we derive
\begin{eqnarray}
\mE u(Z_\infty) - u(\theta^*) &\leq&  \frac{d}{2\beta}\log \bigg(\frac{2\pi e(B+d/\beta)}{Ad}\bigg)  - \frac{1}{\beta} \log\bigg(
\frac{1}{2}\bigg(\frac{2\pi}{K\beta}\bigg)^{d/2}\bigg)  \nonumber \\
&\leq& \frac{1}{\beta}\bigg[\frac{d}{2}\log \bigg(\frac{K e}{A}\bigg(\frac{B}{d}\beta+1\bigg)\bigg)  +\log 2 \bigg]. \label{eq:diff_gibbs_optimal}
\end{eqnarray}
Consequently, from \eqref{eq:diff_gibbs_n} and \eqref{eq:diff_gibbs_optimal}, we derive
\begin{eqnarray*}
\mE u(\theta_n^\lambda) - u(\theta^*) &\leq& C_6 W_2(\mathcal L(\theta_n^\lambda, \pi_\beta))\\ &+&  \frac{1}{\beta}\left(\frac{d}{2}\log \bigg(\frac{K e}{A}\bigg(\frac{B}{d}\beta+1\bigg)\bigg)  +\log 2\right), \\ 
&\leq& C_6 W_2(\mathcal L(\theta_n^\lambda, \pi_\beta)) + \frac{C_7}{\beta},
\end{eqnarray*}
where $C_7 := \frac{d}{2}\log \bigg(\frac{K e}{A}\bigg(\frac{B}{d}\beta+1\bigg)\bigg)  +\log 2$.
\end{proof}

\section{Conclusion and Discussion}\label{sec:con}
This paper begins with an example which illustrates that local Lipschitz continuous gradients can cause serious convergence issues for popular adaptive optimization methods. Such issues manifest themselves as vanishing/exploding gradient phenomena. It proceeds by proposing a novel optimization framework, which is suitable for the fine tuning of neural network models by combining elements of the theory of Langevin SDEs, tamed algorithms and carefully designed boosting functions that handle sparse and super-linearly growing gradients. Further, a detailed convergence analysis of the newly proposed algorithm TheoPouLa is provided along with full theoretical guarantees for obtaining the best known convergence rates. Our experiments confirm that TheoPouLa outperforms other popular stochastic optimization methods.

We believe that this work opens a new door for stochastic (adaptive) optimization methods beyond the popular ADAM-type framework. Also, there is much room for improvement of our novel framework. For example, the improved performance can be further achieved by identifying more efficient taming and boosting functions, which demonstrates the potential of our framework.


\section*{acknolwedgement}This project has received funding from the European Union’s Horizon 2020 research and innovation programme under the Marie Skłodowska-Curie grant agreement No 801215,  the University of Edinburgh Data-Driven Innovation programme, part of the Edinburgh and South East Scotland City Region Deal, and the Ministry of Trade, Industry and Energy (MOTIE) and Korea Institute for Advancement of Technology (KIAT) through the International Cooperative R\&D program (No.P0025828). Dong-Young Lim acknowledges that this work was supported by Institute of Information \& communications Technology Planning \& Evaluation (IITP) grant funded by the Korea government (MSIT) (No.2020-0-01336, Artificial Intelligence Graduate School Program (UNIST)) and by National Research Foundation of Korea (NRF) grant funded by the Korea government (MSIT) (No.RS-2023-00253002). 


\newpage

\appendix

\section{Auxiliary Results}\label{app:remarks}

This section introduces some auxiliary results and their proofs, which are useful to obtain the main results. 

\begin{proof}[\textbf{Proof of Remark \ref{rem:growth_G}}]
By Assumption~\ref{ass:stoc_grad}, it follows that for all $\theta \in \mR^d$, $x\in \mR^m$,
\begin{eqnarray*}
|G(\theta, x)| &\leq& L_G(1+|x|)^\rho (1+|\theta|)^{q-1} |\theta| + |G(0,x)| \\
&\leq& L_G(1+|x|)^\rho (1+|\theta|)^{q}  + |G(0,x)|  (1+|\theta|)^{q} \\
&\leq& K_G(x) (1+|\theta|)^q,
\end{eqnarray*}
where $K_G(x) = L_G(1+|x|)^\rho + |G(0, x)|$.
\end{proof}

\begin{proof}[\textbf{Proof of Remark \ref{rem:dissipativity}}]
From the definition of $H$ in \eqref{eq:stochastic_gradientH}, one obtains, for all $\theta \in \mR^d$,
\begin{eqnarray}
\ip{\theta}{h(\theta)} &=& \ip{\theta}{\mE[G(\theta, X_0)]} + \ip{\theta}{\eta\theta |\theta|^{2r}} \nonumber\\
&\geq& \eta |\theta|^{2r+2}- 2^q\mE[K_G(X_0)](1+|\theta|^{q+1}) \nonumber\\
&\geq& \eta |\theta|^{2r+2}- 2^q\mE[1 + K_G(X_0)](1+|\theta|^{q+1}). \label{eq:diss_0}
\end{eqnarray}
To prove \eqref{eq:diss}, we want to show
$$
\eta|\theta|^{2r+2} + B \geq A|\theta|^2 + 2^q \mE[1+K_G(X_0)](1+|\theta|^{q+1}),
$$
for $A=2^q \mE[1 + K_G(X_0)]$, $B=3 (2^{q+1} \mE[1+K_G(X_0)])^{q+2} /\eta^{q+1}$.

One first observes that, for $|\theta| \geq  2A/\eta>1$ with $\eta\in (0,1)$ and $r \geq q/2\geq 1/2$,
\begin{eqnarray}
\eta |\theta|^{2r+2} + B &\geq& \frac{\eta}{2}|\theta|^3 +
\frac{\eta}{2}|\theta|^{q+2} + \frac{3(2^{q+1} \mE[1+K_G(X_0)])^{q+2} }{\eta^{q+1}} \nonumber \\
&\geq& 2^q \mE[1+K_G(X_0)] |\theta|^2 + 2^q\mE[ 1+K_G(X_0)]|\theta|^{q+1}
+ 2^q \mE[1+K_G(X_0)] \nonumber \\
&\geq&  A|\theta|^2 + 2^q\mE[ 1+K_G(X_0)](1+|\theta|^{q+1} )
\label{eq:diss_1}
\end{eqnarray}
Similarly, one can check that, for $|\theta|< 2A/\eta$,
\begin{eqnarray}
\eta |\theta|^{2r+2} + B &\geq&  \frac{3(2^{q+1} \mE[1+K_G(X_0)])^{q+2}}{\eta^{q+1}} \nonumber \\
&\geq& \eta\bigg(\frac{2^{q+1} \mE[1+K_G(X_0)]}{\eta} \bigg)^3 + \eta\bigg(\frac{2^{q+1} \mE[1+K_G(X_0)]}{\eta} \bigg)^{q+2} \nonumber \\
&+& 2^q \mE[1+K_G(X_0)] \nonumber\\
&\geq& 2^{q+1} \mE[1+K_G(X_0)]|\theta|^2 + 2^{q+1} \mE[1+K_G(X_0)]|\theta|^{q+1} + 2^q \mE[1+K_G(X_0)] \nonumber\\
&\geq& A|\theta|^2 + 2^q\mE[ 1+K_G(X_0)] (1+|\theta|^{q+1}) \label{eq:diss_2}
\end{eqnarray}
Therefore, using \eqref{eq:diss_0}, \eqref{eq:diss_1}, and \eqref{eq:diss_2}, we have for all $\theta \in \mR^d$
\begin{eqnarray*}
\ip{\theta}{h(\theta)} &\geq& \eta |\theta|^{2r+2}- 2^q\mE[1 + K_G(X_0)](1+|\theta|^{q+1}) \\
&\geq& A|\theta|^2 -B
\end{eqnarray*}
where $A=2^q \mE[1 + K_G(X_0)]$, $B=3 (2^{q+1} \mE[1+K_G(X_0)])^{q+2} /\eta^{q+1}$.
\end{proof}

\begin{remark} \label{remark:diff_H}
Let Assumption~\ref{ass:stoc_grad} holds and recall that the definitions of $H$ and $H_{\lambda, c}$ are given in \eqref{eq:stochastic_gradientH} and \eqref{eq:theo_poula}, respectively. Then, one obtains that, for all $\theta \in \mR^d$, $x\in\mR^m$, $i=1,\ldots, d$,
\begin{equation}
|H^{(i)}(\theta, x) - H_{\lambda,c}^{(i)}(\theta, x) | \leq \left( |G^{(i)}(\theta , x)|^2 + 1 + \eta |\theta^{(i)}| |\theta|^{2r}\right) \sqrt \lambda.
\end{equation}
Moreover, it follows that, for all $\theta \in \mR^d$, $x\in\mR^m$,
\begin{equation}
|H(\theta, x) - H_{\lambda,c}(\theta, x) |^2 \leq 9 \bigg[8|K_G(x)|^4(1+|\theta|^{4q}) + d + \eta^2 |\theta|^{4r+2} \bigg]\lambda.
\end{equation}
\end{remark}

\begin{proof}[\textbf{Proof of Remark \ref{remark:diff_H}}]
Recall the expressions of $H$ and $H_{\lambda, c}$ in \eqref{eq:stochastic_gradientH} and \eqref{eq:theo_poula}, respectively. The difference between the $H^{(i)}$ and $H_{\lambda, c}^{(i)}$ can be estimated by, for $i=1,\ldots, d$,
\begin{eqnarray*}
|H^{(i)}(\theta, x) - H_{\lambda,c}^{(i)}(\theta, x) | &\leq& \bigg|G^{(i)}(\theta, x) - \frac{G^{(i)}(\theta, x)}{1+ \sqrt{\lambda }| G^{(i)}(\theta, x)|}\bigg(1 + \frac{\sqrt{\lambda}}{\varepsilon + |G^{(i)}(\theta, x)|}\bigg) \bigg| \\
&+& \bigg| \eta\theta^{(i)}|\theta|^{2r} -  \eta\frac{\theta^{(i)}|\theta|^{2r}}{1 + \sqrt{\lambda}|\theta|^{2r}}\bigg| \\
&\leq& |G^{(i)}(\theta, x)| \frac{\sqrt \lambda |G^{(i)}(\theta, x)|}{1+ \sqrt{\lambda }| G^{(i)}(\theta, x)|} \\
&+&  \frac{\sqrt{\lambda}|G^{(i)}(\theta, x)|}{(1+ \sqrt{\lambda }| G^{(i)}(\theta, x)|)(\varepsilon + |G^{(i)}(\theta, x)|)}\\
&+& \eta |\theta^{(i)}| |\theta|^{2r}\bigg| \frac{\sqrt{\lambda}|\theta|^{2r}}{1 + \sqrt{\lambda}|\theta|^{2r}}\bigg|\\
&\leq&  \sqrt{\lambda} |G^{(i)}(\theta , x)|^2 + \sqrt\lambda +\sqrt \lambda \eta |\theta^{(i)}| |\theta|^{2r}.
\end{eqnarray*}
By Remark \ref{rem:growth_G} and Cauchy-Schwarz inequality, the above estimate further yields that
\begin{eqnarray*}
|H(\theta, x) - H_{\lambda,c}(\theta, x) |^2 &=& \sum_{i=1}^d \bigg(\sqrt{\lambda} |G^{(i)}(\theta , x)|^2 + \sqrt\lambda +\sqrt \lambda \eta |\theta^{(i)}| |\theta|^{2r}\bigg)^2 \\
&\leq& 9\lambda \sum_{i=1}^d \bigg[|G^{(i)}(\theta , x)|^4+ 1 +\eta^2 |\theta^{(i)}|^2 |\theta|^{4r}\bigg] \\
&\leq& 9\lambda \bigg[\bigg(\sum_{i=1}^d |G^{(i)}(\theta , x)|^2\bigg)^2 + d + \eta^2 |\theta|^{8r+2} \bigg]\\
&\leq& 9\lambda \bigg[|G(\theta , x)|^4 + d + \eta^2 |\theta|^{4r+2} \bigg] \\
&\leq& 9\lambda \bigg[8|K_G(x)|^4(1+|\theta|^{4q}) + d + \eta^2 |\theta|^{4r+2} \bigg].
\end{eqnarray*}
\end{proof}

\begin{remark}\label{rem:H_estimate}
Let Assumption~\ref{ass:stoc_grad} holds and recall that the definitions of $H$ and $H_{\lambda, c}$ are given in \eqref{eq:stochastic_gradientH} and \eqref{eq:theo_poula}, respectively. Then, the growth of $H$ can be estimated as follows: for all $\theta \in \mR^d$, $x \in \mR^m$,
\begin{equation*}
|H(\theta, x)|^2 \leq 4|K_G(x)|^2(1+|\theta|^{2q}) +2\eta^2 |\theta|^{4r+2}.
\end{equation*}
Moreover, one obtains, for all $\theta \in \mR^d$, $x \in \mR^m$,
\begin{equation*}
|H_{\lambda, c}(\theta, x)|^2 \leq 18|K_G(x)|^2(1+|\theta|^{2q}) + 9\lambda d +9\eta^2 |\theta|^{4r+2}.
\end{equation*}
\end{remark}

\begin{proof}[\textbf{Proof of Remark \ref{rem:H_estimate}}]
By Remark~\ref{rem:growth_G}, one calculates that, for all $\theta \in \mR^d$ and $x \in \mR^m$,
\begin{eqnarray*}
|H(\theta, x)|^2 = |G(\theta, x) + \eta\theta|\theta|^{2r}|^2 &\leq& 2|G(\theta, x)|^2 +2\eta^2 |\theta|^{4r+2} \\
&\leq& 4|K_G(x)|^2(1+|\theta|^{2q}) +2\eta^2 |\theta|^{4r+2},
\end{eqnarray*}
and
\begin{eqnarray*}
|H_{\lambda, c}(\theta, x)|^2 &=& \sum_{i=1}^d\left( \frac{G^{(i)}(\theta, x)}{1+ \sqrt{\lambda }| G^{(i)}(\theta, x)|}\bigg(1 + \frac{\sqrt{\lambda}}{\varepsilon + |G^{(i)}(\theta, x)|}\bigg) + \eta\frac{\theta^{(i)}|\theta|^{2r}}{1 + \sqrt{\lambda}|\theta|^{2r}}\right)^2 \\
&\leq& \sum_{i=1}^d\bigg( \frac{|G^{(i)}(\theta, x)|}{1+ \sqrt{\lambda }| G^{(i)}(\theta, x)|} + \frac{\sqrt{\lambda}|G^{(i)}(\theta, x)|}{(1+ \sqrt{\lambda }| G^{(i)}(\theta, x)|)(\varepsilon + |G^{(i)}(\theta, x)|)} \\
&+& \eta\frac{|\theta^{(i)|}|\theta|^{2r}}{1 + \sqrt{\lambda}|\theta|^{2r}}\bigg)^2 \\
&\leq& \sum_{i=1}^d\bigg( |G^{(i)}(\theta, x)| +\sqrt{\lambda} + \eta|\theta^{(i)}||\theta|^{2r}\bigg)^2 \\
&\leq& 9\sum_{i=1}^d  \bigg(|G^{(i)}(\theta, x)|^2 + \lambda + \eta^2 |\theta^{(i)}|^2|\theta|^{4r}\bigg) \\
&\leq& 9 \left(|G(\theta, x)|^2 + \lambda d + \eta^2 |\theta|^{4r+2}\right) \\
&\leq& 18|K_G(x)|^2(1+|\theta|^{2q}) + 9\lambda d +9\eta^2 |\theta|^{4r+2}.
\end{eqnarray*}
\end{proof}

\section{Proofs of Lemmas in Appendix~\ref{sec:proofs}}\label{app:proofs_lemma}
\begin{proof}[\textbf{Proof of Lemma~\ref{lem:l2-norm}}]\label{proof:l2-norm}
For each $i=1, \ldots, d$ and fixed $\varepsilon>0$, define
$$
\widehat G_{\lambda, c}^{(i)}(\theta, x) = \frac{G^{(i)}(\theta, x)}{1+ \sqrt{\lambda }| G^{(i)}(\theta, x)|}\bigg(1 + \frac{\sqrt{\lambda}}{\varepsilon + |G^{(i)}(\theta, x)|}\bigg),
$$
for all $\theta\in \mR^d$, $x\in\mR^m$, $0<\lambda \leq \lambda_{\max}$. Denote by $\theta^{(i)}$ the $i$-th component of $\theta \in \mR^d$ for $i=1,\ldots, d$. One then observes that, for all $\theta \in \mR^d$, $x\in \mR^m$, $i=1,\ldots, d$,
\begin{eqnarray}
|\widehat G_{\lambda,c}^{(i)}(\theta, x)| &=& \frac{|G^{(i)}(\theta, x)|}{1+ \sqrt{\lambda }| G^{(i)}(\theta, x )|} + \sqrt{\lambda}\frac{|G^{(i)}(\theta, x)|}{(1+ \sqrt{\lambda }| G^{(i)}(\theta, x )|)(\varepsilon + |G^{(i)}(\theta, x)|)} \nonumber\\
&\leq& \frac{1}{\sqrt{\lambda} } + \sqrt{\lambda}\frac{|G^{(i)}(\theta, x)|/\varepsilon}{1 + |G^{(i)}(\theta, x)|/\varepsilon}  \nonumber \\
&\leq & \frac{1}{\sqrt{\lambda}} + \sqrt{\lambda}. \label{eq:ineq_H}
\end{eqnarray}
By using Cauchy-Schwartz inequality and \eqref{eq:ineq_H}, one can further calculate that for all $\theta \in \mR^d$, $x\in \mR^m$,
\begin{eqnarray}
\ip{\theta}{H_{\lambda, c}(\theta, x)} &=& \sum_{i=1}^d \theta^{(i)}\cdot \widehat G_{\lambda, c}^{(i)} (\theta, x)  +  \eta\frac{|\theta|^{2r+2}}{1 + \sqrt{\lambda}|\theta|^{2r}} \nonumber \nonumber \\
&\geq& \sum_{i=1}^d |\theta^{(i)}| \bigg(-\frac{1}{\sqrt{\lambda}} - \sqrt{\lambda}\bigg)  +  \eta\frac{|\theta|^{2r+2}}{1 + \sqrt{\lambda}|\theta|^{2r}} \nonumber \nonumber \\
&\geq& - \bigg(\frac{1}{\sqrt{\lambda}} + \sqrt{\lambda}\bigg) \sqrt{d}|\theta|  + \eta\frac{|\theta|^{2r+2}}{1 + \sqrt{\lambda}|\theta|^{2r}}, \label{eq:iptheatH}
\end{eqnarray}
which implies that
\begin{eqnarray}
-\frac{2\lambda}{|\theta_n^\lambda|^2} \mE\left[\ip{\theta_n^\lambda}{H_{\lambda, c}(\theta_n^\lambda, X_{n+1})} \bigg|\theta_n^\lambda \right]&\leq&
\frac{2\sqrt d}{|\theta_n^\lambda|}\bigg(\sqrt{\lambda} + \lambda^{\frac{3}{2}}\bigg)  - \frac{2 \eta\lambda|\theta_n^\lambda|^{2r}}{1 + \sqrt{\lambda}|\theta_n^\lambda|^{2r}}. \label{eq:H1}
\end{eqnarray}

Moreover, using \eqref{eq:ineq_H}, it is shown that, for all $\theta\in \mR^d, x\in \mR^m$,
\begin{eqnarray}
|H_{\lambda,c}(\theta,x)|^2 &=& \sum_{i=1}^d \bigg( \widehat G^{(i)}_{\lambda, c}(\theta, x)  + \eta\frac{\theta^{(i)}|\theta|^{2r}}{1 + \sqrt{\lambda}|\theta|^{2r}}\bigg)^2  \nonumber \\
&\leq& \sum_{i=1}^d \bigg( 2|\widehat G^{(i)}_{\lambda, c}(\theta, x)|^2  + 2\eta^2\frac{|\theta^{(i)}|^2|\theta|^{4r}}{(1 + \sqrt{\lambda}|\theta|^{2r})^2}\bigg) \nonumber\\
&\leq& 4d \bigg(\frac{1}{\lambda} + \lambda\bigg) + 2\eta^2 |\theta|^2 \frac{|\theta|^{4r}}{(1 + \sqrt{\lambda}|\theta|^{2r})^2}, \label{eq:Hsquared}
\end{eqnarray}
which yields that
\begin{eqnarray}
\frac{\lambda^2}{|\theta_n^\lambda|^2} \mE\left[|H_{\lambda, c}(\theta_n^\lambda, X_{n+1})|^2 \bigg|\theta_n^\lambda \right] 
\leq 4\lambda d \frac{(1+ \lambda^2)}{|\theta_n^\lambda|^2} + 2\lambda\eta^2.  \label{eq:H2}
\end{eqnarray}

Combining \eqref{eq:H1} and \eqref{eq:H2}, one calculates that
\begin{eqnarray}
\mE\left[ |\theta_{n+1}^\lambda|^2|\theta_n^\lambda \right] &=& \mE\left[|\theta_n^\lambda - \lambda H_{\lambda, c}(\theta_n^\lambda, X_{n+1}) + \sqrt{2\lambda\beta^{-1}}\xi_{n+1} |^2|\theta_n^\lambda\right] \nonumber \\
&=& |\theta_n^\lambda|^2\bigg(1 -\frac{2\lambda}{|\theta_n^\lambda|^2} \mE\left[\ip{\theta_n^\lambda}{H_{\lambda, c}(\theta_n^\lambda, X_{n+1})} \bigg|\theta_n^\lambda \right] \nonumber\\
&+& \frac{\lambda^2}{|\theta_n^\lambda|^2} \mE\left[|H_{\lambda, c}(\theta_n^\lambda, X_{n+1})|^2 \bigg|\theta_n^\lambda \right]\bigg) + \frac{2\lambda d}{\beta} \nonumber\\
&\leq& |\theta_n^\lambda|^2\bigg[1 +\frac{2\sqrt d}{|\theta_n^\lambda|}\left(\sqrt{\lambda} + \lambda^{\frac{3}{2}}\right)  - \frac{2 \eta\lambda|\theta_n^\lambda|^{2r}}{1 + \sqrt{\lambda}|\theta_n^\lambda|^{2r}} \nonumber \\
&+& 4\lambda d \frac{(1+ \lambda^2)}{|\theta_n^\lambda|^2} + 2\lambda\eta^2\bigg] + \frac{2\lambda d}{\beta} \nonumber \\
&=& |\theta_n^\lambda|^2 \left(1 - f^\lambda (\theta_n^\lambda)\right) + \frac{2\lambda d}{\beta} \label{eq:thetan+1},
\end{eqnarray}
where $f^\lambda (\theta):=-\frac{2\sqrt d}{|\theta|}\left(\sqrt{\lambda} + \lambda^{\frac{3}{2}}\right)  + \frac{2 \eta\lambda|\theta|^{2r}}{1 + \sqrt{\lambda}|\theta|^{2r}} - 4\lambda d \frac{(1+ \lambda^2)}{|\theta|^2} - 2\lambda\eta^2$ for all $\theta \in \mR^d \setminus\mathbf{0}$. We note that, for all $0<\lambda \leq \lambda_{\max} \leq (4\eta^2)^{-1}$,
$$
\lim_{|\theta|\rightarrow \infty} f^\lambda(\theta) = 2\eta \sqrt \lambda(1-\sqrt \lambda \eta) \geq \eta \sqrt \lambda >0.
$$

In addition, using the fact that $f(s):=s/(1+\sqrt\lambda s)$ is non-decreasing for all $s \geq 0$, one can choose $M_0>0$ such that
\begin{equation}
f^\lambda(\theta) \geq \eta \sqrt\lambda (1-\sqrt\lambda \eta)\geq \frac{\eta\sqrt\lambda}{2}, \label{eq:flambda}
\end{equation}
for all $\theta \geq M_0$, $0<\lambda\leq\lambda_{\max}$. Therefore, from \eqref{eq:thetan+1} and \eqref{eq:flambda}, we have that
\begin{eqnarray}
\mE\bigg[ |\theta_{n+1}^\lambda|^2 {\bf 1}_{|\theta_n^\lambda| \geq M_0} \bigg| \theta_n^\lambda\bigg]
&\leq& |\theta_n^\lambda|^2\bigg(1-\frac{\eta \sqrt{\lambda}}{2}\bigg)  + \frac{2\lambda d}{\beta} \label{eq: A_nm}.
\end{eqnarray}

Let us consider the case of $|\theta_n^\lambda| < M_0$. From \eqref{eq:iptheatH} and \eqref{eq:Hsquared}, we have for all $|\theta| <M_0$, $x\in\mR^m$,
\begin{eqnarray*}
- 2\lambda\ip{\theta}{H_{\lambda,c}(\theta, x)} + \lambda^2 |H_{\lambda, c}(\theta, x)|^2|\theta]
&\leq& 2\bigg(\sqrt{\lambda}+ \lambda^{\frac{3}{2}}\bigg)\sqrt{d} M_0 \\
&+& 2\eta \sqrt\lambda M_0^2 + 4d(\lambda + \lambda^3)
+ 2\eta^2\lambda M_0^2.
\end{eqnarray*}
The above estimate directly yields that, for $0<\lambda\geq\lambda_{\max}$,
\begin{eqnarray}
\mE\bigg[ |\theta_{n+1}^\lambda|^2 {\bf 1}_{|\theta_n^\lambda| < M_0} \bigg| \theta_n^\lambda\bigg]  \nonumber
&\leq& |\theta_n^\lambda|^2 + \frac{2\lambda d}{\beta}  + 2\bigg(\sqrt{\lambda}+ \lambda^{\frac{3}{2}}\bigg)\sqrt{d} M_0 + 2\eta \sqrt\lambda M_0^2 \nonumber \\
&+& 4d(\lambda + \lambda^3) + 2\eta^2 \lambda M_0^2  \nonumber \\
&\leq& \bigg(1-\frac{\eta \sqrt{\lambda}}{2}\bigg)|\theta_n^\lambda|^2  \nonumber  \\
&+&\sqrt \lambda \bigg( \frac{2d}{\beta} + 4\sqrt{d} M_0 +  \frac{5\eta}{2} M_0^2 +  8d +2\eta^2 M_0^2 \bigg)\label{eq: A_nm_complement}.
\end{eqnarray}
Consequently, \eqref{eq: A_nm} and \eqref{eq: A_nm_complement} yield that
\begin{eqnarray*}
\mE\left[ |\theta_{n+1}^\lambda|^2 \bigg| \theta_n^\lambda\right] &\leq& \bigg(1-\frac{\eta \sqrt{\lambda}}{2}\bigg)|\theta_n^\lambda|^2  + \sqrt \lambda \bigg( \frac{2d}{\beta} + 4\sqrt{d} M_0 + \frac{5\eta}{2} M_0^2
+  8d +2\eta^2 M_0^2   \bigg),
\end{eqnarray*}
and
\begin{eqnarray*}
\mE\bigg[ |\theta_{n+1}^\lambda|^2 \bigg] &\leq& \bigg(1-\frac{\eta \sqrt{\lambda}}{2}\bigg)^n\mE|\theta_0^\lambda|^2 \\
&+& \sqrt\lambda \bigg(  \frac{2 d}{\beta} + 4\sqrt{d} M_0 + \frac{5\eta}{2} M_0^2 +  8d +2\eta^2 M_0^2 \bigg) \sum_{j=0}^\infty \bigg(1-\frac{\eta \sqrt{\lambda}}{2}\bigg)^j \\
&\leq&\bigg(1-\frac{\eta \sqrt{\lambda}}{2}\bigg)^n \mE|\theta_0^\lambda|^2 +\bigg( \frac{4 d}{\beta\eta} + \frac{8\sqrt{d} M_0}{\eta}+ 5M_0^2 +  \frac{16d}{\eta} + 4\eta M_0^2  \bigg).
\end{eqnarray*}

\end{proof}
\begin{proof}[\textbf{Proof of Lemma~\ref{lem:l2p-norm}}]\label{proof:l2p-norm}
For any integer $p\geq 2$, $n\in \mN_0$, $|\theta^\lambda_{n+1}|^{2p}$ is written as
$$
|\theta_{n+1}^\lambda|^{2p}  = \bigg(|\Delta_n|^2 + \frac{2\lambda}{\beta}|\xi_{n+1}|^2 + 2\ip{\Delta_{n}}{\sqrt{\frac{2\lambda}{\beta}}\xi_{n+1}} \bigg)^p
$$
where $\Delta_n = \theta_n^\lambda - \lambda H_{\lambda, c}(\theta_n^\lambda, X_{n+1})$. Then, we obtain
\begin{eqnarray}
\mE [|\theta_{n+1}^\lambda|^{2p} | \theta_n^\lambda] &=& \mE\bigg[\bigg(|\Delta_n|^2 + \bigg|\sqrt{\frac{2\lambda}{\beta}}\xi_{n+1}\bigg|^2 + 2\ip{\Delta_{n}}{\sqrt{\frac{2\lambda}{\beta}}\xi_{n+1}} \bigg)^p \bigg| \theta_n^\lambda\bigg] \nonumber \\
&=& \mE[|\Delta_n|^{2p}|\theta_n^\lambda] + 2p \mE \bigg[ |\Delta_n|^{2p-2}\ip{\Delta_n}{\sqrt{\frac{2\lambda}{\beta}}\xi_{n+1}}\bigg|\theta_n^\lambda\bigg]  \nonumber \\
&+& \sum_{\substack{k_1+k_2+k_3=p\\ \{k_1\neq p-1\} \cap \{k_2 \neq 1\}\\ \{k_1 \neq p\} }} \frac{p!}{k_1!k_2!k_3!} \mE \bigg[ |\Delta_n|^{2k_1} \bigg|\sqrt{\frac{2\lambda}{\beta}}\xi_{n+1}\bigg|^{k_2} \nonumber \\
&\times&\bigg|2\ip{\Delta_{n}}{\sqrt{\frac{2\lambda}{\beta}}\xi_{n+1}}\bigg|^{k_3}\bigg| \theta_n^\lambda\bigg] \nonumber \\
&\leq& \mE[|\Delta_n|^{2p}|\theta_n^\lambda] 
+ \sum_{k=2}^{2p} \begin{pmatrix}
2p \\
k
\end{pmatrix}\mE \bigg[ |\Delta_n|^{2p-k} \bigg|\sqrt{\frac{2\lambda}{\beta}}\xi_{n+1}\bigg|^{k} \bigg| \theta_n^\lambda\bigg] \label{eq:theta_2p}
\end{eqnarray}
where the above inequality follows from the result in \cite[Lemma A.3]{chau:19}. Using the fact that $\xi_{n+1}$ is independent of $\Delta_n$ and $\theta_n^{\lambda}$, one further calculates that
\begin{eqnarray*}
\mE [|\theta_{n+1}^\lambda|^{2p} | \theta_n^\lambda] &\leq& \mE[|\Delta_n|^{2p}|\theta_n^\lambda] + \sum_{l=0}^{2p-2} \begin{pmatrix}
2p \\
l+2
\end{pmatrix}\mE \bigg[ |\Delta_n|^{2p-2-l} \bigg|\sqrt{\frac{2\lambda}{\beta}}\xi_{n+1}\bigg|^{l+2} \bigg| \theta_n^\lambda\bigg] \nonumber \\
&=& \mE[|\Delta_n|^{2p}|\theta_n^\lambda] \nonumber \\
&+& \sum_{l=0}^{2p-2} \frac{2p(2p-1)}{(l+2)(l+1)}\begin{pmatrix}
2p-2 \\
l
\end{pmatrix}\mE \bigg[ |\Delta_n|^{2p-2-l} \bigg|\sqrt{\frac{2\lambda}{\beta}}\xi_{n+1}\bigg|^{l+2} \bigg| \theta_n^\lambda\bigg] \nonumber \\
&\leq& \mE[|\Delta_n|^{2p}|\theta_n^\lambda] \nonumber \\
&+& p(2p-1) \mE\bigg[\left(|\Delta_n| +\bigg|\sqrt{\frac{2\lambda}{\beta}}\xi_{n+1}\bigg|\right)^{2p-2} \bigg|\sqrt{\frac{2\lambda}{\beta}}\xi_{n+1}\bigg|^2 \bigg|\theta_n^\lambda\bigg] \nonumber \\
&\leq& \mE[|\Delta_n|^{2p}|\theta_n^\lambda] \nonumber \\
&+& 2^{2p-3}p(2p-1)\bigg(\mE[|\Delta_n|^{2p-2} |\theta_n^\lambda]\frac{2\lambda d}{\beta} + \bigg(\frac{2\lambda}{\beta}\bigg)^p \mE|\xi_{n+1}|^{2p}\bigg).
\end{eqnarray*}
Define $|\Delta_n|^2 = |\theta_n^\lambda|^2 + r_n$ where $r_n = -2\lambda \ip{\ntheta}{H_{\lambda, c}(\ntheta, X_{n+1})} + \lambda^2 |H_{\lambda,c}(\ntheta, X_{n+1})|^2$
to write
\begin{eqnarray}
\mE \left[|\Delta_n|^{2p} | \theta_n^\lambda\right] &=& \sum_{k=0}^p \begin{pmatrix}
p \\
k
\end{pmatrix}|\ntheta|^{2(p-k)}\mE \left[r_n^k | \theta_n^\lambda\right] \nonumber \\
&=& |\ntheta|^{2p} + p|\ntheta|^{2p-2}\mE\left[r_n|\ntheta\right] + \sum_{k=2}^p \begin{pmatrix}
p\\
k
\end{pmatrix} |\ntheta|^{2(p-k)}\mE\left[r_n^k |\ntheta\right]. \label{eq:delta_2p}
\end{eqnarray}

Now, we focus on the case where $|\ntheta|>M$ where
$$
M := \max\bigg\{M_0, 1, \frac{4d}{(2-\eta)\eta}, \frac{2\sqrt{d}}{\eta(2-\eta)}, \frac{2^{2p-2}p(2p-1)d}{\eta\beta}\bigg\}.
$$
Recall that $M_0$ is defined in the proof of Lemma~\ref{lem:l2-norm}. We need the following relations to estimate the moments of $r_n$: for all $x\in \mR^d$, $0<\lambda\leq \lambda_{\max}$, $0<\eta<1$, $|\theta|\geq M$,
\begin{eqnarray}
\lambda^2 |H_{\lambda, c}(\theta, x)|^2 &\leq& 4d (\lambda + \lambda^3) + 2\eta^2 \lambda|\theta|^2 \frac{\lambda|\theta|^{4r}}{(1 + \sqrt{\lambda}|\theta|^{2r})^2} \nonumber \\
&\leq& 4d\lambda (1 + \lambda^2) + 2\eta^2 \lambda|\theta|^2 \nonumber \\
&\leq& 4d\lambda (1 + \lambda^2)|\theta| + 2\eta^2 \lambda|\theta|^2 \nonumber \\
&\leq& 2\sqrt \lambda\eta\bigg(\frac{4d}{|\theta|\eta} + \eta \bigg)|\theta|^2  \nonumber \\
&\leq& 2\sqrt \lambda\eta\bigg(\frac{4d}{M\eta} + \eta \bigg)|\theta|^2  \nonumber \\
&\leq& 4\sqrt \lambda\eta|\theta|^2, \label{eq: Hsquared_2p}
\end{eqnarray}

where we have used the inequality~\eqref{eq:Hsquared} and
$$
M > \frac{4d}{(2-\eta)\eta}  \Leftrightarrow \bigg(\frac{4d}{M\eta} + \eta \bigg) <2.
$$
Notice that $\frac{4d}{(2-\eta)\eta}$ is finite due to $\lambda_{\max}$ being less than $\frac{1}{4\eta^2}$.
Moreover, from \eqref{eq:iptheatH}, we have the following inequality, for $0<\lambda \leq \lambda_{\max}$,
\begin{eqnarray}
|2\lambda \ip{\theta}{H_{\lambda, c}(\theta, x)}| &\leq&  2(\sqrt{\lambda} + \lambda^{1.5})\sqrt{d} |\theta|  + 2\eta\sqrt\lambda |\theta|^2\frac{\sqrt\lambda|\theta|^{2r}}{1 + \sqrt{\lambda}|\theta|^{2r}} \nonumber \\
&\leq& 2\sqrt\lambda(1+ \lambda)\sqrt d |\theta| + 2\eta\sqrt\lambda|\theta|^2 \nonumber \\
&\leq&2\sqrt\lambda \eta \bigg(\frac{2\sqrt d}{|\theta|\eta} + \eta\bigg)|\theta|^2 \nonumber \\
&\leq&2\sqrt\lambda\eta\bigg(\frac{2\sqrt d}{M\eta} +\eta\bigg)|\theta|^2  \nonumber \\
&\leq&4\sqrt\lambda\eta|\theta|^2, \label{eq:inner_H_2_2p}
\end{eqnarray}
where the last inequality holds since
$$
M > \frac{2\sqrt{d}}{\eta(2-\eta)} \Leftrightarrow \bigg(\frac{2\sqrt d}{M\eta} + \eta\bigg) \leq 2.
$$ Thus, $r_n^k$ can be written as
\begin{eqnarray*}
\mE [{\bf 1}_{\{|\ntheta| > M\}}|r_n|^k|\ntheta] &=& \mE \bigg[ {\bf 1}_{\{|\ntheta| > M\}}\bigg(-2\lambda \ip{\ntheta}{H_{\lambda, c}(\ntheta, X_{n+1})} \nonumber \\
&+& \lambda^2 |H_{\lambda,c}(\ntheta, X_{n+1})|^2\bigg)^k\bigg|\ntheta\bigg] \\
&\leq& \mE \bigg[{\bf 1}_{\{|\ntheta| > M\}}\bigg(|2\lambda \ip{\ntheta}{H_{\lambda, c}(\ntheta, X_{n+1})}| \\
&+& \lambda^2 |H_{\lambda,c}(\ntheta, X_{n+1})|^2\bigg)^k\bigg|\ntheta\bigg] \\
&\leq&  \mE \bigg[{\bf 1}_{\{|\ntheta| > M\}} (8\sqrt\lambda\eta|\theta_n^\lambda|^2)^k\bigg|\ntheta\bigg] \\
&\leq&  \lambda^{\frac{k}{2}} (8\eta)^k|\ntheta|^{2k}.
\end{eqnarray*}
Moreover, \eqref{eq:flambda} implies that
$$
\mE[{\bf 1}_{\{|\ntheta|> M\}}r_n|\ntheta] \leq -\frac{\eta\sqrt \lambda}{2} |\ntheta|^2,
$$
or, equivalently,
\begin{equation}
p|\ntheta|^{2p-2} \mE[{\bf 1}_{\{|\ntheta| > M\}}r_n|\ntheta] \leq -p\frac{\eta\sqrt \lambda}{2} |\ntheta|^{2p}. \label{eq:rn_2p}
\end{equation}

Using \eqref{eq:rn_2p}, the $L_{2p}$-norm of $\Delta_n$ conditional on $\ntheta>M$ is given by
\begin{eqnarray}
\mE \bigg[ {\bf 1}_{\{|\ntheta| > M\}}|\Delta_n|^{2p} \bigg| \theta_n^\lambda\bigg] &\leq& |\ntheta|^{2p} + p|\ntheta|^{2p-2}\mE[{\bf 1}_{\{\ntheta > M\}}r_n|\ntheta] \nonumber \\
&+& \sum_{k=2}^p \begin{pmatrix}
p\\
k
\end{pmatrix} |\ntheta|^{2(p-k)}\mE[{\bf 1}_{\{\ntheta > M\}}|r_n|^k |\ntheta] \nonumber \\
&\leq& |\ntheta|^{2p} - p\frac{\eta\sqrt \lambda}{2} |\ntheta|^{2p} + \sum_{k=2}^p \begin{pmatrix}
p\\
k
\end{pmatrix} |\ntheta|^{2(p-k)}  \lambda^{\frac{k}{2}}(8\eta)^{k} |\ntheta|^{2k} \nonumber \\
&\leq& |\ntheta|^{2p} - p\frac{\eta\sqrt \lambda}{2} |\ntheta|^{2p} + |\ntheta|^{2p} \sum_{k=2}^p \begin{pmatrix}
p\\
k
\end{pmatrix} \lambda^{\frac{k}{2}}(8\eta)^{k}.  \label{eq:delta_2p_cond1}
\end{eqnarray}

Moreover, it follows that, for $0<\lambda\leq \lambda_{\max}$,
$$
\lambda \leq \frac{1}{(2^7\eta {}_{p}\cC_{\ceil{\frac{p}{2}}})^2} =
\frac{1}{2^8(8\eta)^2 {}_p\cC_{\ceil{\frac{p}{2}}}^2}
\leq \frac{1}{ 2^{\frac{8}{k-1}}(8\eta)^2 ({}_p\cC_{\ceil{\frac{p}{2}}}^2)^{\frac{2}{k-1}}},
$$
which is equivalent to
\begin{eqnarray*}
\lambda^{\frac{k-1}{2}}&\leq& \frac{1}{2^4 (8\eta)^{k-1}{}_p\cC_{\ceil{\frac{p}{2}}}} \\
&=& \frac{\eta}{2(8\eta)^{k} {}_p\cC_{\ceil{\frac{p}{2}}} },
\end{eqnarray*}
for all $k \in [2, p]\cap \mN$. Then, one observes the following inequality
\begin{eqnarray*}
\sum_{k=2}^p {}_p\cC_k \lambda^{\frac{k}{2}}(8\eta)^{k} &\leq &\sum_{k=2}^p {}_p\cC_{\ceil{\frac{p}{2}}} \lambda^{\frac{k}{2}}(8\eta)^{k}  \\
&\leq& \frac{1}{2}\sum_{k=2}^p \sqrt \lambda \eta \\
&=& \frac{p-2}{2} \sqrt{\lambda}\eta,
\end{eqnarray*}
to obtain
\begin{equation}
\mE \bigg[ {\bf 1}_{\{|\ntheta| > M\}}|\Delta_n|^{2p} \bigg| \theta_n^\lambda\bigg] \leq
(1- \eta\sqrt \lambda)|\ntheta|^{2p}, \label{eq:theta_2p_final}
\end{equation}
and
\begin{equation}
\mE \bigg[ {\bf 1}_{\{|\ntheta| > M\}}|\Delta_n|^{2p-2} \bigg| \theta_n^\lambda\bigg] \leq
(1- \eta\sqrt \lambda)|\ntheta|^{2(p-2)} \leq  \frac{1}{M^2}(1- \eta\sqrt \lambda)|\ntheta|^{2p}. \label{eq:theta_2p-2_final}
\end{equation}

By combining \eqref{eq:theta_2p}, \eqref{eq:theta_2p-2_final} and \eqref{eq:theta_2p_final}, we derive
\begin{eqnarray}
\mE [{\bf 1}_{\{|\ntheta| > M\}}|\theta_{n+1}^\lambda|^{2p} | \theta_n^\lambda] &\leq&
(1- \eta\sqrt \lambda)|\ntheta|^{2p}
\nonumber \\
&+&\frac{2^{2p-2}p(2p-1)\lambda d}{M^2\beta}(1- \eta\sqrt \lambda)|\ntheta|^{2p} \nonumber \\
&+& 2^{2p-3}p(2p-1)\bigg(\frac{2\lambda}{\beta}\bigg)^p \mE|\xi_{n+1}|^{2p} \nonumber \\
&\leq&(1- \eta\sqrt \lambda)\bigg(1+\frac{2^{2p-2}p(2p-1)\lambda d}{M^2\beta}\bigg)|\ntheta|^{2p}  \nonumber \\
&+& 2^{2p-3}p(2p-1)\bigg(\frac{2\lambda}{\beta}\bigg)^p \mE|\xi_{n+1}|^{2p}  \nonumber \\
&\leq& (1- \eta^2\lambda)|\ntheta|^{2p}
+ 2^{2p-3}p(2p-1)\bigg(\frac{2\lambda}{\beta}\bigg)^p \mE|\xi_{n+1}|^{2p},
\end{eqnarray}
where we used the fact that $M\geq\frac{2^{2p-2}p(2p-1)d}{\eta\beta}$ for the last inequality.

Consider the case of $|\ntheta|\leq M$. By observing that from \eqref{eq:Hsquared}
\begin{eqnarray*}
{\bf 1}_{\{|\theta| \leq M\}}\lambda^2 |H_{\lambda, c}(\theta, x)|^2 &\leq & \lambda \left(8d + 2\eta^2 M^2\right),
\end{eqnarray*}
and
\begin{eqnarray*}
{\bf 1}_{\{|\theta| \leq M\}}|2\lambda \ip{\theta}{H_{\lambda, c}(\theta, x)}| &\leq& 2\lambda \sqrt {|\theta|}\sqrt{|H_{\lambda, c}(\theta, x)|} \\
&\leq&  2\lambda \sqrt M \sqrt{ |G(\theta, x)|+d\sqrt\lambda + 2\eta M^{2r+1}} \\
&\leq& 2\lambda \sqrt M \sqrt { |K_G(x)|(1+M^q)+d\sqrt\lambda + 2\eta M^{2r+1}},
\end{eqnarray*}

it can be shown that
\begin{eqnarray*}
\mE\bigg[{\bf 1}_{\{|\ntheta| \leq M\}}|r_n|^k\bigg| \theta_n^\lambda\bigg] &=&  \mE \bigg[ {\bf 1}_{\{|\ntheta| \leq M\}}\bigg(|2\lambda \ip{\ntheta}{H_{\lambda, c}(\ntheta, X_{n+1})}| \\
&+& \lambda^2 |H_{\lambda,c}(\ntheta, X_{n+1})|^2\bigg)^k \bigg| \theta_n^\lambda\bigg]\\
&\leq& \mE \bigg[ {\bf 1}_{\{|\ntheta| \leq M\}} \bigg(2\lambda \sqrt M \sqrt{ K_G(X_{n+1})(1+M^q)+d + 2\eta M^{2r+1}}\\
&+&\lambda \bigg(8d + 2\eta^2 M^2\bigg)\bigg)^k\bigg| \theta_n^\lambda\bigg] \\
&\leq& \widetilde D_k\lambda^k,
\end{eqnarray*}
where $\widetilde D_k = 2^{k-1}\bigg(  (2 \sqrt M)^k (\mE [K_G(X_0)](1+M^q)+d + 2\eta M^{2r+1})^{k/2} + (8d + 2\eta^2 M^2)^k\bigg)$. Hence, one calculates that
\begin{eqnarray*}
\mE \bigg[ {\bf 1}_{\{|\ntheta| \leq M\}}|\Delta_n|^{2p} \bigg| \theta_n^\lambda\bigg] &\leq& |\ntheta|^{2p} + \sum_{k=1}^p \begin{pmatrix}
p\\
k
\end{pmatrix} |\ntheta|^{2(p-k)}\mE[{\bf 1}_{\{\ntheta \leq M\}}|r_n|^k |\ntheta] \nonumber \\
&\leq& (1-\eta^2 \lambda)|\ntheta|^{2p} + \eta^2  \lambda M^{2p}  + M^{2p}\lambda \sum_{k=1}^p \begin{pmatrix}
p\\
k
\end{pmatrix} \lambda^{k-1}\widetilde D_k,
\end{eqnarray*}
and
\begin{eqnarray*}
\mE \bigg[ {\bf 1}_{\{\ntheta \leq M\}}|\Delta_n|^{2p-2} \bigg| \theta_n^\lambda\bigg] &\leq&  \sum_{k=0}^{p-1} \begin{pmatrix}
p-1\\
k
\end{pmatrix} |\ntheta|^{2(p-1-k)}\mE[{\bf 1}_{\{|\ntheta| \leq M\}}|r_n|^k |\ntheta] \nonumber \\
&\leq& M^{2p-2}\sum_{k=0}^{p-1} \begin{pmatrix}
p\\
k
\end{pmatrix} \widetilde D_k \lambda^k.
\end{eqnarray*}
Consequently, we obtain
\begin{eqnarray}
\mE [{\bf 1}_{\{|\ntheta| \leq M\}}|\theta_{n+1}^\lambda|^{2p} | \theta_n^\lambda] &\leq&
(1-\eta^2 \lambda)|\ntheta|^{2p} + \eta^2\lambda M^{2p}  + \lambda M^{2p}\sum_{k=1}^p \begin{pmatrix}
p\\
k
\end{pmatrix} \lambda^{k-1}\widetilde D_k \nonumber \\
&+&  \frac{\lambda d}{\beta}2^{2p-2}p(2p-1)  M^{2p-2}\sum_{k=0}^{p-1} \begin{pmatrix}
p\\
k
\end{pmatrix} \lambda^{k}\widetilde D_k \\
&+& 2^{2p-3}p(2p-1)\bigg(\frac{2\lambda}{\beta}\bigg)^p \mE|\xi_{n+1}|^{2p} . \nonumber
\end{eqnarray}

By defining
\begin{eqnarray*}
A_p &=&   \eta^2  M^{2p}  + M^{2p}\sum_{k=1}^p \begin{pmatrix}
p\\
k
\end{pmatrix} \widetilde D_k  \nonumber \\
&+&  2^{2p-3}p(2p-1)  \bigg(\frac{2dM^{2p-2}}{\beta}\sum_{k=0}^{p-1} \begin{pmatrix}
p\\
k
\end{pmatrix} \lambda^{k}\widetilde D_k +\frac{2}{\beta} \bigg(\frac{2}{\beta}\bigg)^{p-1} d^p (2p-1)!! \bigg), \nonumber
\end{eqnarray*}
we conclude that
\begin{eqnarray*}
\mE |\theta_{n+1}^\lambda|^{2p} &\leq& (1-\eta^2 \lambda) \mE|\ntheta|^{2p} + \lambda A_p \\
&\leq &
(1-\eta^2\lambda)^n \mE|\theta_0^\lambda|^{2p} + \lambda A_p \sum_{j=0}^\infty (1-\eta^2 \lambda)^j \\
&\leq& (1-\eta^2\lambda)^n \mE|\theta_0^\lambda|^{2p} + \frac{A_p }{\eta^2}.
\end{eqnarray*}

\end{proof}

\begin{proof}[\textbf{Proof of Lemma~\ref{lem3.7}}]\label{proof:V24}
For $p\geq 1$, $0<\lambda \leq \lambda_{\max}$, $t\in (nT, (n+1)T]$, $n\in \mN_0$, It\^o's formula yields that
\begin{align}\label{zetaprocito}
\mE[V_p(\bar{\zeta}_t^{\lambda,n})]
&= \mE[V_p( \bar{\theta}^{\lambda}_{nT})] + \int_{nT}^t \mE\left[\lambda \frac{\Delta V_p(\bar{\zeta}_s^{\lambda,n})}{\beta} - \lambda \langle h(\bar{\zeta}_s^{\lambda,n}), \nabla V_p(\bar{\zeta}_s^{\lambda,n}) \rangle\right] \rmd s \nonumber\\
&\quad +\mE\left[\int_{nT}^t \left\langle   \nabla V_p(\bar{\zeta}_s^{\lambda,n}),\sqrt{2\lambda\beta^{-1}}\,\rmd B^{\lambda}_s \right\rangle \right] \nonumber\\
& = \mE[V_p( \bar{\theta}^{\lambda}_{nT})] + \int_{nT}^t \mE\left[\lambda \frac{\Delta V_p(\bar{\zeta}_s^{\lambda,n})}{\beta} - \lambda \langle h(\bar{\zeta}_s^{\lambda,n}), \nabla V_p(\bar{\zeta}_s^{\lambda,n}) \rangle\right] \rmd s.
\end{align}
Then, differentiating both sides of \eqref{zetaprocito} and applying Lemma~\ref{lem:drift_lyapunov}, we have
\begin{eqnarray*}
\frac{\rmd}{\rmd t} \mE[V_p(\bar{\zeta}_t^{\lambda,n})]
&=& \mE\left[ \lambda \frac{\Delta V_p(\bar{\zeta}_t^{\lambda,n})}{\beta} - \lambda \langle h(\bar{\zeta}_t^{\lambda,n}), \nabla V_p(\bar{\zeta}_t^{\lambda,n}) \rangle\right] \\
&\leq& -\lambda \bar{c}(p)\mE[V_p(\bar{\zeta}_t^{\lambda,n})] + \lambda \tilde{c}(p).
\end{eqnarray*}
Therefore, we have the following inequality:
\begin{eqnarray*}
\mE[V_p(\bar{\zeta}_t^{\lambda,n})] \leq e^{-\lambda (t-nT) \bar{c}(p)} \mE[V_p(\bar{\theta}_{nT}^{\lambda})] + \frac{\tilde{c}(p)}{\bar{c}(p)}\left(1-e^{-\lambda\bar{c}(p)(t-nT)}\right).
\end{eqnarray*}

By setting $p=4$ and applying Lemma~\ref{lem:V4}, we obtain the desired result:
\begin{eqnarray*}
\mE[V_4(\bar{\zeta}_t^{\lambda,n})] &\leq& e^{-\lambda (t-nT) \bar{c}(4)} \mE[V_4(\bar{\theta}_{nT}^{\lambda})] + \frac{\tilde{c}(4)}{\bar{c}(4)}\left(1-e^{-\lambda\bar{c}(4)(t-nT)}\right) \\
&\leq& 2 +  2\mE[|\theta_0|^4] + 2\frac{A_4}{\eta^2} +  \frac{\tilde{c}(4)}{\bar{c}(4)}.
\end{eqnarray*}

\end{proof}

\begin{proof}[\textbf{Proof of Lemma~\ref{Lemma 4.7}}]\label{proof:thetazeta}

We begin by applying It\^o's formula to observe,  for all $n\in\mN_0$ and $t\in (nT, (n+1)T]$,
\begin{align}
    W_2^2\left(\mathcal{L}(\bar{\theta}_t^\lambda), \mathcal{L}(\bar{\zeta}_t^{\lambda,n})\right)&\leq \mathbb{E}\left[|\bar{\theta}^\lambda_t-\bar{\zeta}_{t}^{\lambda, n}|^2\right] \nonumber \\
    &=
     -2\lambda\int_{nT}^{t} \mE \left[ \langle \bar{\zeta}_{s}^{\lambda, n}-\bar{\theta}^\lambda_s,h(\bar{\zeta}_{s}^{\lambda, n})- H_{\lambda}(\bar{\theta}^\lambda_{\lfloor s \rfloor},X_{\left \lceil s  \right \rceil }) \rangle\right] \rd s  \nonumber \\
    &=-2\lambda \int_{nT}^{t}\mE \left[ \ip{\bar{\zeta}_{s}^{\lambda, n} - \btheta_s}{h(\bar{\zeta}_{s}^{\lambda, n}) - h(\btheta_s)} \right] \rd s \nonumber \\
    &-2\lambda \int_{nT}^{t}\mE  \left[\ip{\bar{\zeta}_{s}^{\lambda, n} - \btheta_s}{h(\btheta_s) - h(\btheta_{\lfloor s \rfloor})} \right] \rd s \nonumber \\
    &-2\lambda \int_{nT}^{t}\mE  \left[\ip{\bar{\zeta}_{s}^{\lambda, n} - \btheta_s}{h(\btheta_{\lfloor s \rfloor}) - H(\btheta_{\lfloor s \rfloor}, X_{\lceil s \rceil})}\right]  \rd s \nonumber \\
    &-2\lambda \int_{nT}^{t}\mE\left[\langle \bar{\zeta}_{s}^{\lambda, n}-\bar{\theta}^\lambda_s,  H(\bar{\theta}^\lambda_{\left \lfloor s \right \rfloor},X_{\left \lceil s  \right \rceil }) - H_{\lambda}(\bar{\theta}^\lambda_{\lfloor s \rfloor},X_{\left \lceil s  \right \rceil })\rangle \right]\rd s. \label{L2_estimate}
\end{align}
Furthermore, by applying Proposition~\ref{prop:2.6} to the first term of \eqref{L2_estimate}, and by applying Young's inequality, i.e., $2ab \leq a(2b) \leq L_Ra^2/2 + 2b^2/L_R$ for $a,b \geq0$, to the second and fourth term of \eqref{L2_estimate}, one obtains that
\begin{align}
    \mathbb{E}\left[|\bar{\theta}^\lambda_t-\bar{\zeta}_{t}^{\lambda, n}|^2\right]  &\leq 2\lambda L_R  \int_{nT}^t \mE \left[|\bar{\zeta}_{s}^{\lambda, n} - \btheta_s|^2 \right]\rd s \nonumber \\
    & + \frac{\lambda L_R }{2} \int_{nT}^t \mE \left[|\bar{\zeta}_{s}^{\lambda, n} - \btheta_s|^2\right] \rd s + \int_{nT}^t \frac{2\lambda}{L_R} \mE\left[|h(\btheta_s) - h(\btheta_{\lfloor s \rfloor})|^2\right] \rd s \nonumber \\
    & + \int_{nT}^{t} \bigg(-2\lambda \mE  \left[\ip{\bar{\zeta}_{s}^{\lambda, n} - \btheta_s}{h(\btheta_{\lfloor s \rfloor}) - H(\btheta_{\lfloor s \rfloor}, X_{\lceil s \rceil})}\right] \bigg) \rd s \nonumber \\
    & + \frac{\lambda L_R }{2} \int_{nT}^t\mE \left[|\bar{\zeta}_{s}^{\lambda, n} - \btheta_s|^2\right] \rd s \nonumber \\
    &+ \int_{nT}^t \frac{2\lambda}{L_R} \mE\left[|H(\btheta_{\lrf s}, X_{\lceil s \rceil}) - H_{\lambda, c}(\btheta_{\lfloor s \rfloor}, X_{\lceil s \rceil})|^2\right] \rd s\nonumber \\
    & = 3 \lambda L_R \int_{nT}^t \mE \left[|\bar{\zeta}_{s}^{\lambda, n} - \btheta_s|^2\right] \rd s + \int_{nT}^t\left(A_s^{\lambda, n} +  B_s^{\lambda, n} +D_s^{\lambda, n}\right)\rd s,
\end{align}
where
\begin{eqnarray*}
A_t^{\lambda, n} &:=& \frac{2\lambda}{L_R} \mE\left[|h(\btheta_t) - h(\btheta_{\lfloor t \rfloor})|^2\right]\\
B_t^{\lambda, n} &:=& -2\lambda \mE \left[ \ip{\bar{\zeta}_{t}^{\lambda, n} - \btheta_t}{h(\btheta_{\lfloor t \rfloor}) - H(\btheta_{\lfloor t \rfloor}, X_{\lceil t \rceil})}\right]  \\
D_t^{\lambda, n} &:=& \frac{2\lambda}{L_R} \mE\left[|H(\btheta_{\lrf t}, X_{\lceil t \rceil}) - H_{\lambda, c}(\btheta_{\lfloor t \rfloor}, X_{\lceil t \rceil})|^2\right].
\end{eqnarray*}


In addition, using the definition of $\btheta_t$ and the inequality of \eqref{eq:Hsquared}, we have
\begin{eqnarray*}
|\btheta_t -\btheta_{\lfloor t\rfloor}|^4 &\leq& \bigg|\lambda \int_{\lrf t}^t |H_{\lambda,c }(\btheta_{\lrf s}, X_{\lrc s})| ds + \sqrt{2\lambda\beta^{-1}} |B_t^\lambda - B_{\lrf t}^\lambda |\bigg|^4 \\
&\leq& 8\lambda^2 \bigg( \lambda^2|H_{\lambda,c }(\btheta_{\lrf t}, X_{\lrc t})|^4 + 4\beta^{-2} |B_t^\lambda - B_{\lrf t}^\lambda |^4 \bigg) \\
&\leq& 8\lambda^2 \bigg((4d(1+\lambda^2) + 2\eta^2|\btheta_{\lrf t}|^2)^2 +  4\beta^{-2} |B_t^\lambda - B_{\lrf t}^\lambda |^4\bigg) \\
&\leq& 2^5\lambda^2 \bigg( 2^3d^2(1+\lambda^4) + \eta^4 |\btheta_{\lrf t}|^4+  \beta^{-2} |B_t^\lambda - B_{\lrf t}^\lambda |^4\bigg),
\end{eqnarray*}
which yields that
\begin{eqnarray}
\sqrt{ \mE |\btheta_t -\btheta_{\lrf t}|^4} &\leq& 2^{5/2} \sqrt{16d^2 +\eta^4 \mE\left[|\btheta_{\lrf t}|^4\right] + \beta^{-2}\mE\left[|B_t^\lambda - B_{\lrf t}^\lambda|^4\right]} \lambda\nonumber \\
&\leq&  \widetilde {C}_1 \lambda \label{ineq:diff_theta_4}
\end{eqnarray}
where Lemma~\ref{lem:l2p-norm} is used for the first inequality, and
$$
\widetilde {C}_1:=2^{5/2}\sqrt{16d^2 + \eta^4(\mE\left[|\theta_0|^4\right]+A_{2}/\eta^2)+ \frac{3}{\beta^2}d^2}.
$$

Using Remark~\ref{rem:h_lipschitz}, Lemma~\ref{lem:l2p-norm} and \eqref{ineq:diff_theta_4}, $A_t^{\lambda, m}$ can be bounded as follows:
\begin{eqnarray}
A_t^{\lambda, n} &\leq& \frac{2\lambda L_h^2}{L_R} \mE\left[(1+|\btheta_t| + |\btheta_{\lfloor t \rfloor})^{2l}|\btheta_t - \btheta_{\lfloor t \rfloor}|^2\right]  \nonumber \\
&\leq& \frac{2\lambda L_h^2}{L_R} \sqrt{\mE \left[(1+|\btheta_t| + |\btheta_{\lfloor t \rfloor})^{4l}\right]}\sqrt{\mE\left[|\btheta_t - \btheta_{\lfloor t \rfloor}|^4\right] }  \nonumber \\
&\leq& \frac{2\lambda L_h^2}{L_R}  3^{2l}\sqrt{(1+\mE\left[|\btheta_t|^{4l}\right] +\mE\left[ |\btheta_{\lfloor t \rfloor}|^{4l}\right])}\sqrt{\mE\left[|\btheta_t - \btheta_{\lfloor t \rfloor}|^4\right]}  \nonumber \\
&\leq&\bar{C}_1\lambda^2  \label{eq:A_t}
\end{eqnarray}
where $\bar{C}_1 = \frac{2L_h^29^{l}}{L_R} \sqrt{1+2\mE\left[|\btheta_0|^{4l}\right] +2\frac{A_{2l}}{\eta^2}}\widetilde {C}_1$. To estimate $B_t^{\lambda, n}$, we observe that
\begin{align}
B_t^{\lambda, n} &=  -2\lambda \mE \left[ \ip{\bar{\zeta}_{t}^{\lambda, n} - \btheta_{\lfloor t \rfloor}}{h(\btheta_{\lfloor t \rfloor}) - H(\btheta_{\lfloor t \rfloor}, X_{\lceil t \rceil})}\right]  \nonumber\\
&- 2\lambda \mE \left[\ip{\btheta_{\lfloor t \rfloor}- \btheta_t}{h(\btheta_{\lfloor t \rfloor}) - H(\btheta_{\lfloor t \rfloor}, X_{\lceil t \rceil})}  \right]\nonumber\\
&\leq -2\lambda \mE\bigg[\mE \bigg[ \ip{\bar{\zeta}_{t}^{\lambda, n} - \btheta_{\lfloor t \rfloor}}{h(\btheta_{\lfloor t \rfloor}) - H(\btheta_{\lfloor t \rfloor}, X_{\lceil t \rceil})} \bigg| \bar{\zeta}_{t}^{\lambda, n}, \btheta_{\lfloor t \rfloor}\bigg]\bigg] \nonumber\\
&- 2\lambda \mE \bigg[ \ip{\btheta_{\lfloor t \rfloor}- \btheta_t}{h(\btheta_{\lfloor t \rfloor}) - H(\btheta_{\lfloor t \rfloor}, X_{\lceil t \rceil})} \bigg] \nonumber \\
&\leq - 2\lambda \mE \bigg[ \ip{\lambda\int_{\lfloor t \rfloor}^t H_\lambda(\btheta_{\lfloor s\rfloor}, X_{\lceil s \rceil})\rd s - \sqrt{\frac{2\lambda}{\beta}} B_{t-\lfloor t \rfloor}^\lambda }{ h(\btheta_{\lfloor t \rfloor}) - H(\btheta_{\lfloor t \rfloor}, X_{\lceil t \rceil})} \bigg], \nonumber
\end{align}
where we have used that $H$ is an unbiased estimator of $h$, i.e., $H(\theta, X_0)=h(x)$ for all $x\in \mR^m$, to obtain the last inequality. Moreover,  using Remark~\ref{rem:H_estimate} and Lemma~\ref{lem:l2p-norm}, we have
\begin{align}
B_t^{\lambda, n} &\leq -2\lambda^2 \mE \left[\ip{ H_\lambda(\btheta_{\lfloor t\rfloor}, X_{\lceil t \rceil})}{h(\btheta_{\lfloor t \rfloor}) - H(\btheta_{\lfloor t \rfloor}, X_{\lceil t \rceil})} \right]\nonumber \\
&\leq 2\lambda^2 \mE \left[\ip{ |H_\lambda(\btheta_{\lfloor t\rfloor}, X_{\lceil t \rceil})|}{|h(\btheta_{\lfloor t \rfloor})| + |H(\btheta_{\lfloor t \rfloor}, X_{\lceil t \rceil})|} \right]\nonumber \\
&\leq 4\lambda^2 \mE\left[|H_\lambda(\btheta_{\lfloor t\rfloor}, X_{\lceil t \rceil})|^2\right] \nonumber \\
&\leq 12 \left(\mE\left[|K_G(X_0)|^2(1+\mE\left[|\btheta_{\lfloor t \rfloor}|^{2q}\right])\right] + d +\eta^2 \mE\left[|\btheta_{\lfloor t \rfloor}|^{4r+2}\right]\right)\lambda^2 \nonumber \\
&\leq \bar{C}_2 \lambda^2, \label{eq:B_t},
\end{align}
where the constant $\bar{C}_2$ is given by
\begin{equation*}
\bar{C}_2 = 12\sqrt{\mE\left[|K_G(X_0)|^2\right]\left(1+\mE\left[|\theta_0|^{2q}\right] + \frac{A_q}{\eta^2}\right) +  d +\eta^2 \left(\mE\left[|\theta_0|^{4r+2}\right] + \frac{A_{2r+1}}{\eta^2}\right)}.
\end{equation*}

Moreover, $D_t^{\lambda, n}$ can be estimated as follows, from Remark~\ref{remark:diff_H} and Lemma~\ref{lem:l2p-norm},
\begin{eqnarray}
D_t^{\lambda, n} &\leq& \frac{18\lambda^2}{L_R}\bigg[8\mE\left[|K_G(X_0)|^4\right](1+\mE\left[|\btheta_{\lrf t}|^{4q}\right]) + d + \eta^2 \mE\left[|\btheta_{\lrf t}|^{4r+2}\right] \bigg]  \nonumber \\
&\leq&  \bar{C}_3 \lambda^2, \label{eq:D_t}
\end{eqnarray}
where the independence of $\btheta_{\lfloor s\rfloor}$ and $X_{\lceil s \rceil}$ is used, and $\bar{C}_3$ is given by
$$
\bar{C}_3 = \frac{18}{L_R}\bigg[8\mE\left[|K_G(X_0)|^4\right](1+\mE\left[|\btheta_{0}|^{4q}\right]+ A_{2q}/\eta^2) + d + \eta^2 (\mE\left[|\btheta_{0}|^{4r+2}\right] + A_{2r+1}/\eta^2) \bigg].
$$

Substituting \eqref{eq:A_t}, \eqref{eq:B_t}, and \eqref{eq:D_t} into \eqref{L2_estimate}, one can derive
\begin{eqnarray*}
 \mathbb{E}\left[|\bar{\theta}^\lambda_t-\bar{\zeta}_{t}^{\lambda, n}|^2\right] &\leq& 3\lambda L_R \int_{nT}^t \mE\left[|\bar{\theta}^\lambda_s-\bar{\zeta}_{s}^{\lambda, n}|^2\right] \rd s + \int_{nT}^t (\bar{C}_1+\bar{C}_2 + \bar{C}_3)\lambda^2 \rd s \\
 &\leq& 3\lambda L_R \int_{nT}^t \mE\left[|\bar{\theta}^\lambda_s-\bar{\zeta}_{s}^{\lambda, n}|^2\right] \rd s +  (\bar{C}_1+\bar{C}_2 + \bar{C}_3)\lambda < \infty
\end{eqnarray*}
where the second inequality follows from the fact that $(t-nT) \leq T \leq \frac{1}{\lambda}$
and the use of Gronwall's inequality gives
$$
\mE | \btheta_t - \bar{\zeta}_{t}^{\lambda, n} |^2 \leq e^{3 L_R}(\bar{C}_1+\bar{C}_2+\bar{C}_3) \lambda.
$$
\end{proof}

\begin{proof}[\textbf{Proof of Lemma~\ref{Lemma 4.8}}]\label{proof:zetaZ1}
Recall that $Z_t^\lambda = \bar{\zeta}_{t}^{\lambda, 0}$. For $t\in (nT, (n+1)T]$, $n\in \mN_0$, we can write
\begin{eqnarray}
W_{1}\left(\mathcal{L}(\bar{\zeta}_{t}^{\lambda, n}), \mathcal{L}(Z_{t}^{\lambda})\right) &\leq&  \sum_{k=1}^n W_{1}\left(\mathcal{L}(\bar{\zeta}_{t}^{\lambda, k}), \mathcal{L}(\bar{\zeta}_{t}^{\lambda, k-1})\right) \nonumber \\
&\leq& \sum_{k=1}^n w_{1,2}\left(\mathcal{L}(\bar{\zeta}_{t}^{\lambda, k}), \mathcal{L}(\bar{\zeta}_{t}^{\lambda, k-1})\right), \label{eq:w1zetaz}
\end{eqnarray}
where we have used the fact $W_1(\mu, \nu) \leq w_{1,2}(\mu, \nu)$ for $\mu, \nu \in \mathcal{P}_{V_2}(\mR^d)$ for the second inequality. Using Lemma~\ref{eberle} and $\lambda (t-kT)\geq n-k$, we further calculate
\begin{align}
& w_{1,2}\left(\mathcal{L}(\bar{\zeta}_{t}^{\lambda, k}), \mathcal{L}(\bar{\zeta}_{t}^{\lambda, k-1})\right) \nonumber \\
&\leq \hat{c} e^{-C_0 \lambda(t-kT)} w_{1,2}\left(\mathcal{L}(\bar \theta_{kT}^\lambda), \mathcal{L}( \bar \zeta_{kT}^{\lambda, k-1})\right) \nonumber \\
&\leq \hat{c} e^{-C_0 (n-k) } w_{1,2}\left(\mathcal{L}(\bar \theta_{kT}^\lambda), \mathcal{L}( \bar \zeta_{kT}^{\lambda, k-1})\right) \nonumber\\
&\leq \hat{c} e^{-C_0 (n-k)} W_{2}\left(\mathcal{L}(\bar \theta_{kT}^\lambda), \mathcal{L}( \bar \zeta_{kT}^{\lambda, k-1})\right) \sqrt{\mE\left[\left|1 + V_2(\bar \theta_{kT}^\lambda) + V_2(\bar \zeta_{kT}^{\lambda, k-1})\right|^2\right]} \nonumber\\
&\leq \hat{c} e^{-C_0 (n-k)} W_{2}\left(\mathcal{L}(\bar \theta_{kT}^\lambda), \mathcal{L}( \bar \zeta_{kT}^{\lambda, k-1})\right) \bigg( 1 + \sqrt{\mE[V_4(\bar \theta_{kT}^\lambda)]} + \sqrt {\mE[V_4(\bar \zeta_{kT}^{\lambda, k-1})]}\bigg) \nonumber\\
&\leq \sqrt \lambda \hat{c} e^{-C_0 (n-k)}\sqrt{e^{3 L_R} (\bar{C}_1+\bar{C}_2+\bar{C}_3)} \bigg(1 + \sqrt{2\mE |\theta_0|^4 + 2 + 2\frac{A_2}{\eta^2}}  \nonumber \\
&+  \sqrt{2\mE |\theta_0|^4 + 2 + 2\frac{A_2}{\eta^2} + \frac{\tilde c(4)}{\bar c(4)}}\bigg) \label{ineq:w_12}
\end{align}
where Lemma~\ref{lem:V4}, \ref{Lemma 4.7} and \ref{lem3.7} are used for the last inequality. By substituting \eqref{ineq:w_12} into \eqref{eq:w1zetaz}, we obtain
\begin{eqnarray*}
W_{1}\left(\mathcal{L}\left(\bar{\zeta}_{t}^{\lambda, n}\right), \mathcal{L}\left(Z_{t}^{\lambda}\right)\right) &\leq& \sqrt \lambda \hat{c}\sqrt{e^{3
L_R} (\bar{C}_1+\bar{C}_2+\bar{C}_3)}\bigg[1 + \sqrt{2\mE |\theta_0|^4 + 2 + 2\frac{A_2}{\eta^2}} \\
&+&  \sqrt{2\mE |\theta_0|^4 + 2 + 2\frac{A_2}{\eta^2} + \frac{\tilde c(4)}{\bar c(4)}}\bigg] \sum_{k=1}^n  e^{-C_0 (n-k)} \\
&\leq& z_1  \sqrt \lambda,
\end{eqnarray*}
where
\begin{eqnarray*}
z_1&=& \frac{\hat{c}}{1-\exp(-C_0)}\sqrt{e^{3L_R} (\bar{C}_1+\bar{C}_2+\bar{C}_3)} \\
&\times&\bigg[1 + \sqrt{2\mE |\theta_0|^4 + 2 + 2\frac{A_2}{\eta^2}}
+  \sqrt{2\mE |\theta_0|^4 + 2 + 2\frac{A_2}{\eta^2} + \frac{\tilde c(4)}{\bar c(4)}}\bigg].
\end{eqnarray*}
\end{proof}

\begin{proof}[\textbf{Proof of Lemma~\ref{Lemma A.5.}}]\label{proof:zetaZ2}
We begin by observing that, for $t\in (nT, (n+1)T]$, $n\in \mN_0$,
\begin{eqnarray*}
W_2 \left(\mathcal{L}(\bar{\zeta}_{t}^{\lambda, k}), \mathcal{L}(\bar{\zeta}_{t}^{\lambda, k-1})\right) &\leq& \sqrt{2 w_{1,2}\left(\mathcal{L}\left(\bar{\zeta}_{t}^{\lambda, k}\right), \mathcal{L}\left(\bar{\zeta}_{t}^{\lambda, k-1}\right)\right)} \\
&\leq& \lambda^{1/4} e^{-C_0 (n-k)/2}\bigg[\hat{c} \sqrt{e^{3 L_R} (\bar{C}_1+\bar{C}_2+\bar{C}_3)}\\
&\times&\bigg(1 + \sqrt{2\mE |\theta_0|^4 + 2 + 2\frac{A_2}{\eta^2}} \\
&+& \sqrt{2\mE |\theta_0|^4 + 2 + 2\frac{A_2}{\eta^2} + \frac{\tilde c(4)}{\bar c(4)}}\bigg)\bigg]^{1/2},
\end{eqnarray*}
where the first inequality holds due to \eqref{ineq:semi_norm}, and the second inequality follows from \eqref{ineq:w_12}. Consequently, we derive
\begin{eqnarray*}
W_2 \left(\mathcal{L}(\bar{\zeta}_{t}^{\lambda, n}), \mathcal{L}(Z_t^\lambda\right)) &\leq& \sum_{k=1}^n W_2 \left(\mathcal{L}\left(\bar{\zeta}_{t}^{\lambda, k}\right), \mathcal{L}(\bar{\zeta}_{t}^{\lambda, k-1})\right) \\
&\leq&  \lambda^{1/4}\bigg[\hat{c} \sqrt{e^{3 L_R} (\bar{C}_1+\bar{C}_2+\bar{C}_3)}\bigg(1 + \sqrt{2\mE |\theta_0|^4 + 2 + 2\frac{A_2}{\eta^2}} \nonumber\\
&+& \sqrt{2\mE |\theta_0|^4 + 2 + 2\frac{A_2}{\eta^2} + \frac{\tilde c(4)}{\bar c(4)}}\bigg)\bigg]^{1/2} \sum_{k=1}^n  e^{-C_0 (n-k)/2} \\
&\leq& \lambda^{1/4} z_2
\end{eqnarray*}
where
$$
z_2 = \frac{\sqrt {(1-\exp(-C_0))z_1}}{1-\exp(-C_0/2)}.
$$
\end{proof}

\section{Table of Constants}
Table~\ref{tab:constant} displays full expressions for constants which appear in the main results of this paper. In addition, Table~\ref{tab:basic constants} shows all main constants and their dependency on key parameters such as $d$ and $\beta$.
\begin{table}[htb]
\renewcommand{\arraystretch}{2}
\centering
\caption{Explicit expressions for main constants.}\label{tab:constant} 

\begin{sc}
\scriptsize
\begin{tabular}{c  c}
\hline
Symbol & Full Expression  \\
\midrule
$\bar{C}_1$ & 
$\frac{L^2 2^{2\rho +5/2}3 ^{2l} }{L_R} (1 + \mE |X_0|^{2\rho})\sqrt{(1+2\mE|\btheta_0|^{4l} +2\frac{A_{2l}}{\eta^2})\left(16d^2 + \eta^4(\mE|\theta_0|^4+A_{2}/\eta^2)+ \frac{3}{\beta^2}d^2\right)}$\\
\hline
$\bar{C}_2$ &
$
12\sqrt{\mE|K_G(X_0)|^2\left(1+\mE|\theta_0|^{2q} + \frac{A_q}{\eta^2}\right) +  d +\eta^2 \left(\mE|\theta_0|^{4r+2} + \frac{A_{2r+1}}{\eta^2}\right)}
$ \\
\hline
$\bar{C}_3$ & $
\frac{18}{L_R}\bigg[8\mE|K_G(X_0)|^4(1+\mE|\btheta_{0}|^{4q}+ A_{2q}/\eta^2) + d + \eta^2 (\mE|\btheta_{0}|^{4r+2} + A_{2r+1}/\eta^2) \bigg].
$ \\ \hline
$z_1$ & $\frac{\hat{c}\sqrt{e^{3L_R} (\bar{C}_1+\bar{C}_2+\bar{C}_3)}}{1-\exp(-C_0)}\bigg[1 + \sqrt{2\mE |\theta_0|^4 + 2 + 2\frac{A_2}{\eta^2}}
+  \sqrt{2\mE |\theta_0|^4 + 2 + 2\frac{A_2}{\eta^2} + \frac{\tilde c(4)}{\overline c(4)}}\bigg]$ \\
\hline
$z_2$ & $\frac{\sqrt {(1-\exp(-C_0))z_1}}{1-\exp(-C_0/2)}$ \\
\hline
$\hat{c}$ & See Lemma~\ref{eberle}. \\
\hline
$C_0$ & See Lemma~\ref{eberle}. \\
\hline
$C_1$ & $(z_1+\sqrt{e^{3 L_R}(\bar{C}_1+\bar{C}_2+\bar{C}_3)})$ \\
\hline
$C_2$ & $\hat{c}  \bigg(1 + \mE[V_2(\theta_0)] + \int_{\mR^d} V_2(\theta) \pi_\beta (d\theta)\bigg)$
\\ \hline
$C_3$ & $\sqrt{e^{3a}(\bar{C}_1+ \bar{C}_2+\bar{C}_3)} + z_2$ \\ \hline
$C_4$ & $\sqrt{2 \hat{c} \left(1+\mathbb{E}\left[V_{2}\left(\theta_{0}\right)\right]+\int_{\mathbb{R}^{d}} V_{2}(\theta) \pi_{\beta}(d \theta)\right)}$ \\ \hline
$C_5$ & $\frac{C_0}{2}$ \\ \hline
$C_6$ & $\bigg( r_1 + \frac{r_12^{2r}}{2r+2}\sqrt{\mE|\theta_0^\lambda|^{4r+2} + \frac{A_{2r+1}}{\eta^2}} + \frac{r_12^{2r}}{2r+2}\sqrt{\mE|Z_\infty|^{4r+2}}\bigg)$ \\ \hline
$C_7$ & $\frac{d}{2}\log \bigg(\frac{K e}{A}\bigg(\frac{B}{d}\beta+1\bigg)\bigg)  +\log 2$ \\ \hline
$A$ & $2^q \mE[1 + K_G(X_0)]$ \\ \hline
$B$ & $3 (2^{q+1} \mE[1+K_G(X_0)])^{q+2} /\eta^{q+1}$ \\ \hline
$K$ & $L\mE[(1+|X_0|)^\rho](1+4R_0)^{2r+1}$ \\\hline
\end{tabular}
\end{sc}
\end{table}

\newpage

\begin{table}[htb]
\renewcommand{\arraystretch}{2}
\centering
    \caption{Main constants and their dependency to key parameters}
    \label{tab:basic constants}
\begin{sc}
\scriptsize
    \begin{tabular}{@{}c|cc@{}}
    \hline
     \multirow{2}{*}{Constant} &  \multicolumn{2}{c}{Key parameters}   \\ \cline{2-3}
    \phantom{Constant} &$d$&$\beta$ \\\hline
$A_p$ & $\mathcal O\left(poly\left(\frac{d}{\beta}\right)\right)$ & $\mathcal O\left(poly\left(\frac{d}{\beta}\right)\right)$ \\ \hline 
$\hat c$&  $\mathcal O\left(e^{poly(d)}\right)$  & $\mathcal O\left(e^{poly\left(\frac{d}{\beta}\right)}\right)$\\ \hline 
$C_0$&  $\mathcal O\left(poly\left(\frac{d}{\beta}\right)\right)$  & $\mathcal O\left(poly\left(\frac{d}{\beta}\right)\right)$\\ \hline 
$C_1$&  $\mathcal O\left(e^{poly(d)}\right)$  & $\mathcal O\left(e^{poly\left(\frac{d}{\beta}\right)}\right)$\\ \hline 
$C_2$&  $\mathcal O\left(e^{poly(d)}\right)$  & $\mathcal O\left(e^{poly\left(\frac{d}{\beta}\right)}\right)$\\ \hline 
$C_3$&  $\mathcal O\left(e^{poly(d)}\right)$  & $\mathcal O\left(e^{poly\left(\frac{d}{\beta}\right)}\right)$\\ \hline 
$C_4$&  $\mathcal O\left(e^{poly(d)}\right)$  & $\mathcal O\left(e^{poly\left(\frac{d}{\beta}\right)}\right)$\\ \hline 
$C_5$&  $\mathcal O\left(poly\left(\frac{d}{\beta}\right)\right)$  & $\mathcal O\left(poly\left(\frac{d}{\beta}\right)\right)$\\ \hline 
$C_6$&  $\mathcal O\left(poly\left(\frac{d}{\beta}\right)\right)$  & $\mathcal O\left(poly\left(\frac{d}{\beta}\right)\right)$\\ \hline 
$C_7$&  $\mathcal O(d\log d)$  & $\mathcal O\left(\log\left(\frac{d}{\beta}\right)\right)$\\ \hline 
\end{tabular}
\end{sc}
\end{table}

\bibliographystyle{plainnat}
\bibliography{references}

\newpage

\end{document}